\documentclass{article}

\PassOptionsToPackage{numbers, compress}{natbib}
 \usepackage[preprint]{nips_2018}

\usepackage[utf8]{inputenc} % allow utf-8 input
\usepackage[T1]{fontenc}    % use 8-bit T1 fonts
\usepackage{hyperref}       % hyperlinks
\usepackage{url}            % simple URL typesetting
\usepackage{booktabs}       % professional-quality tables
\usepackage{amsfonts}       % blackboard math symbols
\usepackage{nicefrac}       % compact symbols for 1/2, etc.
\usepackage{microtype}      % microtypography
\usepackage{makecell}

\usepackage[cmex10]{amsmath}
\usepackage{algorithm}
\usepackage{algpseudocode}
\usepackage[small,bf]{caption}
\usepackage{subcaption}
\usepackage{mathtools}
\DeclarePairedDelimiter{\ceil}{\lceil}{\rceil}

\usepackage{lipsum}
\usepackage{color}
\definecolor{Sijia_color}{rgb}{0.858, 0.188, 0.478}
\definecolor{Lisa_color}{rgb}{0.5, 0.388, 0.878}
\definecolor{BK_color}{rgb}{0.5, 0.8, 0.5}
\definecolor{PY_color}{rgb}{0.3, 0.8, 0.8}

\newtheorem{myprop}{\bf{Proposition}}
\newtheorem{mycor}{\bf{Corollary}}
\newtheorem{mythr}{\bf{Theorem}}
\newtheorem{mylemma}{\bf{Lemma}}
\newtheorem{myremark}{\bf{Remark}}

\DeclarePairedDelimiterX{\inp}[2]{\langle}{\rangle}{#1, #2}

\DeclareMathOperator{\tr}{tr}

\DeclareMathOperator{\cov}{cov}

\DeclareMathOperator*{\minimize}{\text{minimize}}

\DeclareMathAlphabet\mathbfcal{OMS}{cmsy}{b}{n}
\newcommand{\Def}[0]{\mathrel{\mathop:}=}

\usepackage{adjustbox}

\title{\textcolor{black}{Zeroth-Order Stochastic Variance Reduction for Nonconvex Optimization}}

\author{
Sijia Liu$^{1}$~~Bhavya Kailkhura$^{2}$~~Pin-Yu Chen$^{1}$~~Paishun Ting$^{3}$~~\textbf{Shiyu Chang}$^{1}$~~\textbf{Lisa Amini}$^{1}$\\
  $^1$MIT-IBM Watson AI Lab, IBM Research\\
    $^2$Lawrence Livermore National Laboratory \\
    $^3$University of Michigan, Ann Arbor  %\\
}

\begin{document}
% \nipsfinalcopy is no longer used

\maketitle

\vspace*{-0.2in}

\begin{abstract}
\vspace*{-0.1in}
% \textcolor{Lisa_color}{[Lisa's edits]}.
% \textcolor{Sijia_color}{[Sijia's edits]}.
% \textcolor{BK_color}{[BK's edits]}.
% \textcolor{PY_color}{[PY's edits]}.
% lisa: it seems like going through the additional error and the 2 accelerations to mitigate is too detailed for an abstract, so I tried to make it focus on the bigger contributions.  If we go this route, it needs another pass...I don't like the a,b,c wording...
%\textcolor{Lisa_color}{
As application demands for zeroth-order (gradient-free) optimization accelerate, the need for variance reduced and faster converging approaches
is also intensifying. This paper addresses these challenges by 
presenting: a) a comprehensive theoretical analysis of variance reduced zeroth-order (ZO) optimization, b) a novel variance reduced ZO algorithm, called ZO-SVRG, and c) an experimental evaluation of our approach in the context of two compelling applications, black-box chemical material classification and generation of  adversarial examples from black-box deep neural network models. Our theoretical analysis uncovers an essential difficulty in the analysis of ZO-SVRG: the unbiased assumption on gradient estimates 
no longer holds. We prove that compared to its first-order counterpart, ZO-SVRG with a two-point random gradient estimator could suffer an additional error of order $O(1/b)$, where $b$ is the mini-batch size.
To mitigate this error, we propose two accelerated versions of ZO-SVRG utilizing  variance reduced gradient estimators, which achieve  %$O(1/T)$ 
the best rate  known for ZO stochastic optimization (in terms of iterations).
Our extensive experimental results show that our approaches outperform other state-of-the-art ZO algorithms, and strike a balance  between the convergence rate and the function query complexity.
%}
 
%In this paper, we propose and analyze ZO-SVRG, 
%a new zeroth-order (gradient-free) variance reduced  method for nonconvex optimization. 
%The novelty of our work lies at that it   provides
%a  comprehensive study of ZO optimization and stochastic variance reduction. 
%On theoretical side, 
%we show that compared to its first-order counterpart, 
%ZO-SVRG achieves a similar convergence rate that 
%{decays linearly with $T$}
%linearly scales with $O(1/T)$ 
%but suffers an additional error of order $O(1/b)$ %due to the use of random gradient estimators based only on two function evaluations. Here $T$ and $b$ represent the number of iterations and the mini-batch size, respectively. 
%To mitigate  the \textcolor{black}{critical} $O(1/b)$ side effect,  we propose two accelerated versions of ZO-SVRG using gradient estimators of reduced variances, which achieve a faster convergence rate %towards $O(d/T)$ comparably to the zeroth-order (batch) gradient descent method. In brief, our work 
%provides a  comprehensive study on variance reduced ZO optimization.
%On the piratical side,   the effectiveness of our approaches is demonstrated via two novel applications, black-box chemical material classification and generation of  adversarial examples from black-box deep neural network models. 

%Extensive experimental results show that our approaches outperform
% other state-of-the-art zeroth-order   algorithms, and strike a balance  between the convergence rate and the function query complexity.
\end{abstract}

\vspace*{-0.2in}
\section{Introduction}
\vspace*{-0.1in}
% first paragraph needs to strengthen motivation for ZO, needs to be checked for accuracy/intent
Zeroth-order (gradient-free) optimization is increasingly embraced for solving machine learning problems where explicit expressions of the gradients are difficult or infeasible to obtain.
Recent examples have shown zeroth-order (ZO) based generation of prediction-evasive, black-box adversarial attacks on deep neural networks (DNNs) as effective as state-of-the-art white-box attacks, despite leveraging only the inputs and outputs of the targeted DNN \citep{papernot2016practical,madry17,chen2017zoo}. Additional classes of applications include %multi-agent resource management, where each agent can only partially evaluate the objective function 
network control and management with time-varying constraints and limited computation capacity
\citep{chen2017bandit,liu2017zeroth}, and parameter inference of black-box systems \citep{fu2002optimization,lian2016comprehensive}. ZO algorithms achieve gradient-free optimization by approximating the full gradient via gradient estimators based on only the function values \citep{brent2013algorithms,spall2005introduction}.  

%Zeroth-order (gradient-free) optimization approximates the full gradient via a gradient estimator based on only the function values \citep{brent2013algorithms,spall2005introduction}.  It has attracted an increasing amount of attention for solving machine learning problems in scenarios where explicit expressions for the gradients are difficult or infeasible to obtain.  Some recent important applications of zeroth-order (ZO) optimization include: 1) parameter inference of black-box systems \citep{fu2002optimization,lian2016comprehensive}; 2) dynamic network resource management \citep{liu2017zeroth,chen2017bandit}; and 3) generation of prediction-evasive adversarial examples \citep{papernot2016practical,madry17,chen2017zoo}.  

Although many ZO algorithms have recently been developed and analyzed \citep{liu2017zeroth,flaxman2005online,shamir2013complexity,agarwal2010optimal,nesterov2015random,duchi2015optimal,shamir2017optimal,gao2014information,dvurechensky2018accelerated,wangdu18}, they often suffer \textcolor{black}{from} the high variances of ZO gradient estimates, and in turn, hampered 
%\textcolor{BK_color}{slow} 
convergence rates.  In addition, these algorithms are mainly designed for convex settings, which limits their applicability in a wide range of \textcolor{black}{(non-convex) machine learning problems}.  

In this paper, we study the problem of design and analysis of variance reduced and faster converging nonconvex ZO optimization methods. 
%To reduce the variance of ZO gradient estimates, the first thought is to retrieve similar ideas from the first-order regime.  
To reduce the variance of ZO gradient estimates, one can draw motivations from similar ideas in the first-order regime. The stochastic variance reduced gradient (SVRG)  is a commonly-used, effective first-order approach to reduce the variance \citep{johnson2013accelerating,reddi2016stochastic,nitanda2016accelerated,allen2016improved,lei2017non}. Due to the variance reduction, it improves the convergence rate of stochastic gradient descent (SGD) from $O({1}/{\sqrt{T}})$\footnote{In the big $O$ notation,  the constant numbers are ignored, and the dominant factors are kept.} to $O({1}/{{T}})$, where  $T$ is the total number of iterations.  

Although SVRG has shown a great promise, applying similar ideas to ZO optimization is not a trivial task.   The main challenge arises due to the fact that   SVRG relies upon the assumption that a \textit{stochastic} gradient is an \textit{unbiased} estimate of the \textit{true} batch/full gradient, which unfortunately does \textit{not} hold in the ZO case.  
Therefore, it is an open question  whether  the  ZO stochastic variance reduced gradient \textcolor{black}{could enable}   faster convergence of ZO algorithms.  
In this paper, 
%we will answer this question.
%take some first steps to answer this question.
we attempt to fill the gap between ZO optimization and SVRG.

%In this paper,  
\textbf{Contributions} We propose and evaluate a novel {ZO algorithm for  nonconvex stochastic optimization}, ZO-SVRG, which integrates SVRG with ZO gradient estimators.
{We show that compared to SVRG, 
ZO-SVRG achieves a similar convergence rate that 
decays linearly with $O(1/T)$}
%We show that compared to SVRG, ZO-SVRG achieves a similar convergence rate, 
%but with an additional correction %\textcolor{BK_color}{/error} 
{but up to an additional error correction} 
term of order $1/b$, where $b$ is the mini-batch size. {Without a careful treatment, this correction term (e.g., when $b$ is small)  could be a critical factor affecting the optimization performance.}
%To mitigate this correction term, we propose two accelerated ZO-SVRG  variants, utilizing reduced variance gradient estimators.
{To mitigate this error term, we propose two accelerated ZO-SVRG  variants, utilizing reduced variance gradient estimators.}
These yield a faster convergence rate towards $O(d/T)$, the best  known iteration complexity bound for ZO stochastic optimization. 
%(e.g., the ZO gradient descent method for deterministic programming).
%the ordinary random gradient estimator.  
%Both of them trade-off more on the number of queries of the function value. 

{Our work offers a comprehensive study on how    ZO gradient estimators affect SVRG on both   iteration complexity (i.e., convergence rate)  and       function query complexity.
Compared to the existing ZO algorithms, our methods can 
strike a balance  between  iteration complexity  and function query complexity.}  
To demonstrate the flexibility of our approach in managing this trade-off, we conduct an empirical evaluation of 
%Further, 
%we compare 
our proposed algorithms and other state-of-the-art algorithms
 on two diverse applications: black-box chemical material classification and generation of universal adversarial perturbations from black-box deep neural network models.
Extensive experimental results and theoretical analysis validate the effectiveness of our approaches. 

\vspace*{-0.1in}
\section{Related work}
\vspace*{-0.1in}
 In ZO algorithms, a full gradient is typically approximated using either a one-point or a two-point gradient estimator, where the former  acquires a gradient estimate $\hat{\nabla} f(\mathbf x)$ by  querying $f (\cdot)$ at a single random location close to $\mathbf x$ \citep{flaxman2005online,shamir2013complexity}, and the latter computes a finite difference using two random function queries \citep{agarwal2010optimal,nesterov2015random}.
 %/citep{lian2016comprehensive,liu2017zeroth,agarwal2010optimal,nesterov2015random,duchi2015optimal,dvurechensky2018accelerated,wangdu18, ghadimi2013stochastic,hajinezhadzenith,gu2016zeroth}.
 In this paper, we focus on the \textit{two-point} gradient estimator     since it has a lower variance  and thus improves the complexity bounds of ZO algorithms. %\citep{lian2016comprehensive,liu2017zeroth,agarwal2010optimal,nesterov2015random,duchi2015optimal,dvurechensky2018accelerated,wangdu18, ghadimi2013stochastic,hajinezhadzenith,gu2016zeroth}.  In this paper,
 %we focus on \textit{two-point} based ZO algorithms in this paper.

  Despite the meteoric rise of two-point based ZO algorithms, most of the work is restricted to convex problems \citep{liu2017zeroth,duchi2015optimal,shamir2017optimal,gao2014information,dvurechensky2018accelerated,wangdu18}.
 %The last decade has seen   significant progress for the development of ZO \textit{convex} optimization algorithms. 
 For example, a ZO mirror descent  algorithm proposed by \citep{duchi2015optimal} has an exact rate $O({\sqrt{d}}/{\sqrt{T}})$, where $d$ is the number of optimization variables. The same rate is  obtained by bandit convex optimization  \citep{shamir2017optimal}
 and ZO online alternating direction method of multipliers  \citep{liu2017zeroth}. Current studies suggested  that ZO algorithms typically agree with the iteration complexity of first-order  algorithms up to a small-degree polynomial of   the problem size $d$.

In contrast to the convex setting, non-convex ZO algorithms are comparatively under-studied except  a  few recent attempts   \citep{lian2016comprehensive,nesterov2015random,ghadimi2013stochastic,hajinezhad2017zeroth,gu2016zeroth}. Different from convex optimization, the stationary condition is used to measure
the convergence   of nonconvex methods.
In \citep{nesterov2015random},  the ZO gradient descent (ZO-GD) algorithm was   proposed for \textit{deterministic} nonconvex programming, which yields    $O({d}/{T})$ convergence rate.
%in \citep{nesterov2015random}, the  convergence rate $O({d}/{T})$  of ZO gradient descent (ZO-GD) was established for deterministic nonconvex programming.
A stochastic version of ZO-GD (namely, ZO-SGD) studied in  \citep{ghadimi2013stochastic} achieves
the rate of $O({\sqrt{d}}/{\sqrt{T}})$.
In \citep{hajinezhad2017zeroth}, a ZO distributed  algorithm  %(known as ZONE-M)
was developed for multi-agent optimization, leading to   $O ({1}/{T} + {d}/{q})$ convergence rate. Here  $q$ is  the number of random directions used to construct a  gradient estimate.
In \citep{lian2016comprehensive}, an asynchronous ZO stochastic coordinate descent (ZO-SCD) was derived for parallel optimization and  achieved the   rate of $O  ( {\sqrt{d}}/{\sqrt{T}}  )$. In  \citep{gu2016zeroth},  a variant of ZO-SCD, known as   ZO stochastic variance reduced coordinate (ZO-SVRC) descent, improved the convergence  rate from $O  ( {\sqrt{d}}/{\sqrt{T}}  )$  to $O  ( {{d}}/{{T}}  )$ under the same parameter setting for the gradient estimation.
Although the authors in \citep{gu2016zeroth} considered the  stochastic variance reduced technique, only a coordinate descent algorithm using a coordinate-wise (deterministic) gradient estimator was studied. 
This motivates our  study on a more general framework ZO-SVRG under different   gradient estimators.

\vspace*{-0.1in}
\section{Preliminaries}
\label{sec: preliminaries}
\vspace*{-0.1in}
Consider a nonconvex finite-sum problem of the form
\begin{equation}
    \small
\begin{array}{ll}
\displaystyle \minimize_{\mathbf x \in \mathbb R^d} & \displaystyle f(\mathbf x) \Def  \frac{1}{n}\sum_{i=1}^n
f_i(\mathbf x),
\end{array}\label{eq: prob_ori}
\end{equation}
%where $\mathbf x \in \mathbb R^d$ is the optimization varaible, and there exist 
%$n$ component functions $\{ f_i(\mathbf x) \}$, each of which could be nonconvex.
where  $\{ f_i(\mathbf x) \}_{i=1}^n$ are $n$   individual nonconvex cost functions.
%$\mathbf x \in \mathbb R^d$ is {\color{black}high-dimensional optimization variable, and there exist $n$ potentially nonconvex functions $\{ f_i(\mathbf x) \}_{i=1}^n$.}
The generic form \eqref{eq: prob_ori}
encompasses many machine learning problems, ranging from generalized linear models to   neural networks.
We next elaborate on assumptions of   problem \eqref{eq: prob_ori}, and provide a background on ZO gradient estimators. 

\vspace*{-0.1in}
\subsection{Assumptions}
\vspace*{-0.1in}
%First, the smoothness of individual functions $\{ f_i \}_{i=1}^n$ will be made.

\textbf{A1:} Functions $\{f_i\}$ have
$L$-Lipschitz continuous gradients ($L$-smooth), i.e.,
$\| \nabla f_i (\mathbf x)  - \nabla f_i (\mathbf y)  \|_2 \leq L \| \mathbf x - \mathbf y \|_2$ for any $\mathbf x$ and $\mathbf y$,  $i \in [n]$, and   some $L < \infty$. Here 
$\| \cdot \|_2$ denotes the Euclidean norm, and
for ease of notation $[n]$ represents the integer set $\{1,2,\ldots, n \}$.

\textbf{A2:} The variance of stochastic gradients is bounded as 
$   \frac{1}{n} \sum_{i=1}^n \| \nabla f_i(\mathbf x) - \nabla f(\mathbf x)\|_2^2 \leq \sigma^2$. Here $\nabla f_i(\mathbf x)$ can be viewed as a stochastic gradient of $\nabla f(\mathbf x)$ by randomly picking an index $i \in [n]$. 
 
  Both A1 and A2  are the standard assumptions used in    nonconvex optimization literature \citep{lian2016comprehensive,nesterov2015random,lei2017non,ghadimi2013stochastic,hajinezhad2017zeroth,gu2016zeroth}. 
Note that A2 is milder than the assumption of bounded gradients \citep{liu2017zeroth,hajinezhad2017zeroth}. For example, if   $\| \nabla f_i(\mathbf x)\|_2 \leq \tilde{\sigma}$, then   A2 is satisfied with $\sigma = 2 \tilde{\sigma}$.

\vspace*{-0.05in}
\subsection{ZO gradient estimation}
\vspace*{-0.1in}
% To avoid explicit gradient calculations, 
% random gradient estimator based on two function queries become popular  \citep{gao2014information,liu2017zeroth,nesterov2015random,ghadimi2013stochastic,duchi2015optimal,shamir2017optimal,hajinezhad2017zeroth}. 
Given an individual cost function $f_i$ (or an arbitrary function under A1 and A2), a two-point \underline{rand}om \underline{grad}ient \underline{est}imator   $\hat{\nabla}f_i (\mathbf x)$ is defined by \citep{nesterov2015random, gao2014information}
\begin{equation}
\small
 \hat{\nabla}f_i  (\mathbf x ) =
   (d/\mu)\left  [  f_i ( \mathbf x + \mu \mathbf u_i ) - f_i ( \mathbf x  ) \right ]  \mathbf u_i,  ~ \text{for}~i\in[n],
    \tag{RandGradEst}
    \label{eq: grad_rand}
\end{equation}
where recall that $d$ is the number of optimization variables, $\mu> 0$ is a smoothing  parameter\footnote{The parameter $\mu$  can be generalized to $\mu_i$ for $i \in [n]$. Here  we assume $\mu_i = \mu$ for ease of representation.}, and  $\{ \mathbf u_i \}$ are i.i.d.  random directions drawn from
%Here 
%The distribution of $\mathbf u$ is often assumed to be either a standard Gaussian distribution \citep{nesterov2015random} or
a uniform distribution over a unit sphere   \citep{flaxman2005online,shamir2017optimal,gao2014information}.
%The latter is used in this paper, since the uniform distribution is defined in a {bounded} space  rather than the whole real space required for Gaussian.
%And it can provide 
%a sharper bound on the second moment of the random gradient estimator \eqref{eq: grad_rand} \cite{gu2016zeroth}. 
In general, %$ \hat{\nabla}f  (\mathbf x ) $ 
\ref{eq: grad_rand}
is a biased approximation to   the true gradient ${\nabla}f_i  (\mathbf x ) $, and its bias reduces as $\mu$ approaches zero. However, in a practical system, % a large value of $\mu$ is often preferred. That is because 
if $\mu$ is too small, then the function difference could be dominated by the system noise and   fails to represent the function differential \citep{lian2016comprehensive}.

%since the function difference would be dominated by noise and does not really reflect the function
% differential.

% a large μ to obtain a less exact estimation
% for the stochastic gradient without affecting the convergence rate. From the practical view of
% point, it always tends to choose a large value for μ. Recall the zeroth-order method uses the
% function difference at two different points (e.g., x + μei and x − μei) to estimate the differential.
% In a practical system (e.g., a concrete control system), there usually exists some system noise
% while querying the function values. If two points are too close (in other words μ is too small),
% the obtained function difference is dominated by noise and does not really reflect the function
% differential.

%e.g, an intuitive 1-dimensional case ($d = 1$). 
%however, the extremely small $\beta$ makes the convergence of ZO algorithms much slow. 
%We elaborate on  properties of \eqref{eq: grad_rand} in  Supplementary  Material.

%The rationale behind \eqref{eq: grad_rand} is that  $ \hat{\nabla}f  (\mathbf x ) $ becomes an   unbiased estimator of ${\nabla}f  (\mathbf x ) $ when  $\beta$ approaches zero, e.g, an intuitive 1-dimensional case ($d = 1$).  %We elaborate on more properties of \eqref{eq: grad_rand} in  Supplementary  Material.
\vspace*{-0.05in}
\begin{myremark}
Instead of  using a single sample $\mathbf u_i$   in \ref{eq: grad_rand}, the average of $q$ i.i.d. %sub-samples
samples
 $\{ \mathbf u_{i,j} \}_{j=1}^q$ can also be used for gradient estimation % leading to an \underline{ave}rage \underline{rand}om \underline{grad}ient \underline{est}imator %(Ave-RnadGradEst) 
 %due to resulting reduced variance 
 \citep{liu2017zeroth,duchi2015optimal,hajinezhad2017zeroth},
 \begin{equation}\label{eq: grad_rand_ave}
 \small
\textstyle \hat \nabla f_i(\mathbf x) = (d/(\mu q)) \sum_{j=1}^q 
 \left [  f_i ( \mathbf x + \mu \mathbf u_{i,j} ) - f_i ( \mathbf x  ) \right ]\mathbf u_{i,j},~ \text{for}~i\in [n],
\tag{Avg-RandGradEst}
\end{equation}
which we call an \underline{ave}rage \underline{rand}om \underline{grad}ient \underline{est}imator. %(Ave-RnadGradEst)
\end{myremark}

In addition to \ref{eq: grad_rand} and \ref{eq: grad_rand_ave}, the work \cite{lian2016comprehensive,gu2016zeroth,  choromanski2017blackbox} considered a  \underline{coord}inate-wise  \underline{grad}ient \underline{est}imator. Here every partial derivative is estimated via the two-point querying scheme under fixed direction vectors,
\begin{equation}\label{eq: grad_e}
\small
   \textstyle \hat{\nabla}f_i (\mathbf x ) = \sum_{\ell = 1}^d \left ( 1/(2\mu_\ell) \right )
    \left [ f_i ( \mathbf x + \mu_{\ell} \mathbf e_\ell ) - f_i( \mathbf x - \mu_{\ell} \mathbf e_\ell ) \right ] \mathbf e_l,~ \text{for}~i\in [n],
    \tag{CoordGradEst}
\end{equation}
where $\mu_\ell > 0$ is a  coordinate-wise smoothing parameter, and $\mathbf e_\ell \in \mathbb R^d$ is a standard basis vector with $1$ at its $\ell$th coordinate, and $0$s elsewhere. Compared to \ref{eq: grad_rand}, 
\ref{eq: grad_e} is \textit{deterministic} and requires $d$ times more 
function queries. 
%, leading to a poor query complexity. 
However, as will be evident later, it yields  an improved iteration complexity (i.e., convergence rate).
More details on  ZO  gradient estimation can be found in Appendix\,\ref{supp: ZOgrad}. %(\hyperref[sec: supp]{Supplementary Material}). 
%and will investigate their impacts on our proposed ZO algorithm in the following sections.
%by Lemma\,\ref{lemma: smooth_f_random} and Lemma\,\ref{lemma: deter_smooth_gradest} in Supplementary Material.
%A detailed comparison between  random and deterministic gradient estimators  can be found in Supplementary Material.

%\subsection{Stochastic variance reduced gradient (SVRG) method}
\vspace*{-0.1in}
\section{ZO stochastic variance reduced gradient (ZO-SVRG)}
\label{sec: ZOSVRG}
\vspace*{-0.05in}
%In this section, we present ZO-SVRG algorithm and provide its convergence analysis. 

\subsection{SVRG: from first-order to zeroth-order %{\color{red} change}
}
\vspace*{-0.05in}
% The  stochastic variance reduced gradient (SVRG)
% method blends GD and SGD, where the former is used to control the effect
% of the variance of the latter
It has been shown in
\citep{johnson2013accelerating,reddi2016stochastic} that the first-order SVRG achieves the convergence rate $O({1}/{T})$, yielding  $O(\sqrt{T})$ less iterations than the ordinary SGD for solving   finite sum problems. 
%\textcolor{blue}{(nonvex setting?)}
%, where $T$ is the total number of iterations. %We review SVRG in Algorithm\,1.
The key step of SVRG\footnote{Different from the standard SVRG  \citep{johnson2013accelerating}, we consider its mini-batch variant in \citep{reddi2016stochastic}.} (Algorithm\,1) 
%\footnote{Different from the standard SVRG algorithm in \citep{johnson2013accelerating}, we consider a mini-batch variant  of SVRG in \citep{reddi2016stochastic}.} (Algorithm\,1) 
is to generate an auxiliary sequence $\hat{\mathbf x}$ at which the full gradient is used as a reference in building a modified   stochastic
gradient estimate
\begin{equation}\label{eq: SVRG_update}
\small
  \textstyle  \hat{\mathbf g} = \nabla f_{\mathcal I} (\mathbf x) - ( \nabla f_{\mathcal I} (\hat{\mathbf x}) - \nabla f(\hat{\mathbf x}) ),~  \nabla f_{\mathcal I} (\mathbf x)  = (1/b) \sum_{i \in {\mathcal I}}     {\nabla}f_i ({\mathbf x} )
\end{equation}
where $\hat{\mathbf g}$ denotes the gradient estimate at $\mathbf x$,  
$\mathcal I \subseteq [n]$ is a mini-batch of size $b$ (chosen uniformly randomly\footnote{\textcolor{black}{For mini-batch $\mathcal I$,  SVRG \cite{reddi2016stochastic} assumes i.i.d. samples with replacement, while  a variant of SVRG (called SCSG) assumes samples without replacement \cite{lei2017non}. This paper considers both sampling strategies.}}),
and
$
 \nabla f  (\mathbf x)  = \nabla f_{[n]} (\mathbf x)
 $.
 \textcolor{black}{The key property of \eqref{eq: SVRG_update} is that $\hat {\mathbf g}$ is an unbiased gradient estimate of $\nabla f(\mathbf x)$.}
{The gradient blending \eqref{eq: SVRG_update} is also motivated by a variance reduced  technique known as control variate \citep{tucker2017rebar,grathwohl2017backpropagation,chatterji2018theory}. The link between SVRG and control variate is discussed in Appendix\,\ref{supp: control_var}. %(\hyperref[sec: supp]{Supplementary Material}).
}
 
%  is similar to the technique of control variate for variance reduction \citep{tucker2017rebar,grathwohl2017backpropagation,chatterji2018theory}. If we view $\hat {\mathbf g}_0 \Def \nabla f_{\mathcal I} (\mathbf x)$ as the raw gradient estimate at $\mathbf x$, and    $\mathbf c \Def \nabla f_{\mathcal I}(\hat{\mathbf x})$ as a control variate satisfying 
% $
%  \mathbb E[\mathbf c]
% = \nabla f(\hat{\mathbf x})
% $, then the gradient blending \eqref{eq: SVRG_update} becomes a   gradient estimate modified by a  control variate,
% $
% \hat{\mathbf g} = \hat {\mathbf g}_0 - (\mathbf c - \mathbb E[\mathbf c])
% $. 
% Here $\hat{\mathbf g}$  has the same expectation as $\hat {\mathbf g}_0$, $\mathbb E[\hat{\mathbf g}] = \mathbb E[ \hat {\mathbf g}_0 ] = \nabla f(\mathbf x)$, but has a lower variance when $\mathbf c$ is positively
% correlated with $\mathbf g_0$ \citep{tucker2017rebar,grathwohl2017backpropagation,chatterji2018theory};
% see more details on control variate in 
% Sec.\,\ref{supp: control_var} (\hyperref[sec: supp]{Supplementary Material}).

In the ZO setting, the gradient blending \eqref{eq: SVRG_update} is   approximated  using only  function values,
\begin{equation}\label{eq: ZOSVRG_update}
  \small
  \textstyle  \hat{\mathbf g} = \hat {\nabla} f_{\mathcal I} (\mathbf x) - ( \hat {\nabla} f_{\mathcal I} (\hat{\mathbf x}) - \hat {\nabla} f(\hat{\mathbf x}) ),~ \hat{\nabla} f_{\mathcal I} (\mathbf x)  = (1/b)  \sum_{i \in {\mathcal I}}    \hat {\nabla}f_i ({\mathbf x} ),
\end{equation}
where 
$\hat{\nabla} f (\mathbf x) = \hat{\nabla} f_{[n]} (\mathbf x) 
$,
% $
%  \hat{\nabla} f_{\mathcal I} (\mathbf x)  = (1/b) \sum_{i \in {\mathcal I}}    \hat {\nabla}f_i ({\mathbf x} )
%  $, $
%  \hat{\nabla} f (\mathbf x)  = (1/n) \sum_{i =1}^n    \hat {\nabla}f_i ({\mathbf x} )
%  $, 
and $\hat {\nabla} f_i$ is a ZO gradient estimate   specified by \ref{eq: grad_rand}, \ref{eq: grad_rand_ave} or \ref{eq: grad_e}. 
%and  random direction samples   used for \ref{eq: grad_rand} and \ref{eq: grad_rand_ave} are i.i.d. for each of $\{ \hat {\nabla} f_i\}_{i=1}^n$. 
Replacing \eqref{eq: SVRG_update} with \eqref{eq: ZOSVRG_update} in SVRG (Algorithm\,1) leads to a  new ZO algorithm, which we call ZO-SVRG (Algorithm\,2).
 We highlight that although
 ZO-SVRG  is similar to SVRG  except the use of ZO gradient estimators to estimate batch, mini-batch, as well as blended gradients,  this seemingly minor difference yields an essential difficulty in the analysis of ZO-SVRG. That is, the unbiased assumption on   gradient estimates used in  SVRG no longer holds. Thus,  a careful analysis of ZO-SVRG is much needed.
% there exist gaps (estimation errors) from the ZO gradient estimates to the true gradient.

% The combination of \eqref{eq: ZOSVRG_update} and SVRG leads to a new algorithm, ZO-SVRG (Algorithm\,2).
% %a new algorithm: zeroth-order SVRG (ZO-SVRG).
% However, different from \eqref{eq: SVRG_update}, $\hat{\mathbf g}$ in \eqref{eq: ZOSVRG_update} is no longer an unbiased estimate of $\nabla f(\mathbf x)$.
%  \textcolor{blue}{Therefore, understanding and applying ZO-SVRG requires a careful analysis, including answering the following fundamental questions} : a) What is the impact of ZO gradient estimates on the variance reduced step? b) What is its convergence  improvement/degradation  compared to ZO-SGD and SVRG? 
% %c) What is the degradation in convergence rate compared to SVRG?
% c) What is the function query complexity?  
% We will answer these questions in the rest of the paper.
 
\vspace*{-0.1in}
\subsection{ZO-SVRG and convergence analysis}
\vspace*{-0.05in}
% %We show the pseudo-code of ZO-SVRG in Algorithm\,2. 
%  We note that \textcolor{blue}{although}
%  ZO-SVRG (Algorithm\,2) is similar to SVRG (Algorithm\,1) except the use of ZO gradient estimators to estimate batch, mini-batch, as well as blended gradients,  this \textcolor{blue}{seemingly minor} difference yields an essential difficulty in the analysis of ZO-SVRG: there exist  gaps (in terms of bias and variance) from the ZO gradient blending $\hat {\mathbf g}$
% %to the ZO full gradient estimate $\hat{\nabla }f$, and then
% to the original gradient ${\nabla }f$ {\color{red} not clear}. 
% %By contrast,  the gradient unbiasedness (with respect to $\nabla f$) is a prerequisite  to analyze the first-order variants of SVRG \citep{reddi2016stochastic,lei2017non}.
% %In what follows, we focus on ZO-SVRG under the random gradient estimator \eqref{eq: grad_rand}. 
% %The studies on \eqref{eq: grad_rand_ave} and \eqref{eq: grad_e}  will be elaborated on later.
 In what follows, we focus on the analysis of ZO-SVRG using \ref{eq: grad_rand}. 
Later, we will study ZO-SVRG with \ref{eq: grad_rand_ave} and   \ref{eq: grad_e}.
%Towards  a comprehensive convergence analysis,  
We start by investigating the second-order moment of the blended ZO gradient estimate  $\hat{\mathbf v}_{k}^s$ in the form of \eqref{eq: ZOSVRG_update}; see  Proposition\,\ref{prop: vks}. 
%Here we recall that the superscript $s$ denotes the epoch index, and the subscript $k$ represents the iteration within an epoch.

\begin{figure*}
 \begin{subfigure}{.47\textwidth}
\begin{algorithm}[H]
  \caption*{\textbf{Algorithm\,1}: $\text{SVRG}(T,m,\{\eta_k \}, b,  \tilde {\mathbf x}_0)$}
  \begin{algorithmic}[1]
     \State \textbf{Input}:  total number of iterations $T$, epoch length $m$,  number of epochs $S = \ceil{T/m}$,  step sizes $\{ \eta_k \}_{k=0}^{m-1}$, mini-batch  $b$,  and  $ \tilde {\mathbf x}_0$.
      \For{$s =  1,2,\ldots, S$}
      \State set  $ {\mathbf g}_s =  \nabla f(\tilde{\mathbf x}_{s-1})$,    $\mathbf x_0^{s} = \tilde {\mathbf x}_{s-1}$,
       \For{$k=  0,1,\ldots, m-1$}
 %  \State uniformly randomly pick $i \in [n]$,
  \State choose mini-batch ${\mathcal I}_{k}$ of size $b$,
   \State compute gradient blending via \eqref{eq: SVRG_update}:
    \hspace*{0.39in}
$
\mathbf v_{k}^{s} = {\nabla} f_{\mathcal I_k} ( {\mathbf x}_{k}^{s}) - {\nabla} f_{\mathcal I_k} ( {\mathbf x}_{0}^{s}) +  {\mathbf g}_s $,
\State update $\mathbf x_{k+1}^{s} = \mathbf  x_{k}^{s} - \eta_{k} \mathbf v_{k}^{s}$,
   \EndFor
    \State set $\tilde{\mathbf x}_{s} = \mathbf x_{m}^{s}$,
      \EndFor
       \State \textbf{return}  $\bar{\mathbf x}$ chosen uniformly random from $\{ \{ {\mathbf x}_k^s \}_{k=0}^{m-1} \}_{s=1}^S$.
  \end{algorithmic}
\end{algorithm}
 \end{subfigure}
  \hspace*{0.03in}
 \begin{subfigure}{.5\textwidth}
\begin{algorithm}[H]
  \caption*{\textbf{Algorithm\,2}: $\text{ZO-SVRG}(T,m,\{\eta_k \}, b,  \tilde {\mathbf x}_0, \mu)$}
  \begin{algorithmic}[1]
   \State \textbf{Input}:  In addition to parameters in SVRG, set smoothing parameter $\mu>0$.
   \For{$s =  1,2,\ldots, S$}
   \State \textit{compute ZO estimate $\hat {\mathbf g}_s = \hat \nabla f(\tilde{\mathbf x}_{s-1})$,} % via
  % \hspace*{0.18in}
   %\ref{eq: grad_rand} or  \ref{eq: grad_e},
   \State set $\mathbf x_0^{s} = \tilde {\mathbf x}_{s-1}$,
   \For{$k=  0,1,\ldots, m-1$}
   \State choose mini-batch ${\mathcal I}_{k}$ of size $b$,
   %a  mini-batch  ${\mathcal I}_{k} \in [n]$ \hspace*{0.36in} of   size $b$ (uniformly random with \hspace*{0.39in} replacement),
   \State \textit{compute ZO gradient blending \eqref{eq: ZOSVRG_update}:}
    \hspace*{0.39in}
$
\hat {\mathbf v}_{k}^{s} = \hat{\nabla} f_{{\mathcal I}_{k}} ( {\mathbf x}_{k}^{s}) - \hat{\nabla} f_{{\mathcal I}_{k}} ( {\mathbf x}_{0}^{s}) + \hat {\mathbf g}_s $,
\State update $\mathbf x_{k+1}^{s} = \mathbf  x_{k}^{s} - \eta_{k} \hat {\mathbf v}_{k}^{s}$,
   \EndFor
   \State set $\tilde{\mathbf x}_{s} = \mathbf x_{m}^{s}$,
   \EndFor
      \State \textbf{return}  
      $\bar{\mathbf x}$ chosen uniformly random from $\{ \{ {\mathbf x}_k^s \}_{k=0}^{m-1} \}_{s=1}^S$.
  \end{algorithmic}
\end{algorithm}
 \end{subfigure}
   \vspace*{-0.15in}
\end{figure*}

\vspace*{-0.05in}
\begin{myprop}
\label{prop: vks}
Suppose  A2 holds and   \ref{eq: grad_rand} is used in Algorithm\,2. The blended ZO gradient estimate $\hat{\mathbf v}_{k}^s$ in Step\,7 of Algorithm\,2 satisfies
%The blended ZO gradient estimate  $\hat{\mathbf v}_{k}^s$ in ZO-SVRG yields
\begin{equation}
\label{eq: vks_norm_SVRG2}
\small
%\begin{array}{c}
         \mathbb E[ \| \hat{ \mathbf v}_{k}^s \|_2^2]   \hspace*{-0.03in} \leq \hspace*{-0.03in}
  \frac{4(b+18\textcolor{black}{\delta_n})d}{b}  \mathbb E\left [ \| \nabla f  (\mathbf x_k^{s}) \|_2^2  \right ]+ \frac{6(4d+1)L^2 \textcolor{black}{\delta_n}}{b} \mathbb E \left [ \| \mathbf x_k^s - \mathbf x_0^s \|_2^2 \right ] + \frac{  (6\textcolor{black}{\delta_n} +b)  L^2 d^2 \mu^2}{b} + \frac{ 72d \sigma^2 \textcolor{black}{\delta_n}}{b},
%\end{array}
%     \mathbb E[ \| \hat{ \mathbf v}_{k}^s \|_2^2]  \leq &
%   \frac{4(b+18)d}{b}  \mathbb E\left [ \| \nabla f  (\mathbf x_k^{s}) \|_2^2  \right ]+ \frac{6(4d+1)L^2}{b} \mathbb E \left [ \| \mathbf x_k^s - \mathbf x_0^s \|_2^2 \right ] \nonumber \\
%   &+ \frac{  (6 +b)  L^2 d^2 \mu^2}{b} + \frac{ 72d \sigma^2}{b}.
\end{equation}
\textcolor{black}{where $\delta_n = 1$ if the mini-batch contains i.i.d. samples from $[n]$ with replacement, and $\delta_n = I(b < n)$ if samples are randomly selected without replacement. Here $I(b < n)$ is $1$ if $b < n$, and $0$ if $b = n$.}
\end{myprop}
\textbf{Proof}:
See Appendix\,\ref{supp: lemma_vks}. %(\hyperref[sec: supp]{Supplementary Material}). 
\hfill $\square$

Compared to SVRG and its variants \citep{reddi2016stochastic,lei2017non},
the  error bound \eqref{eq: vks_norm_SVRG2}  involves a new error term $ O( { d \sigma^2}/{b})$ \textcolor{black}{for $b < n$}, which is induced by the second-order moment
%while upper bounding 
 of \ref{eq: grad_rand} (Appendix\,\ref{supp: ZOgrad}).
 %(Lemma\,\ref{lemma: smooth_f_random} in \hyperref[sec: supp]{Supplementary Material}).
%For nonconvex optimization \citep{ghadimi2013stochastic,reddi2016stochastic}, we judge the convergence performance of an iterative algorithm by bounding its stationary error 
 %$\mathbb E[ \| \nabla f(\bar{\mathbf x})\|^2 ] $. Spurred by that and
With the aid of Proposition\,\ref{prop: vks}, Theorem\,\ref{thr: ZO_SVRG_rand} provides the convergence rate of ZO-SVRG in terms of an upper bound on
 $\mathbb E[ \| \nabla f(\bar{\mathbf x})\|^2 ] $ at the solution $\bar{\mathbf x}$.

%we  next show  the convergence rate  of ZO-SVRG in Theorem\,\ref{thr: ZO_SVRG_rand}.
%Following \citep{ghadimi2013stochastic,reddi2016stochastic}, 
% the convergence performance of ZO-SVRG is evaluated by bounding the stationary error
% $\mathbb E[ \| \nabla f(\bar{\mathbf x})\|^2 ] $ at the algorithm output $\bar{\mathbf x}$.
 \vspace*{-0.05in}
 \begin{mythr}\label{thr: ZO_SVRG_rand}
 Suppose A1 and A2 hold, and the random gradient estimator \eqref{eq: grad_rand} is used. The output $\bar{\mathbf x}$ of   Algorithm\,2 satisfies
 \begin{equation} \label{eq: grad_norm_SVRG}
 \small
\displaystyle    \mathbb E \left [ \| \nabla f(\bar { \mathbf x }) \|_2^2 \right ] \leq 
\frac{f(\tilde{\mathbf x}_0) - f^*}{T \bar \gamma } + \frac{L \mu^2 }{T \bar \gamma }
+ \frac{S \chi_m }{ T \bar \gamma},
\end{equation}
where $T = Sm$, $f^* = \min_{\mathbf x} f(\mathbf x)$, $\bar{\gamma} = \min_{k \in [m]} \gamma_k$,  $\chi_m = \sum_{k=0}^{m-1} \chi_k$, and  
\begin{align} %\label{eq: paras_SVRG}
&  \begin{array}{c}
     \gamma_k = \frac{1}{2} \left (1-  \frac{c_{k+1} }{\beta_k} \right ) \eta_k
    - \left ( \frac{L}{2} + c_{k+1} \right ) \frac{4db + 72d \textcolor{black}{\delta_n}}{b } \eta_k^2  
\end{array}  \label{eq: coeff_gammak} \\
&  \begin{array}{c}
\chi_k =   \left (1-  \frac{c_{k+1} }{\beta_k} \right ) \frac{\mu^2 d^2 L^2}{4}\eta_k + \left ( \frac{L}{2}  + c_{k+1} \right )\frac{ (6\textcolor{black}{\delta_n}+b) L^2 d^2 \mu^2 + 72d \sigma^2 \textcolor{black}{\delta_n}}{b}\eta_k^2.
\end{array}
%  &   \gamma_k = \frac{1}{2} \left (1-  \frac{c_{k+1} }{\beta_k} \right ) \eta_k
%     - \left ( \frac{L}{2} + c_{k+1} \right ) \frac{4db + 72d }{b } \eta_k^2,  \label{eq: coeff_gammak} \\
% &\chi_k =   \left (1-  \frac{c_{k+1} }{\beta_k} \right ) \frac{\mu^2 d^2 L^2}{4}\eta_k + \left ( \frac{L}{2}  + c_{k+1} \right )\frac{ (6+b) L^2 d^2 \mu^2 + 72d \sigma^2 }{b}\eta_k^2. 
\label{eq: coeff_chik}
\end{align}
In \eqref{eq: coeff_gammak}-\eqref{eq: coeff_chik}, $\beta_k$ is a positive parameter ensuring $\gamma_k > 0$, and
 the coefficients $\{ c_k \}$ are given by
\begin{equation}\label{eq: ck_coeff}
\small
   c_k = \left [1 + \beta_k \eta_k + \frac{6 (4d+1)  L^2 \textcolor{black}{\delta_n} \eta_k^2}{b} \right ] c_{k+1} + \frac{3 (4d+1) L^3 \textcolor{black}{\delta_n} \eta_k^2}{b},
   \quad c_m = 0.
\end{equation}
 \end{mythr}
 \textbf{Proof}: See Appendix\,\ref{supp: thr1}. 
 %(\hyperref[sec: supp]{Supplementary Material}). 
 \hfill $\square$

 Compared to the convergence rate of SVRG as given in \citep[Theorem\,2]{reddi2016stochastic}, Theorem\,\ref{thr: ZO_SVRG_rand} exhibits two additional errors  $
 ({L \mu^2 }/{(T \bar \gamma )})$
and $ ({S \chi_m }/{ (T \bar \gamma)})$ due to the use of ZO gradient estimates.
 Roughly speaking, if we choose the smoothing parameter $\mu$ reasonably small, then 
 the error $ ({L \mu^2 }/{(T \bar \gamma )})$ would reduce, leading to non-dominant effect on the convergence rate of ZO-SVRG. For the term $ ({S \chi_m }/{ (T \bar \gamma)})$, the quantity $\chi_m $ is more involved, relying on   the epoch length $m$, the step size $\eta_k$, the smoothing parameter $\mu$, the mini-batch size $b$, and the number of optimization variables $d$. In order to acquire explicit dependence on these parameters and   to explore deeper insights of convergence, we simplify \eqref{eq: grad_norm_SVRG} for a specific parameter setting, as formalized below.

\vspace*{-0.05in}
\begin{mycor}\label{col: SVRG_simple}
 Suppose we set
\begin{equation}\label{eq: para_specific_SVRG}
\small
 \mu = \frac{1}{ \sqrt{dT} }, \quad \eta_k = \eta= \frac{\rho}{L d},  
\end{equation}
$ \beta_k = \beta = L$, and $m = \ceil{ \frac{d}{31 \rho} } $, 
where
$ 
0 < \rho \leq 1$ is a universal constant that is independent of $b$, $d$, $L$, and $T$. 
%with $\rho \leq %\min \{ \frac{d}{7b\sqrt{m}},
%\frac{d^{\frac{2}{3}} \sqrt{m} }{6d^{\frac{2}{3}} + 104 (db + 6d) } 
%\} 
%$, 
Then Theorem\,\ref{thr: ZO_SVRG_rand} implies 
$\frac{f(\tilde{\mathbf x}_0) - f^*}{T \bar \gamma } \leq  O\left (  \frac{d}{T}\right )$, $\frac{L \mu^2 }{T \bar \gamma } \leq O\left (  \frac{1}{T^2}\right )$, and $ \frac{S \chi_m}{T \bar{\gamma}} \leq O\left ( 
  \frac{d }{T} + \frac{\textcolor{black}{\delta_n}}{b}
\right )$,
% \begin{equation}
% \label{eq: O_terms_ZOSVRG}
% \small
%   \frac{f(\tilde{\mathbf x}_0) - f^*}{T \bar \gamma } \leq  O\left (  \frac{d}{T}\right ),~
%   \frac{L \mu^2 }{T \bar \gamma } \leq O\left (  \frac{1}{T^2}\right ),~
%   \frac{S \chi_m}{T \bar{\gamma}} \leq O\left ( 
%   \frac{d }{T} + \frac{1}{b}
% \right ),
% \end{equation}
which yields
\begin{equation} \label{eq: grad_norm_SVRG_simplify0}
\small
    \mathbb E \left [ \| \nabla f(\bar { \mathbf x }) \|_2^2 \right  ] \leq O\left ( 
  \frac{d }{T} + \frac{\textcolor{black}{\delta_n}}{b}
\right ).
\end{equation}
\end{mycor}
\textbf{Proof}: See Appendix\,\ref{supp: cor1}.
%(\hyperref[sec: supp]{Supplementary Material}).
\hfill $\square$

%It is worth mentioning that the value of the smoothing parameter $ \mu$ in  Corollary\,\ref{col: SVRG_simple} is less restrictive than many other ZO algorithms. 
{\color{black}It is worth mentioning that the condition on the value of smoothing parameter $ \mu$ in  Corollary\,\ref{col: SVRG_simple} is less restrictive than several ZO algorithms\footnote{One exception is ZO-SCD \citep{lian2016comprehensive} (and its variant ZO-SVRC \citep{gu2016zeroth}), where $\mu \leq O(1/\sqrt{T})$.}.}
For example, ZO-SGD  in \citep{ghadimi2013stochastic} required  
$\mu \leq O(d^{-1}T^{-1/2})$, and ZO-ADMM   \citep{liu2017zeroth} and ZO-mirror descent \citep{duchi2015optimal} considered
$\mu_t = O({d^{-1.5} t^{-1}})$. Moreover similar to \cite{liu2017zeroth},  we set the step size $\eta$   linearly scaled  with $1/d$.
Compared to the aforementioned ZO algorithms \citep{liu2017zeroth,duchi2015optimal,ghadimi2013stochastic},  the convergence performance of ZO-SVRG in \eqref{eq: grad_norm_SVRG_simplify0}   has an improved   (linear rather than sub-linear) dependence on $1/T$. However, it suffers an additional error of order $O({\textcolor{black}{\delta_n}}/{b})$ inherited from $({S \chi_m}/{(T \bar{\gamma})})$ in \eqref{eq: grad_norm_SVRG}, which is also a consequence of the last error term in \eqref{eq: vks_norm_SVRG2}.
\textcolor{black}{We recall from the definition of $\delta_n$ in Proposition\,\ref{prop: vks} that if $b < n$ or  samples in the mini-batch  are chosen independently   from $[n]$, then $\delta_n = 1$. The error term is  eliminated only when $\mathcal I_k = [n]$ for any $k$ (i.e., $\delta_n = 0$). In this case, ZO-SVRG (Algorithm\,2) reduces to ZO-GD in \cite{nesterov2015random} since Step\,7 of Algorithm\,2 becomes
$\hat{\mathbf v}_k^s = \hat{\nabla}f(\mathbf x_k^s)$.}
{A recent work \citep[Theorem\,1]{hajinezhad2017zeroth} also identified the possible  side effect $O(1/b)$ for $b < n$ in the context of   ZO nonconvex multi-agent optimization using a method of multipliers. Therefore, a naive combination of \ref{eq: grad_rand} and SVRG could make the algorithm converging to a neighborhood of a stationary point, where the size of neighborhood is controlled by the mini-batch size $b$. 
Our work and   reference \citep{liu2017zeroth}  show that
a large mini-batch indeed reduces the variance of \ref{eq: grad_rand} and  improves the convergence of ZO optimization methods.
% It has been shown in \citep[Corollary\,3 ]{liu2017zeroth} that a large mini-batch indeed reduces the variance of \ref{eq: grad_rand} and  improves the convergence of ZO optimization methods.
Although the tightness of the error bound \eqref{eq: grad_norm_SVRG_simplify0} is not proven, 
%our rate is consistent with that of ZO-GD and ZO-SVRG \citep{nesterov2015random,gu2016zeroth}.
%Thus a better dependence in $d$ (e.g., $\sqrt{d}$) could be achieved for other  parameter settings.
%However, 
we conjecture that the dependence on $T$ and $b$ could be optimal, since the form is consistent with SVRG, and the latter does not rely on the selected parameters in \eqref{eq: para_specific_SVRG}.
}
% an effect of   \ref{eq: grad_rand} also appears in a ZO nonconvex optimization method, called ZONE-M in \citep{hajinezhadzenith}.

% This demonstrates the significant effect of ZO gradient estimators on the convergence performance, yielding the essential difference with SVRG (Algorithm\,1). 

% It is clear from \eqref{eq: grad_norm_SVRG_simplify0} that a large mini-batch size $b$ reduces the convergence error of ZO-SVRG until  $O({d}/{T})$ dominates the error bound.
% We   remark that the convergence rate of ZO-SVRG   in Corollary\,\ref{col: SVRG_simple} may not be tight. Thus a better dependence in $d$ (e.g., $\sqrt{d}$) could be achieved for other  parameter settings.
% However, we conjecture that the dependence on $T$ and $b$ will not be improved, since the form is consistent with SVRG, and the latter does not rely on the selected parameters in \eqref{eq: para_specific_SVRG}.

\vspace*{-0.1in}
\section{Acceleration of ZO-SVRG}
\label{sec: acceleration}
\vspace*{-0.1in}

In this section, we improve the iteration complexity of ZO-SVRG (Algorithm\,2) by using  \ref{eq: grad_rand_ave} and  \ref{eq: grad_e}, respectively. 
We start by comparing the squared errors of different gradient estimates %\eqref{eq: grad_rand}-\eqref{eq: grad_e} 
to  the true gradient $\nabla f$, as  formalized in Proposition\,\ref{prop: grad_err}. 
%which helps us to  investigate   the convergence performance of ZO-SVRG under \eqref{eq: grad_rand_ave} and \eqref{eq: grad_e}.
%as formalized in Proposition\,\ref{prop: grad_err}.

\vspace*{-0.05in}
\begin{myprop}\label{prop: grad_err}
Consider a gradient estimator $\hat{\nabla}f(\mathbf x) = \nabla f(\mathbf x) + \boldsymbol \omega$, then
the squared error $\mathbb E[\| \boldsymbol \omega \|_2^2]$ 
%for gradient estimators
%\eqref{eq: grad_rand}-\eqref{eq: grad_e}  is given by
\begin{equation}\label{eq: err_specific}
\small
\left \{ 
\begin{array}{ll}
      \mathbb E \left [
    \| \boldsymbol \omega\|_2^2
    \right ] 
    \leq  O \left ( d \right ) \| \nabla f(\mathbf x) \|_2^2  + O \left ( \mu^2  L^2 d^2 \right ) & \text{for \ref{eq: grad_rand}},\\
       \mathbb E \left [
 \|    \boldsymbol{\omega} \|_2^2
    \right ] \leq
    O \left (
    \frac{q+d}{q} \right ) \| \nabla f(\mathbf x) \|_2^2 + O \left ( \mu^2 L^2 d^2
    \right )  & \text{for \ref{eq: grad_rand_ave}}, \\
        \| \boldsymbol{\omega} \|_2^2 \leq O\left( L^2 d \sum_{\ell = 1}^d \mu_\ell^2 \right )  & \text{for \ref{eq: grad_e}}.
\end{array}
\right.
\end{equation}
\end{myprop}
\textbf{Proof}: See 
Appendix\,\ref{supp: grad_err}.
%(\hyperref[sec: supp]{Supplementary Material}). 
\hfill $\square$

%\textcolor{blue}{estimators (2), (3), (4) appear many times. Should we consider give them some names and indicate the names in the equations?}
 
Proposition\,\ref{prop: grad_err} shows that compared to  \ref{eq: grad_e},  
   \ref{eq: grad_rand} and \ref{eq: grad_rand_ave}   involve an additional error term within a factor $O(d)$ and $O({(q+d)}/{q})$ of  
 $   \| \nabla f(\mathbf x) \|_2^2$, respectively.
 %The above characteristic  of ZO random gradient estimators was first identified by \citep{nesterov2015random,duchi2015optimal}. 
Such an   error is introduced by the second-order moment of gradient estimators using random direction samples \citep{nesterov2015random,duchi2015optimal}, and it
decreases as the number of direction samples $q$ increases.
%so that  the slowdown factor proportional to the problem size $d$ is mitigated.  
On the other hand, all gradient estimators  have a common error bounded by $O(\mu^2 L^2 d^2)$, where let $\mu_\ell = \mu$ for $\ell \in [d]$  in \ref{eq: grad_e}. \textit{If   $\mu $ is  specified as in \eqref{eq: para_specific_SVRG}, then we obtain the error term $O({d}/{T})$, consistent with  the convergence rate of ZO-SVRG in Corollary\,\ref{col: SVRG_simple}.}

In Theorem\,\ref{thr: ZO_SVRG_rand_ave}, we  
show the effect of   \ref{eq: grad_rand_ave} on 
 the convergence rate of ZO-SVRG.

\vspace*{-0.05in}
 \begin{mythr}\label{thr: ZO_SVRG_rand_ave}
 Suppose A1 and A2 hold, and   \ref{eq: grad_rand_ave} is used in  Algorithm\,2. Then $\mathbb E \left [ \| \nabla f(\bar { \mathbf x }) \|_2^2 \right ]$ is bounded same as given in \eqref{eq: grad_norm_SVRG},
 %is same bounded  as \eqref{eq: grad_norm_SVRG}, 
 where the parameters $\gamma_k$, $\chi_k$ and $c_k$ for $k \in [m]$ are modified by
 $ \gamma_k =  \frac{1}{2} \left (1-  \frac{c_{k+1} }{\beta_k} \right ) \eta_k    - \left ( \frac{L}{2} + c_{k+1} \right ) \frac{(72\textcolor{black}{\delta_n} + 4b)(q+d) }{b q } \eta_k^2
 $,
 $\chi_k =    \left (1-  \frac{c_{k+1} }{\beta_k} \right ) \frac{\mu^2 d^2 L^2}{4}\eta_k + \left ( \frac{L}{2}  + c_{k+1} \right )\frac{ (6\textcolor{black}{\delta_n}+b)(q+1) L^2 d^2 \mu^2 + 72 (q+d) \sigma^2 \textcolor{black}{\delta_n}}{bq}\eta_k^2
 $,  
 $
 c_k  =  \left [ 1 +  \beta_k \eta_k+ \frac{6(4d+5q)L^2\textcolor{black}{\delta_n}}{bq} \eta_k^2 \right ] c_{k+1} + 
\frac{3(4d+5q)L^3\textcolor{black}{\delta_n}}{bq} \eta_k^2
 $ with $c_m = 0$.
% \begin{equation} %\label{eq: paras_SVRG_ave}
%   \gamma_k = & \frac{1}{2} \left (1-  \frac{c_{k+1} }{\beta_k} \right ) \eta_k
%     - \left ( \frac{L}{2} + c_{k+1} \right ) \frac{(72 + 4b)(q+d) }{b q } \eta_k^2,  \label{eq: coeff_gammak_ave} \\
%  \chi_k = &   \left (1-  \frac{c_{k+1} }{\beta_k} \right ) \frac{\mu^2 d^2 L^2}{4}\eta_k + \left ( \frac{L}{2}  + c_{k+1} \right )\frac{ (6+b)(q+1) L^2 d^2 \mu^2 + 72 (q+d) \sigma^2 }{bq}\eta_k^2 , \label{eq: coeff_chik_ave} \\
%   c_k  = & \left [ 1 +  \beta_k \eta_k+ \frac{6(4d+5q)L^2}{bq} \eta_k^2 \right ] c_{k+1} + 
% \frac{3(4d+5q)L^3}{bq} \eta_k^2, ~ c_m = 0. \label{eq: ck_coeff_ave}
% \end{equation}
%Here $\beta_k$ is a parameter ensuring $\gamma_k > 0$, and $c_m = 0$.
Given the     setting in Corollary\,\ref{col: SVRG_simple} and $m = \ceil{\frac{d}{55 \rho} }$, the convergence rate simplifies to
\begin{equation} \label{eq: grad_norm_SVRG_simplify0_ave}
\small
    \mathbb E\left [ \| \nabla f(\bar { \mathbf x }) \|_2^2 \right ] \leq O\left ( 
  \frac{d}{T} + \frac{\textcolor{black}{\delta_n}}{b \min\{d,q \}}
\right ).
\end{equation}
\end{mythr}
 \textbf{Proof}: See Appendix\,\ref{supp: thr2} %(\hyperref[sec: supp]{Supplementary Material}). 
 \hfill $\square$
 
 By contrast with Corollary\,\ref{col: SVRG_simple}, it can be seen from \eqref{eq: grad_norm_SVRG_simplify0_ave} that the use of   \ref{eq: grad_rand_ave} reduces the error $O({\delta_n}/{b})$   in \eqref{eq: grad_norm_SVRG_simplify0} through multiple ($q$)   direction samples. 
And the   convergence rate ceases to be significantly improved as $q \geq d$. 
Our empirical results show that a moderate choice of $q$ can significantly speed up the convergence of ZO-SVRG.

%\subsection{ZO-SVRG with a coordinate-wise gradient estimator}
We next study the   effect of the coordinate-wise gradient estimator \eqref{eq: grad_e} on 
 the convergence rate of ZO-SVRG, as formalized in  Theorem\,\ref{thr: ZO_SVRG_coord}.
 
 \vspace*{-0.05in}
  \begin{mythr}\label{thr: ZO_SVRG_coord}
 Suppose A1 and A2 hold, and \ref{eq: grad_e} with $\mu_\ell = \mu$ is used in  Algorithm\,2. Then     
 \begin{equation} \label{eq: grad_norm_SVRG_coord}
 \small
    \mathbb E \left [ \| \nabla f(\bar { \mathbf x }) \|_2^2 \right ] \leq 
\frac{f(\tilde{\mathbf x}_0) - f^*}{T \bar \gamma } +  \frac{S \chi_m }{ T \bar \gamma},
\end{equation}
 where $T$, $f^*$, $\bar{\gamma}$ and $\chi_m$ have been defined in \eqref{eq: grad_norm_SVRG},    the parameters $\gamma_k$, $\chi_k$ and $c_k$ for $k \in [m]$ are given by
 %$T = Sm$, $f^* = \min_{\mathbf x} f(\mathbf x)$, $\bar{\gamma} = \min_{k} \gamma_k$,  $\chi_m = \sum_{k=0}^{m-1} \chi_k$, 
$ \gamma_k = 
\frac{1}{2}\left( 1 - \frac{c_{k+1}}{\beta_k} \right ) \eta_k - 4\left(
\frac{L}{2} + c_{k+1}
\right ) \eta_k^2$, 
$
\chi_k =  \left( \frac{1}{4} + \frac{c_{k+1}}{\beta_{k}} \right) \frac{ L^2  \mu^2 d^2}{2} \eta_k
 + \left (\frac{L}{2} + c_{k+1} \right ) \mu^2 L^2 d^2  \eta_k^2
$,
$
c_k = \left ( 1 +   \beta_k \eta_k + \frac{2d L^2 \textcolor{black}{\delta_n} \eta_k^2}{b}  \right )  c_{k+1} + \frac{ d L^3 \textcolor{black}{\delta_n} \eta_k^2}{b}
$ with $c_m = 0$, and  $\beta_k$ is a positive parameter ensuring $\gamma_k > 0$.
% \begin{equation} %\label{eq: paras_SVRG_ave}
% \gamma_k = & 
% \frac{1}{2}\left( 1 - \frac{c_{k+1}}{\beta_k} \right ) \eta_k - 4\left(
% \frac{L}{2} + c_{k+1}
% \right ) \eta_k^2,  \label{eq: coeff_gammak_coord} \\
% \chi_k =  & 
% \left( \frac{1}{4} + \frac{c_{k+1}}{\beta_{k+1}} \right) \frac{ L^2  \mu^2 d^2}{2} \eta_k
% + \left (\frac{L}{2} + c_{k+1} \right ) \mu^2 L^2 d^2  \eta_k^2, \label{eq: coeff_chik_coord} \\
%   c_k =& \left ( 1 +   \beta_k \eta_k + \frac{2d L^2 \eta_k^2}{b}  \right )  c_{k+1} + \frac{ d L^3 \eta_k^2}{b}, ~ c_m = 0. \label{eq: ck_coeff_coord}
% \end{equation}
%Here $\beta_k$ is a positive parameter ensuring $\gamma_k > 0$.
Given the specific   setting in Corollary\,\ref{col: SVRG_simple} and $m = \ceil{\frac{d}{3 \rho} }$, the convergence rate simplifies to
\begin{equation} \label{eq: grad_norm_SVRG_simplify0_coord}
\small
    \mathbb E\left [ \| \nabla f(\bar { \mathbf x }) \|_2^2 \right ] \leq  O\left ( 
  \frac{d}{T} 
\right ).
\end{equation}
\end{mythr}
 \textbf{Proof}: See Appendix\,\ref{supp: thr3}. %(\hyperref[sec: supp]{Supplementary Material}). 
 \hfill $\square$

 %%%% some analysis.
% Compared \eqref{eq: grad_norm_SVRG_simplify0_coord} with \eqref{eq: grad_norm_SVRG_simplify0} and  \eqref{eq: grad_norm_SVRG_simplify0_ave}, 
 Theorem\,\ref{thr: ZO_SVRG_coord} shows that 
 the use of  \ref{eq: grad_e}  improves the iteration complexity, where the   error of order $O({1}/{b})$   in Corollary\,\ref{col: SVRG_simple}  or  $O({1}/{(b\min\{d,q\})})$ in  Theorem\,\ref{thr: ZO_SVRG_rand_ave} has been eliminated in \eqref{eq: grad_norm_SVRG_simplify0_coord}. 
 This improvement  is benefited from the low variance of   \ref{eq: grad_e} shown by Proposition\,\ref{prop: grad_err}. We can also see this benefit by comparing $\chi_k$ in Theorem\,\ref{thr: ZO_SVRG_coord}  with \eqref{eq: coeff_chik}: the former avoids the term  $(d\sigma^2/b)$.
The disadvantage of \ref{eq: grad_e} is the  need of 
$d$ times more function queries than \ref{eq: grad_rand} in gradient estimation.

% We provide a more detailed complexity analysis in what follows.

%\paragraph{Function query complexity}
Recall that  \ref{eq: grad_rand}, \ref{eq: grad_rand_ave} and \ref{eq: grad_e}  require $O(1)$, $O(q)$ and $O(d)$ function queries, respectively. In ZO-SVRG (Algorithm\,2), the total number of gradient evaluations is given by $nS + bT$, where $T = mS$. 
Therefore, by fixing the number of iterations $T$, the function query complexity of ZO-SVRG using the studied estimators   is then given by 
$O(nS + bT)$, $O(q(nS + bT))$ and $O(d(nS + bT))$, respectively. 
In Table\,\ref{table: SZO_complexity_T}, we summarize the convergence rates  and the function query complexities of  ZO-SVRG and its two   variants, which we call ZO-SVRG-Ave and ZO-SVRG-Coord, respectively. 
For comparison, we also present the results of ZO-SGD \citep{ghadimi2013stochastic} and ZO-SVRC \citep{gu2016zeroth}, where the later updates $J$ coordinates per iteration within an epoch. Table\,\ref{table: SZO_complexity_T} shows that ZO-SGD has the lowest query complexity but has the worst convergence rate.  ZO-SVRG-coord yields the best convergence rate   in the cost of high query complexity. By contrast, ZO-SVRG (with an appropriate mini-batch size) and ZO-SVRG-Ave could achieve better trade-offs between the convergence rate  and the query complexity. 
 
% \begin{table}[t]
% \caption{Sample table title}
% \label{sample-table}
% \begin{center}
% \begin{tabular}{ll}
% \multicolumn{1}{c}{\bf PART}  &\multicolumn{1}{c}{\bf DESCRIPTION}
% \\ \hline \\
% Dendrite         &Input terminal \\
% Axon             &Output terminal \\
% Soma             &Cell body (contains cell nucleus) \\
% \end{tabular}
% \end{center}
% \end{table}

 \begin{table}[htb]
   \vspace*{-0.15in}
\centering
\caption{Summary of convergence rate and function query complexity of our proposals given $T$ iterations.}
\label{table: SZO_complexity_T}
\begin{adjustbox}{max width=\textwidth }
\begin{tabular}{|c|c|c|c|c|}
\hline
Method        & \begin{tabular}[c]{@{}c@{}}Grad. estimator\end{tabular}       & Stepsize 
%&     \begin{tabular}[c]{@{}c@{}}Iteration   \end{tabular} 
&  \begin{tabular}[c]{@{}c@{}}Convergence  rate \\ (worst case as $b < n$)  \end{tabular}         & \begin{tabular}[c]{@{}c@{}}Query  complexity \end{tabular} 
\\ \hline  
ZO-SVRG  & \eqref{eq: grad_rand} &  $O\left (\frac{1}{d} \right )$
%&  $T$ 
& $O\left (\frac{d}{T}+\frac{1}{b} \right )$  &
$O\left (nS + bT \right ) $ \\ \hline
ZO-SVRG-Ave  & \eqref{eq: grad_rand_ave} &  $O(\frac{1}{d})$
%&   $T$ 
& $O \left (\frac{d}{T}+\frac{1}{b\min\{ d,q\}} \right )$   &
$O\left (qnS + qbT  \right ) $
%, $q \ll d$ 
\\ \hline
ZO-SVRG-Coord & \eqref{eq: grad_e} &  $O(\frac{1}{d})$  
%&$T$ 
& $O(\frac{d}{T})$   & 
$O(dnS + dbT ) $ \\ \hline
ZO-SGD \citep{ghadimi2013stochastic} & \eqref{eq: grad_rand}   &  $O \left ( \min\{ \frac{1}{d} , \frac{1}{\sqrt{dT}}\} \right )$  
%& $T$ 
& $O\left ( \frac{\sqrt{d}}{\sqrt{T}} \right )$ &  $O(bT)$
\\ \hline 
ZO-SVRC \citep{gu2016zeroth} &
\eqref{eq: grad_e}
&
$O\left ( \frac{1}{n^\alpha} \right )$, $\alpha \in (0,1)$    
%&  $T$ 
&
$O\left ( \frac{  d}{T}\right )$ &
\text{$O \left ( d n S + J b T \right )$}
\\ \hline
\end{tabular}
\end{adjustbox}
  \vspace*{-0.15in}
\end{table}

\vspace*{-0.1in}
\section{Applications and experiments}
\label{sec: experiments}
\vspace*{-0.05in}
We evaluate the performance of our proposed algorithms on two applications: black-box classification and \textcolor{black}{generating adversarial examples from black-box DNNs}. The first application is motivated by a real-world material science problem, where a material is classified to   either  be a conductor or an insulator from a  density function theory (DFT) based black-box simulator  \citep{dft}. The second application arises in \textcolor{black}{testing the robustness of a deployed DNN via iterative model queries}
 \citep{papernot2016practical,chen2017zoo}.
 
 \vspace*{-0.05in}
\paragraph{Black-box binary classification}
\textcolor{black}{
We consider a non-linear least square problem \citep[Sec.\,3.2]{xu2017second}, i.e., problem \eqref{eq: prob_ori} with $f_i(\mathbf x) = \left ( y_i - \phi(\mathbf x; \mathbf a_i) \right )^2$ for $i\in[n]$. Here
   $(\mathbf a_i, y_i)$ is the $i$th data sample containing feature vector $\mathbf a_i \in \mathbb R^d$ and label $y_i \in \{0,1 \}$, and $\phi(\mathbf x; \mathbf a_i) $ is a \textit{black-box} function that only returns the function value   given an input.
The used dataset consists of   $N = 1000$ crystalline materials/compounds extracted from   Open Quantum Materials Database \citep{oqmd}. Each compound has $d = 145$ chemical features, and its label   ($0$ is conductor and $1$ is insulator)  is determined by a DFT simulator~\citep{vasp}. Due to the black-box nature of DFT, the true $\phi$ is unknown\footnote{ 
\textcolor{black}{One can mimic DFT simulator using a logistic function once the parameter $\mathbf x$ is learned from  ZO algorithms.}}.
%One can mimic it using a logistic function once the parameter $\mathbf x$ is learned from  ZO algorithms.
We split the dataset into two equal parts, leading
to $n=500$ training samples and $(N-n)$ testing samples.
%In addition to a real dataset, we also perform the non-linear least square problem over a synthetic dataset.
We refer   readers to   Appendix\,\ref{supp: experiment_classification} 
%(\hyperref[sec: supp]{Supplementary Material}) 
for more details on our dataset and the setting of experiments.
}

   \begin{figure}[htb]
     \vspace*{-0.15in}
\centerline{
\begin{tabular}{cc}
\includegraphics[width=.46\textwidth,height=!]{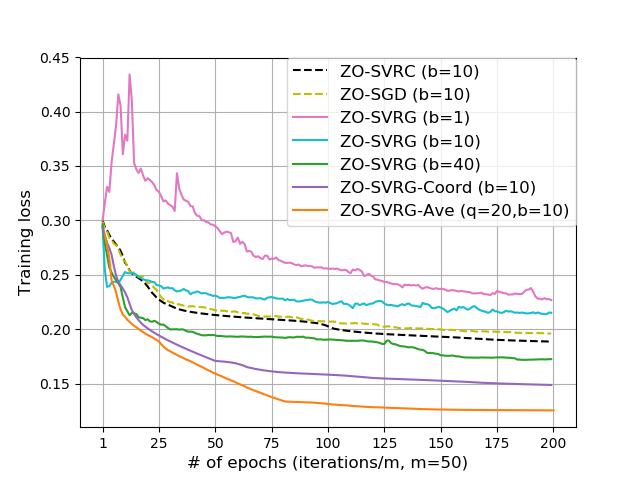}  &
\includegraphics[width=.46\textwidth,height=!]{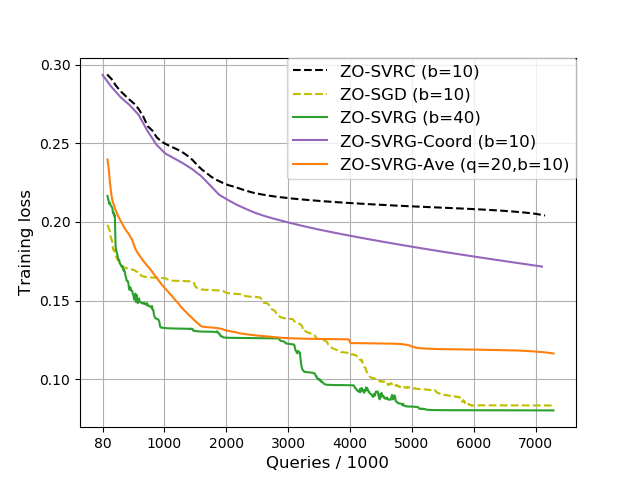}
\\
\footnotesize{(a) Training loss versus  iterations} &   \footnotesize{(b) Training loss versus function queries} 
\end{tabular}}
\caption{\footnotesize{Comparison of different ZO algorithms for  the task of chemical material classification.}}
  \label{fig: DFT}
   \vspace*{-0.05in}
\end{figure}

    \begin{table}[htb]
     \vspace*{-0.15in}
\centering
\caption{Testing error for chemical material classification using $7.3\times 10^6$ function queries. }
\label{table: test_real}
\begin{adjustbox}{max width=\textwidth }
\begin{tabular}{|l|c c c c c|}
\hline
Method & ZO-SGD \citep{ghadimi2013stochastic}  
&  ZO-SVRC \citep{gu2016zeroth}   &ZO-SVRG & ZO-SVRG-Coord & ZO-SVRG-Ave    
\\ \hline  
\# of epochs  &$14600$ & $100$  & $2920$    & $50$ 
&  $365$ 
\\ \hline
Error ($\%$) &$12.56\%$ & $23.70\%$  & $11.18\%$    &  $20.67\%$ & $15.26\%$ 
\\ \hline
\end{tabular}
\end{adjustbox}
  %\vspace*{-0.15in}
\end{table}
     
\textcolor{black}{
 In Fig.\,\ref{fig: DFT}, we present the training loss against   the number of epochs (i.e., iterations divided by the epoch length $m = 50$) and  function  queries. We compare  our proposed algorithms ZO-SVRG, ZO-SVRG-Coord and ZO-SVRG-Ave with   ZO-SGD \citep{ghadimi2013stochastic} and ZO-SVRC \citep{gu2016zeroth}. 
 Fig.\,\ref{fig: DFT}-(a) presents the convergence trajectories of   ZO algorithms as   functions  of the number of epochs, where ZO-SVRG is evaluated under different mini-batch sizes  $b \in \{1,10,40 \}$.
We observe that the convergence error of ZO-SVRG decreases as $b$ increases, and for a small mini-batch size $b \leq 10$, ZO-SVRG likely converges to a neighborhood of a critical point as shown
by  Corollary\,\ref{col: SVRG_simple}.
We also note that our proposed algorithms ZO-SVRG ($b= 40$), ZO-SVRG-Coord and ZO-SVRG-Ave have faster convergence speeds (i.e., less iteration complexity) than the existing algorithms ZO-SGD and ZO-SVRC. Particularly, the use of multiple random direction samples in \ref{eq: grad_rand_ave} significantly accelerates ZO-SVRG since the error of order $O(1/b)$  is reduced to $O(1/(bq))$ (see Table\,\ref{table: SZO_complexity_T}), leading to a non-dominant factor versus $O(d/T)$ in the convergence rate of ZO-SVRG-Ave. Fig.\,\ref{fig: DFT}-(b) presents the training loss against the number of function queries. For the same experiment,  Table\,\ref{table: test_real} shows the number of iterations and the testing error of algorithms studied in Fig.\,\ref{fig: DFT}-(b)
using   $7.3\times 10^6$  function queries.
% We also note from Table\,\ref{table: test_real}
% that the same function query complexity   implies different  iteration numbers for   ZO algorithms studied in Table\,\ref{table: test_real}. 
%\footnote{The first row of Table\,\ref{table: test_real} shows the   number of iterations required to achieve the  number of function queries $7.3\times 10^6$.}. 
We observe that the performance of \ref{eq: grad_e} based algorithms (i.e., ZO-SVRC and ZO-SVRG-Coord) degrade due to the need of large number of function queries to construct  coordinate-wise gradient estimates.
 By contrast, algorithms based on random gradient estimators (i.e., ZO-SGD, ZO-SVRG and ZO-SVRG-Ave) yield better both training and testing results, while ZO-SGD consumes an extremely large number of   iterations ($14600$ epochs). As a result,  ZO-SVRG ($b=40$)  and ZO-SVRG-Ave   achieve better tradeoffs     between the  iteration   and  the function query complexity.  
}

\paragraph{Generation of adversarial examples from black-box DNNs} In image classification, adversarial examples refer to carefully crafted perturbations such that, when added to the natural images, are visually imperceptible but will lead the target model to misclassify.  In the setting of `zeroth order' attacks \citep{madry17,chen2017zoo,carlini2017towards}, the model parameters are hidden and acquiring its gradient is inadmissible. Only the model evaluations are accessible.
We can then regard the task of generating a universal adversarial perturbation (to  $n$ natural images) as an ZO optimization problem of the form \eqref{eq: prob_ori}. We elaborate on the problem formulation %\eqref{eq: prob_ori} 
for generating adversarial examples  in Appendix\,\ref{appendix: adv_app}.

\begin{figure}[t]
\begin{minipage}[t]{\textwidth}
	\centering
	%\vspace{-7.5mm}
	\begin{minipage}[t]{.43\textwidth}
		\includegraphics[width=\textwidth]{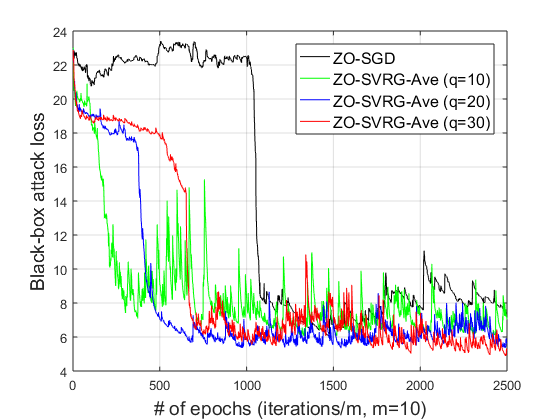}
		%\caption{Training loss versus iterations  for generating universal adversarial perturbations.}
	\end{minipage}
	\hspace{1mm}
	\begin{minipage}[t]{.45\textwidth}
    %\caption{My caption}
    %\label{my-label}
    \vspace{-35mm}
	\centering
\begin{tabular}{l|l}
\hline
Method           & \multicolumn{1}{l}{$\ell_2$ distortion}  \\ \hline
ZO-SGD           & $5.22$                                 \\
ZO-SVRG-Ave ($q=10$) & $4.91$    ($6 \% $)                             \\
ZO-SVRG-Ave ($q=20$) & $3.91$       ($25 \% $)                            \\
ZO-SVRG-Ave ($q=30$) & $3.67$           ($30 \% $)                        \\
%ZO-SVRG-Ave ($q=50$) & \multicolumn{1}{l}{}                 \\ 
\hline
\end{tabular}
    \end{minipage}
    		\caption{Comparison of ZO-SGD and ZO-SVRG-Ave for generation of universal adversarial perturbations from a black-box DNN. Left:  Attack loss versus iterations. Right: $\ell_2$ distortion and improvement $(\%)$ with respect to ZO-SGD.}
    		\label{Fig3}
\end{minipage}
	 \vspace*{-0.2in}
	\end{figure}

We use a well-trained DNN\footnote{\url{https://github.com/carlini/nn_robust_attacks}} on the MNIST handwritten digit classification task as the target black-box model, which achieves 99.4\% test accuracy on natural examples. Two ZO optimization methods, ZO-SGD and ZO-SVRG-Ave, are performed in our experiment. Note that ZO-SVRG-Ave reduces to ZO-SVRG when $q = 1$. We choose $n = 10$ images from the same class, and set the same parameters $b=5$  and constant step size $30/d$ for both ZO methods, where $d=28 \times 28 $ is the image dimension. For ZO-SVRG-Ave, we set $m=10$ and vary the number of random direction samples $q \in \{10,20,30\}$.
In Fig.\,\ref{Fig3}, we show the black-box attack loss (against the number of epochs) as well as the least $\ell_2$ distortion of the successful (universal) adversarial perturbations. 
{To reach the same attack loss (e.g., $7$ in our example), ZO-SVRG-Ave requires roughly $30\times$ ($q=10$), $77 \times$ ($q=20$) and $380\times$ ($q=30$) more function evaluations than ZO-SGD. The sharp drop of attack loss in each method could be caused by
%in each method could be explained by an artifact of
the hinge-like loss as part of the total loss function, which turns to $0$ only if the attack becomes successful.
 Compared to ZO-SGD, ZO-SVRG-Ave 
 %yields two advantages in generation of adversarial examples. First, its convergence is much faster in earlier iterations (e.g., as the number of epochs is less than $1000$). Second, its convergence becomes  more stable and more accurate (at the final iteration) as   $q$ increase due to the smaller variance of  \ref{eq: grad_rand_ave}.
offers a faster convergence to a  more accurate solution, and its convergence trajectory is more stable as $q$ becomes larger (due to the reduced variance of \ref{eq: grad_rand_ave}). 
In addition,
%the $\ell_2$ distortion of ZO-SVRG-Ave with $q=10$ is comparable to that of ZO-SGD. 
%as $q$ increases, 
ZO-SVRG-Ave improves
the $\ell_2$ distortion of adversarial examples compared to ZO-SGD (e.g., $30\%$ improvement when $q = 30$). We present the corresponding adversarial examples    in Appendix\,\ref{appendix: adv_app}. 
%(Supplementary Material).
}

   %It is also worth mentioning that compared to ZO-SVRG-Acc2, ZO-SVRG-Acc1 yields the compelling performance only 

%%%% Fig. 2... Material Science Problem

%Due to the black-box behavior of DFT, the ground truth of $\phi(\mathbf x; \mathbf a_i)$ is not known. 
%Here we choose the logistic function to mimic it while presenting the training loss.

% solving three
% tasks, namely generalized LASSO, robust regression with a low-rank regularizer (RR-LR) and
% learning low-rank representation. For generalized lasso, our experiment focuses on comparing the
% proposed LA-SADMM with SADMM. For the latter tasks, we focus on comparing the proposed
% LA-ADMM with previous linearized ADMM with and without self-adaptive penalty parameters.

\vspace*{-0.05in}
\section{Conclusion}
\label{sec: conclusion}
\vspace*{-0.05in}
In this paper, we   studied ZO-SVRG, a new ZO nonconvex optimization method. We presented 
%its %\textcolor{red}{superior}
new
convergence results beyond
the existing work on ZO nonconvex optimization. We show that ZO-SVRG improves the convergence rate of ZO-SGD from $O(1/\sqrt{T})$ to $O(1/{T})$ %with respect to the number of iterations $T$ 
but suffers a new correction term of order $O(1/b)$.  The is the side effect of combining a two-point random gradient estimators with SVRG. We then propose two accelerated variants of ZO-SVRG based on improved gradient estimators of reduced variances. We show an illuminating trade-off between the iteration and the function query complexity.
  Experimental results and theoretical analysis validate the effectiveness of our approaches compared to other state-of-the-art algorithms.

% \subsubsection*{Acknowledgments}

% Use unnumbered third level headings for the acknowledgments. All
% acknowledgments go at the end of the paper. Do not include
% acknowledgments in the anonymized submission, only in the final paper.
{\small{
\bibliographystyle{IEEEbib}
\bibliography{Ref}
}}

\newpage
\appendix

 \section{Supplementary material}\label{sec: supp}

\subsection{Zeroth-order (ZO) gradient estimators}\label{supp: ZOgrad}
With an abuse of notation, in this section let $f$ be an arbitrary function under assumptions A1 and A2.
Lemma\,\ref{lemma: smooth_f_random} shows the second-order statistics of  \ref{eq: grad_rand}.

\begin{mylemma}
\label{lemma: smooth_f_random}
%\citep[Lemma\,4.1]{gao2014information}
Suppose that Assumption A1 holds, and define $f_\mu = \mathbb E_{\mathbf u \in U_{\mathrm{b}}}[f(\mathbf x + \mu \mathbf u)]$, where $U_{\mathrm{b}}$ is a uniform distribution over the unit Euclidean ball. Then  \ref{eq: grad_rand} yields:

1) $f_\mu$ is $L$-smooth, and
\begin{equation}\label{eq: smooth_est_grad}
    \nabla f_\mu (\mathbf x) 
    %  = \mathbb E_{\mathbf u }
    % \left [
    % \frac{d}{\mu} f(\mathbf x + \mu \mathbf u) \mathbf u
    % \right ] 
    =  \mathbb E_{\mathbf u }
    \left [
    \hat \nabla f(\mathbf x)
    \right ],
\end{equation}
where $\mathbf u$ is drawn from 
the uniform distribution over the unit Euclidean
sphere, and 
$\hat \nabla f(\mathbf x)$ is given by \ref{eq: grad_rand}.

2) For any $\mathbf x \in \mathbb R^d$,
\begin{align}
& | f_\mu(\mathbf x) - f(\mathbf x) | \leq \frac{L \mu^2}{2} \label{eq: dist_f_smooth_true_random}  \\
& \| \nabla f_\mu (\mathbf x) - \nabla f(\mathbf x) \|_2^2 \leq \frac{\mu^2 L^2 d^2}{4}, \label{eq: dist_smooth_true} \\
&  
\frac{1}{2}  \| \nabla f(\mathbf x) \|_2^2 - \frac{\mu^2 L^2 d^2}{4}  \leq \| \nabla f_\mu (\mathbf x)\|_2^2 \leq 2 \| \nabla f(\mathbf x) \|_2^2
    + \frac{\mu^2 L^2 d^2 }{2}. \label{eq: up_grad_fsmooth}
\end{align}

 3) For any $\mathbf x \in \mathbb R^d$,
 \begin{align}
  \mathbb E_{\mathbf u} \left [ \| \hat \nabla f(\mathbf x) - \nabla f_\mu (\mathbf x) \|_2^2 
    \right ] \leq 
 \mathbb E_{\mathbf u} \left [
\| 
\hat \nabla f(\mathbf x)
\|_2^2
\right ] \leq 2 d \| \nabla f(\mathbf x) \|_2^2 + \frac{\mu^2  L^2 d^2}{2}.
\label{eq: second_moment_grad_random}
\end{align}
% \begin{align}
%   & \mathbb E_{\mathbf u} \left [ \| \hat \nabla f(\mathbf x) - \nabla f_\mu (\mathbf x) \|^2 
%     \right ]  \leq  \mathbb E_{\mathbf u} \left [
% \| 
% \hat \nabla f(\mathbf x)
% \|_2^2
% \right ] 
% \nonumber \\
% & \leq 
%  2d \| \nabla f(\mathbf x) \|_2^2 + \frac{\mu^2}{2} L^2 d^2.
%  \label{eq: var_grad_est_random}
% \end{align}
\end{mylemma}
\textbf{Proof:}
First, by using \citep[Lemma\,4.1.a]{gao2014information} (also see \citep{shalev2012online} and \citep{nesterov2015random}), we immediately obtain that 
$f_\mu$ is $L_\mu$ smooth with $L_\mu \leq L$, and 
\begin{align}
  \nabla f_\mu (\mathbf x) 
     = \mathbb E_{\mathbf u }
    \left [
    \frac{d}{\mu} f(\mathbf x + \mu \mathbf u) \mathbf u
    \right ].    
\end{align}
Since $\mathbb E_{\mathbf u} [(d/\mu) f(\mathbf x) \mathbf u] = 0 $, we obtain
\eqref{eq: smooth_est_grad}.

% The first equality in \eqref{eq: smooth_est_grad} has been proved by several work, e.g.,
% \citep[Lemma\,4.1.a]{gao2014information}, \citep{shalev2012online} and \citep{nesterov2015random}.
% The second equality in \eqref{eq: smooth_est_grad} holds due to 
% $\mathbb E_{\mathbf u} [(d/\mu) f(\mathbf x) \mathbf u] = 0 $.

Next, we obtain  \eqref{eq: dist_f_smooth_true_random}-\eqref{eq: up_grad_fsmooth} based on \citep[Lemma\,4.1.b]{gao2014information}.
Moreover, we have
\begin{align}
     \| \nabla f_\mu (\mathbf x)\|_2^2 =  &   \| \nabla f_\mu (\mathbf x) - \nabla f (\mathbf x) + \nabla f (\mathbf x)  \|_2^2  \nonumber \\
     \leq & 2   \|  \nabla f (\mathbf x)  \|_2^2  +
     2  \| \nabla f_\mu (\mathbf x) - \nabla f (\mathbf x)   \|_2^2 \nonumber \\
     \overset{\eqref{eq: dist_smooth_true}}{\leq}  &  2 \| \nabla f(\mathbf x) \|_2^2
    + \frac{\mu^2 d^2 L^2}{2},
\end{align}
where the first inequality holds due to Lemma\,\ref{lemma: sum_exp_bd}.
Similarly, we have 
\begin{align}
    \| \nabla f (\mathbf x)\|_2^2 = \| \nabla f_\mu (\mathbf x) + \nabla f (\mathbf x) - \nabla f_\mu (\mathbf x)   \|_2^2   \overset{\eqref{eq: dist_smooth_true}}{\leq}  2 \| \nabla f_\mu(\mathbf x) \|_2^2
    + \frac{\mu^2 d^2 L^2}{2},
\end{align}
which yields
\begin{align}
    \| \nabla f_\mu(\mathbf x) \|_2^2 \geq \frac{1}{2}  \| \nabla f(\mathbf x) \|_2^2 - \frac{\mu^2 L^2 d^2}{4}.
\end{align}

In \eqref{eq: second_moment_grad_random}, the first inequality
holds due to \eqref{eq: smooth_est_grad} and  $\mathbb E[\| \mathbf a - \mathbb E[ \mathbf a ]\|_2^2] \leq \mathbb E [\| \mathbf a \|_2^2]$ for a random variable $\mathbf a$. And the second inequality of 
 \eqref{eq: second_moment_grad_random} holds due to  \citep[Lemma\,4.1.b]{gao2014information}.
The proof is now complete. \hfill $\square$

%%%%%%%%%%% average random gradient estimator

In Lemma\,\ref{lemma: smooth_f_random_ave}, we show the properties of   \ref{eq: grad_rand_ave}.

\begin{mylemma}
\label{lemma: smooth_f_random_ave}
%\citep[Lemma\,4.1]{gao2014information}
Following the conditions of Lemma\,\ref{lemma: smooth_f_random}, then   \ref{eq: grad_rand_ave} yields:

1) For any $\mathbf x \in \mathbb R^d$
\begin{align}\label{eq: smooth_est_grad_ave}
    \nabla f_\mu (\mathbf x) = \mathbb E 
    \left [
    \hat \nabla f(\mathbf x)
    \right ],
\end{align}
where $\hat \nabla f(\mathbf x)$ is given by \ref{eq: grad_rand_ave}.

% 2) For any $\mathbf x \in \mathbb R^d$
% \begin{align}
%  \mathbb E_{\mathbf u} \left [
% \| 
% \hat \nabla f(\mathbf x)
% \|_2^2
% \right ] \leq & 2 \left (1 + \frac{d}{q} \right ) \| \nabla f(\mathbf x) \|_2^2  + \left (1+\frac{1}{q} \right ) \frac{\mu^2 L^2 d^2}{2}.
% %2 d \| \nabla f(\mathbf x) \|^2 + \frac{\mu^2}{2} L^2 d^2
% \label{eq: second_moment_grad_random_ave}
% \end{align}

2) For any $\mathbf x \in \mathbb R^d$
\begin{align}
   & \mathbb E \left [ \| \hat \nabla f(\mathbf x) - \nabla f_\mu (\mathbf x) \|_2^2 
    \right ]  \leq  \mathbb E  \left [
\| 
\hat \nabla f(\mathbf x)
\|_2^2
\right ] 
 \leq 2 \left (1 + \frac{d}{q} \right ) \| \nabla f(\mathbf x) \|_2^2 + \left (1+\frac{1}{q} \right ) \frac{\mu^2 L^2 d^2}{2}.
 %2d \| \nabla f(\mathbf x) \|_2^2 + \frac{\mu^2}{2} L^2 d^2.
 \label{eq: var_grad_est_random_ave}
\end{align}
\end{mylemma}
\textbf{Proof}:
Since $\{ \mathbf u_i \}_{i=1}^q$ are i.i.d. random vectors, we have
\begin{align}
    \mathbb E \left [
    \hat \nabla f(\mathbf x)
    \right ] = \mathbb E_{\mathbf u_i} \left [
    \hat \nabla f(\mathbf x; \mathbf u_i)
    \right ] \overset{\eqref{eq: smooth_est_grad}}{=} \nabla f_\mu (\mathbf x),
\end{align}
where 
$
 \hat \nabla f(\mathbf x; \mathbf u_i) \Def
  \frac{d}{\mu} [  f ( \mathbf x + \mu \mathbf u_i ) - f ( \mathbf x  )]  \mathbf u_i
$.

In  \eqref{eq: var_grad_est_random_ave}, the first inequality 
holds due to \eqref{eq: smooth_est_grad_ave} and  $\mathbb E[\| \mathbf a - \mathbb E[ \mathbf a ]\|_2^2] \leq \mathbb E [\| \mathbf a \|_2^2]$ for a random variable $\mathbf a$.
Next, we bound the second moment of $\hat \nabla f(\mathbf x)$
\begin{align}
\mathbb E \left [ \left \| \hat{\nabla} f(\mathbf x) \right  \|_2^2  \right ]  =& \mathbb E \left [ \left  \| \frac{1}{q}\sum_{i=1}^{q}  \left ( \hat \nabla f(\mathbf x; \mathbf u_i)   - {\nabla} f_\mu(\mathbf x)  \right ) + {\nabla} f_\mu(\mathbf x)  \right \|_2^2 \right ] \nonumber \\
= &  \left \| {\nabla} f_\mu(\mathbf x)  \right \|_2^2 + \mathbb E \left [ \left  \| \frac{1}{q }\sum_{i=1}^{q}    \left ( \hat \nabla f(\mathbf x; \mathbf u_i)   - {\nabla} f_\mu(\mathbf x)  \right )  \right \|_2^2 \right ] \nonumber \\
 = &   \left \| {\nabla} f_\mu(\mathbf x)  \right \|_2^2 + \frac{1}{q} 
 \mathbb E \left [ \left  \|  \hat \nabla f(\mathbf x; {\color{black}\mathbf u_1})   - {\nabla} f_\mu(\mathbf x)  \right \|_2^2 \right  ],
% = & \| \bar{\mathbf g} \|^2 + \frac{1}{q} \mathbb E[ \| \hat{\mathbf g}_{1} \|^2 ]  - \frac{1}{q}  \| \bar{\mathbf g} \|^2,
 \label{eq: second_est_ave}
\end{align}
where the expectation is taken with respect to i.i.d. random vectors $\{ \mathbf u_i \}$, and we have used the fact that 
 $\mathbb E[ \|\hat \nabla f(\mathbf x; \mathbf u_i)   - {\nabla} f_\mu(\mathbf x) \|_2^2] = \mathbb E[ \| \hat \nabla f(\mathbf x; \mathbf u_1)   - {\nabla} f_\mu(\mathbf x)  \|_2^2]$ for any $i$. 
 Substituting \eqref{eq: up_grad_fsmooth} and \eqref{eq: second_moment_grad_random} into \eqref{eq: second_est_ave}, we obtain
\eqref{eq: var_grad_est_random_ave}. %{\color{green}using $\mathbf u_1$ is little confusing but it may still be ok}
\hfill $\square$

In Lemma\,\ref{lemma: deter_smooth_gradest}, we demonstrate the properties of   \ref{eq: grad_e}.

    \begin{mylemma}\label{lemma: deter_smooth_gradest}
    Let Assumption A1  hold and define $ f_{\mu_\ell} (\mathbf x) = \mathbb E_{u \sim U [-\mu_\ell, \mu_\ell]} f(\mathbf x + u \mathbf e_\ell)$, where $U [-\mu_\ell, \mu_\ell]$ denotes the uniform distribution at the interval $[-\mu_\ell, \mu_\ell]$. We then have:
    %1) The smoothing function $f_\ell$ is differentiable and $L$-smooth. 
    
    1)  $f_{\mu_\ell}$ is L-smooth, and 
    \begin{align}\label{eq: grad_f_smooth_coord_all}
    { \hat \nabla } f(\mathbf x) = \sum_{\ell = 1}^d  \frac{\partial  f_{\mu_\ell} (\mathbf x) }{\partial x_\ell} \mathbf e_\ell,
\end{align}
where $  { \hat \nabla } f(\mathbf x) $ is defined by \ref{eq: grad_e}, and $\partial f / \partial x_\ell$ denotes the partial derivative with respect to 
      the $\ell$th coordinate.

    2) For $\ell \in [d]$,
      \begin{align}
       & |  f_{\mu_\ell} (\mathbf x) - f (\mathbf x)|    \leq \frac{L \mu_\ell^2}{2},
          \label{eq: diff_f_coordinate_abs}
          \\
          & \left | \frac{\partial  f_{\mu_\ell} (\mathbf x) }{\partial x_\ell}   -  \frac{\partial f  (\mathbf x) }{\partial x_\ell} \right | \leq \frac{L \mu_\ell }{2}. \label{eq: diff_grad_coordinate_abs}
    \end{align}

    3) For $\ell \in [d]$,  
    \begin{align}\label{eq: grad_diff_allcoord}
 \left \| \hat{\nabla} f(\mathbf x) - \nabla f(\mathbf x)  \right \|_2^2 
  \leq \frac{L^2 d}{4} \sum_{\ell = 1}^d \mu_\ell^2.
\end{align}
    % \begin{align}
    %   % & | f_\ell (\mathbf x) - f (\mathbf x)| \leq \frac{L \mu_\ell^2}{2}, \label{eq: f_diff}\\
    %     \left \|  \frac{\partial  f_{\mu_\ell} (\mathbf x) }{\partial x_\ell}   -  \frac{\partial f  (\mathbf x) }{\partial x_\ell}  \right \|_2 \leq \frac{L \mu_\ell }{2}. \label{eq: grad_diff} %\\
    %     % & \mathbb E_\ell \|  \frac{\partial f_\ell (\mathbf x) }{\partial x_\ell}   -  \frac{\partial f  (\mathbf x) }{\partial x_\ell}   \|_2^2\leq \sum_{\ell=1}^d \frac{L^2 \mu_\ell^2}{4d},  \label{eq: sec_grad_diff} 
    % \end{align}
%\textcolor{red}{PY: extra notation $|$ in the LHS of (10)}    
    %where the expectation is taken with respect to the coordinate index $\ell$ which is randomly picked from $[d]$.
    
   %  4) If $f$ is convex, then $f_\ell$ is also convex.
    \end{mylemma}
    \textbf{Proof}: 
  For the $\ell$th coordinate, it is known from \citep[Lemma\,6]{lian2016comprehensive} that $f_{\mu_\ell}$ is $L$-smooth and %{\color{green} equality is only in the limit $\mu_l \xrightarrow{0}$ for next equation right}
  \begin{align}\label{eq: grad_f_smooth_coord}
    \frac{\partial  f_{\mu_\ell} (\mathbf x)}{\partial x_\ell}  = \frac{f( \mathbf x + \mu_{\ell} \mathbf e_\ell ) - f( \mathbf x - \mu_{\ell} \mathbf e_\ell )}{2\mu_\ell}.
    \end{align}
    Based on \eqref{eq: grad_f_smooth_coord} and the definition of \ref{eq: grad_e}, we then obtain \eqref{eq: grad_f_smooth_coord_all}.
    
    The inequalities \eqref{eq: diff_f_coordinate_abs} and \eqref{eq: diff_grad_coordinate_abs}  have   been proved by \citep[Lemma\,6]{lian2016comprehensive}.
    
Based on \eqref{eq: grad_f_smooth_coord_all} and \eqref{eq: diff_grad_coordinate_abs}, we have
    \begin{align}
& \left \| \hat{\nabla} f(\mathbf x) - \nabla f(\mathbf x)  \right \|_2^2 
\overset{\eqref{eq: grad_f_smooth_coord_all}}{=} \left \| \sum_{\ell=1}^d \left ( \frac{\partial  f_{\mu_\ell} (\mathbf x) }{\partial x_\ell} - \frac{\partial f  (\mathbf x)}{\partial x_\ell} \right ) \mathbf e_\ell  \right \|_2^2 \nonumber \\
\leq & d \sum_{\ell=1}^d \left \|  \frac{\partial  f_{\mu_\ell} (\mathbf x) }{\partial x_\ell} - \frac{\partial f  (\mathbf x)}{\partial x_\ell} \right \|_2^2 \leq \frac{L^2 d}{4} \sum_{\ell = 1}^d \mu_\ell^2, \nonumber
\end{align}
where the first inequality holds due to Lemma\,\ref{lemma: sum_exp_bd} in Sec.\,\ref{app: aux}.
The proof is now complete. \hfill $\square$

\subsection{Control variates}\label{supp: control_var}
 The gradient blending   
 in Step\,6 of SVRG (Algorithm\,1) can be interpreted using control variate  \citep{tucker2017rebar,grathwohl2017backpropagation,chatterji2018theory}. If we view $\hat {\mathbf g}_0 \Def \nabla f_{\mathcal I} (\mathbf x)$ as the raw gradient estimate at $\mathbf x$, and    $\mathbf c \Def \nabla f_{\mathcal I}(\hat{\mathbf x})$ as a control variate satisfying 
$
 \mathbb E[\mathbf c]
= \nabla f(\hat{\mathbf x})
$, then the gradient blending \eqref{eq: SVRG_update} becomes a   gradient estimate modified by a  control variate,
$
\hat{\mathbf g} = \hat {\mathbf g}_0 - (\mathbf c - \mathbb E[\mathbf c])
$. 
Here $\hat{\mathbf g}$  has the same expectation as $\hat {\mathbf g}_0$, i.e., $\mathbb E[\hat{\mathbf g}] = \mathbb E[ \hat {\mathbf g}_0 ] = \nabla f(\mathbf x)$, however, has a lower variance when $\mathbf c$ is positively
correlated with $\mathbf g_0$ (see a detailed analysis as below).

Consider the following gradient estimator,
\begin{align}
    \hat{\mathbf g} = \hat {\mathbf g}_0 - \eta (\mathbf c - \mathbb E[\mathbf c]),
\end{align}
where $\hat {\mathbf g}_0$ is a given (raw) gradient estimate, $\eta$ is an unknown coefficient,  and $\mathbf c$ is a control variate. 
It is clear that $ \hat{\mathbf g} $ has the same expectation as $\hat {\mathbf g}_0$. We then study the effect of $\mathbf c$ on the variance of $ \hat{\mathbf g} $,
\begin{align} \label{eq: trace_cov_cv}
   \tr ( \cov(\hat {\mathbf g} ) )  =  \tr( \cov ( {\hat {\mathbf g}_0}) )
    + \eta^2  \tr (\cov( \mathbf  c ) ) - 2 \eta  \tr ( \cov(\hat {\mathbf g}_0,\mathbf c ) ),
\end{align}
where $\tr(\cdot)$ denotes the trace operator, and $\cov(\cdot)$ is the covariance operator. 
When  $
\eta = \frac{\tr (\cov(\hat {\mathbf g}_0,\mathbf c) ) }{\tr (\cov( \mathbf c ))}
%- \mathrm{Cov}(\hat {\mathbf g},\mathbf c_{\boldsymbol \phi} ) \var^{-1}( \mathbf  c_{\boldsymbol \phi} )
$, the  variance of $\hat {\mathbf g}$  in \eqref{eq: trace_cov_cv} is then minimized, leading to
\begin{align}\label{eq: minvar_con}
   & \tr ( \cov(\hat {\mathbf g} )   )  = \tr( \cov (\hat {\mathbf g}_0) ) \left (
   1 - \rho(\hat{\mathbf g}_0, \mathbf c)^2
   \right ),
\end{align}
where
$  \rho(\hat{\mathbf g}_0, \mathbf c) = 
\frac{\tr ( \cov(\hat {\mathbf g}_0,\mathbf c) ) }{ \sqrt{\tr( \cov (\hat {\mathbf g}_0) )} \sqrt{ \tr (\cov(\mathbf  c ))} }
$. In \eqref{eq: minvar_con}, $ \rho(\hat{\mathbf g}_0, \mathbf c)$ indicates  the correlation strength between $\hat{\mathbf g}_0$ and $ \mathbf c$. Therefore, the   gradient estimate $\hat{\mathbf g}$ has a smaller variance than $\hat {\mathbf g}_0$  when the control variate $\mathbf c$ is \textit{positively} correlated with the latter. Moreover, if $\mathbf c$ is chosen  similar to $\hat{\mathbf g}$, then $\eta  $ would be close to 1. 

\subsection{Proof of Proposition~\ref{prop: vks}}\label{supp: lemma_vks}
In  Algorithm\,2, we recall that the mini-batch $\mathcal I$ is chosen uniformly randomly (with  replacement). \textcolor{black}{It is known from Lemma\,\ref{lemma: minibatch} and Lemma\,\ref{lemma: minibatch_noReplace}} that
\begin{align}\label{eq: E_Iv_v}
    \mathbb E_{\mathcal I_k} [\hat \nabla f_{\mathcal I_k}(\mathbf x_k^s) - \hat \nabla f_{\mathcal I_k}(\mathbf x_0^s)] = \hat \nabla f(\mathbf x_k^s) - \hat \nabla f(\mathbf x_0^s).
    %= \hat \nabla f(\mathbf x_k^s)  + \mathbf e_s.
\end{align}
We   then rewrite $\hat{\mathbf v}_k^s$ as
 \begin{align}\label{eq: E_Iv_v_2}
     \hat {\mathbf v}_k^s = & \hat \nabla f_{\mathcal I_k}(\mathbf x_k^s) - \hat \nabla f_{\mathcal I_k}(\mathbf x_0^s)  -  \mathbb E_{\mathcal I_k} [\hat \nabla f_{\mathcal I_k}(\mathbf x_k^s) - \hat \nabla f_{\mathcal I_k}(\mathbf x_0^s)] +  \hat \nabla f(\mathbf x_k^s).
 \end{align}
% Since  $\mathbb E [ \| \mathbf a - \mathbb E [\mathbf a ] \|_2^2 ] \leq \mathbb E [ \| \mathbf a \|_2^2]$ for a random variable $\mathbf a$, we have
Taking the expectation of $\|  \mathbf v_k^s \|_2^2$ with repsect to all the random variables, we have
 \begin{align}\label{eq: vks_norm_SVRG_step1}
      \mathbb E \left [ \| \hat {\mathbf v}_k^s \|_2^2  \right ]  \leq 
      & 2\mathbb E \left  [ \| \hat \nabla f_{\mathcal I_k}(\mathbf x_k^s) - \hat \nabla f_{\mathcal I_k}(\mathbf x_0^s)  -  \mathbb E_{\mathcal I_k} [\hat \nabla f_{\mathcal I_k}(\mathbf x_k^s) - \hat \nabla f_{\mathcal I_k}(\mathbf x_0^s)]    \|_2^2  \right ]   + 2 \mathbb E \left [\|\hat \nabla f(\mathbf x_k^s) \|_2^2  \right ] \nonumber \\
   \leq &
   2\mathbb E \left  [ \| \hat \nabla f_{\mathcal I_k}(\mathbf x_k^s) - \hat \nabla f_{\mathcal I_k}(\mathbf x_0^s)  -  \mathbb E_{\mathcal I_k} [\hat \nabla f_{\mathcal I_k}(\mathbf x_k^s) - \hat \nabla f_{\mathcal I_k}(\mathbf x_0^s)]    \|_2^2  \right ]  \nonumber \\
   & +
   4d \mathbb E \left [
     \| \nabla f(\mathbf x_k^s)  \|_2^2 
     \right ] + \mu^2 d^2 L^2,
%   \textcolor{black}{
%   2 \mathbb E \left [
%   \| \hat \nabla f_{\mathcal I_k}(\mathbf x_k^s) - \hat \nabla f_{\mathcal I_k}(\mathbf x_0^s)  \|_2^2
%   \right ]+ 4d \mathbb E \left [
%      \| \nabla f(\mathbf x_k^s)  \|_2^2 
%      \right ] + \mu^2 d^2 L^2,
%   }
    %  \textcolor{black}{\frac{2l}{b}}  
    %  \mathbb E \left [ \| \hat \nabla f_{i} ( \mathbf x_k^s ) - \hat \nabla f_{i} ( \mathbf x_0^s ) -   \mathbb E_{i} [\hat \nabla f_{i}(\mathbf x_k^s) - \hat \nabla f_{i}(\mathbf x_0^s) ]  \|_2^2 \right ] + 2 \mathbb E \left [\|\hat \nabla f(\mathbf x_k^s) \|_2^2  \right ] \nonumber \\
    % \leq   &  \frac{2}{b}  \mathbb E \left [ \| \hat \nabla f_{i} ( \mathbf x_k^s ) - \hat \nabla f_{i} ( \mathbf x_0^s ) \|_2^2 \right ] + 4d \mathbb E \left [
    % \| \nabla f(\mathbf x_k^s)  \|_2^2 
    % \right ] + \mu^2 d^2 L^2,
 \end{align}
 where the  first inequality holds due to Lemma\,\ref{lemma: sum_exp_bd}, and  the second inequality holds due to   \eqref{eq: second_moment_grad_random}.
 %the second inequality holds due to Lemma\,\ref{lemma: minibatch}, and the third inequality holds due to $\mathbb E[\| \mathbf a - \mathbb E[ \mathbf a ]\|_2^2] \leq \mathbb E [\| \mathbf a \|_2^2]$ and \eqref{eq: up_grad_fsmooth}.
% and we recall that $\mu$ is a smoothing parameter.
 \textcolor{black}{Based on \eqref{eq: E_Iv_v}, we note that the following holds
\begin{align}\label{eq: vks_norm_SVRG_step11}
    &\sum_{i=1}^n \left \{
    \hat{\nabla} f_i(\mathbf x_k^s) - \hat{\nabla} f_i(\mathbf x_k^s) -   \mathbb E_{\mathcal I_k} [\hat \nabla f_{\mathcal I_k}(\mathbf x_k^s) - \hat \nabla f_{\mathcal I_k}(\mathbf x_0^s)]  
    \right \} \nonumber \\
     = &  n (  \hat \nabla f(\mathbf x_k^s) - \hat \nabla f(\mathbf x_0^s) ) - n  (  \hat \nabla f(\mathbf x_k^s) - \hat \nabla f(\mathbf x_0^s) )  = \mathbf 0.
\end{align}}
\textcolor{black}{Based on \eqref{eq: vks_norm_SVRG_step11} and applying  Lemma\,\ref{lemma: minibatch} and \ref{lemma: minibatch_noReplace}, the first term at the right hand side (RHS) of \eqref{eq: vks_norm_SVRG_step1} yields
\begin{align}\label{eq: vks_norm_SVRG_step2}
%   & \mathbb E  \left [
%   \| \hat \nabla f_{\mathcal I_k}(\mathbf x_k^s) - \hat \nabla f_{\mathcal I_k}(\mathbf x_0^s)  \|_2^2 
%   \right ]  \leq  \frac{\delta_n}{bn} \sum_{i=1}^n   \mathbb E \left [ \| \hat \nabla f_{i} ( \mathbf x_k^s ) - \hat \nabla f_{i} ( \mathbf x_0^s )   \|_2^2 \right ] \nonumber \\
%     = & \frac{\delta_n}{b}   \mathbb E \left [ \| \hat \nabla f_{i} ( \mathbf x_k^s ) - \hat \nabla f_{i} ( \mathbf x_0^s )   \|_2^2 \right ],
&\mathbb E \left  [ \| \hat \nabla f_{\mathcal I_k}(\mathbf x_k^s) - \hat \nabla f_{\mathcal I_k}(\mathbf x_0^s)  -  \mathbb E_{\mathcal I_k} [\hat \nabla f_{\mathcal I_k}(\mathbf x_k^s) - \hat \nabla f_{\mathcal I_k}(\mathbf x_0^s)]    \|_2^2  \right ] \nonumber \\
\leq &  \frac{\delta_n}{bn} \sum_{i=1}^n   \mathbb E \left [ \| \hat \nabla f_{i} ( \mathbf x_k^s ) - \hat \nabla f_{i} ( \mathbf x_0^s ) - (  \hat \nabla f(\mathbf x_k^s) - \hat \nabla f(\mathbf x_0^s) )   \|_2^2 \right ] \nonumber \\
= & \mathbb E \left [ \frac{\delta_n}{b} \left (
\frac{1}{n} \sum_{i=1}^n  \| \hat \nabla f_{i} ( \mathbf x_k^s ) - \hat \nabla f_{i} ( \mathbf x_0^s ) \|_2^2 - \|\hat \nabla f(\mathbf x_k^s) - \hat \nabla f(\mathbf x_0^s) \|_2^2
\right )
\right ] \nonumber \\
\leq & \frac{\delta_n}{bn } \sum_{i=1}^n \mathbb E \left [
\| \hat \nabla f_{i} ( \mathbf x_k^s ) - \hat \nabla f_{i} ( \mathbf x_0^s ) \|_2^2
\right ].
\end{align}
where the first inequality holds due to  Lemma\,\ref{lemma: minibatch} and \ref{lemma: minibatch_noReplace} (taking the expectation   with respect to mini-batch $\mathcal I$), we define $\delta_n$ as
\begin{align}
    \delta_n = \left \{ 
    \begin{array}{ll}
        1 & \text{if $\mathcal I$ contains i.i.d. samples with replacement (Lemma\,\ref{lemma: minibatch})} \\
       I(b < n)  & \text{if $\mathcal I$ contains samples without replacement (Lemma\,\ref{lemma: minibatch_noReplace})},
    \end{array}
    \right.
\end{align}
$I(b<n) = 1$ if $b < n$ and $0$ otherwise,
 and the second equality in \eqref{eq: vks_norm_SVRG_step2} holds since $\frac{1}{n} \sum_{i=1}^n \| \mathbf x_i - \mathbf a \|_2^2 = \frac{1}{n} \sum_{i=1}^n \| \mathbf x_i \|_2^2 - \| \mathbf a\|_2^2$ when $\mathbf a = \frac{1}{n}\sum_{i=1}^n \mathbf x_i$.
}

\textcolor{black}{Substituting \eqref{eq: vks_norm_SVRG_step2} into
\eqref{eq: vks_norm_SVRG_step1}, we obtain
\begin{align}\label{eq: vks_norm_SVRG}
      \mathbb E \left [ \| \hat {\mathbf v}_k^s \|_2^2  \right ]  
    \leq   & 
    \frac{2 \delta_n}{bn}  \sum_{i=1}^n \mathbb E \left [
\| \hat \nabla f_{i} ( \mathbf x_k^s ) - \hat \nabla f_{i} ( \mathbf x_0^s ) \|_2^2
\right ]+ 4d \mathbb E \left [
    \| \nabla f(\mathbf x_k^s)  \|_2^2 
    \right ] + \mu^2 d^2 L^2.
 \end{align}
}
 
 Similar to Lemma\,\ref{lemma: smooth_f_random}, we introduce 
 a smoothing function  $f_{i,\mu}$ of $f_i$, and continue to bound the first term at the right hand side (RHS) of \eqref{eq: vks_norm_SVRG}. This yields
\begin{align}\label{eq: 1st_term_norm_vks}
 & \mathbb E \left [ \| \hat \nabla f_{i} ( \mathbf x_k^s ) - \hat \nabla f_{i} ( \mathbf x_0^s )   \|_2^2 \right ]
 \nonumber \\  \overset{\eqref{eq: sum_exp_bd}}{\leq}   & 3 \mathbb E \left [ \| \hat \nabla f_{i} ( \mathbf x_k^s ) - \nabla f_{i,\mu} ( \mathbf x_k^s )\|_2^2 \right ]    + 3 \mathbb E \left [ \|\hat \nabla f_{i,\mu} ( \mathbf x_0^s )
     -   \nabla f_{i} ( \mathbf x_0^s )   \|_2^2 \right  ]  \nonumber \\
     & +       3 \mathbb E \left [ \| \nabla f_{i,\mu} ( \mathbf x_k^s ) -  \nabla f_{i,\mu} ( \mathbf x_0^s ) \|_2^2 \right ] \nonumber \\
     \overset{\eqref{eq: second_moment_grad_random}}{\leq}
     & 6d \mathbb E[ \| \nabla f_{i} (\mathbf x_k^{s}) \|_2^2 ] +  6d \mathbb E[ \| \nabla f_{i} (\mathbf x_0^{s}) \|_2^2 ]   + 3  L^2 d^2 \mu^2 +  3 \mathbb E \left [ \| \nabla f_{i,\mu} ( \mathbf x_k^s ) -  \nabla f_{i,\mu} ( \mathbf x_0^s ) \|_2^2 \right ].
 \end{align}
Since both $f_i$ and $f_{i,\mu}$ are $L$-smooth (A1 and Lemma\,\ref{lemma: smooth_f_random}), we have
\begin{align}
& \mathbb E \left [ \| \nabla f_{i,\mu} ( \mathbf x_k^s ) -  \nabla f_{i,\mu} ( \mathbf x_0^s ) \|_2^2 \right ] \leq  L^2 \mathbb E \left [ \| \mathbf x_k^s - \mathbf x_0^s \|_2^2 \right ], \label{eq: vks_bd1} \\
  &  \mathbb E \left [ \| \nabla f_{i} (\mathbf x_0^{s}) \|_2^2 \right ] \leq 
 2 \mathbb E \left [ \| \nabla f_{i} (\mathbf x_0^{s})  - \nabla f_{i} (\mathbf x_k^{s}) \|_2^2  \right ] + 2 \mathbb E \left [\| \nabla f_{i} (\mathbf x_k^{s}) \|_2^2 \right ] \nonumber \\
& \hspace*{0.9in} \leq  2L^2 \mathbb E \left [ \| \mathbf x_0^{s} - \mathbf x_k^{s} \|_2^2 \right ]+ 2 \mathbb E \left [\| \nabla f_{i} (\mathbf x_k^{s}) \|_2^2 \right ]. %\nonumber \\
 %\leq &  2L^2 \| \mathbf x_0^{s} - \mathbf x_k^{s} \|_2^2+ 4  \mathbb E \left [  \| \nabla f_{i} (\mathbf x_k^{s}) - \nabla f (\mathbf x_k^{s})  \|_2^2 \right ]  + 4 \mathbb E \left [\| \nabla f  (\mathbf x_k^{s}) \|_2^2 \right ]  \nonumber \\
 %\leq &  2L^2 \| \mathbf x_0^{s} - \mathbf x_k^{s} \|_2^2+ 4\sigma^2 + 4 \mathbb E \left [\| \nabla f  (\mathbf x_k^{s}) \|_2^2 \right ],
 \label{eq: vks_bd2}
\end{align}
%where the last inequality holds due to Assumption A2. 
 Substituting \eqref{eq: vks_bd1} and \eqref{eq: vks_bd2} into \eqref{eq: 1st_term_norm_vks}, we obtain
 \begin{align}\label{eq: 1st_term_norm_vks_2}
  &   \mathbb E \left [ \| \hat \nabla f_{i} ( \mathbf x_k^s ) - \hat \nabla f_{i} ( \mathbf x_0^s )   \|_2^2 \right ] \nonumber \\
     \leq & 18d \mathbb E[ \| \nabla f_{i} (\mathbf x_k^{s}) \|_2^2 ] +  (12d+3)L^2  \mathbb E \left [ \| \mathbf x_0^{s} - \mathbf x_k^{s} \|_2^2 \right ] + 3  L^2 d^2 \mu^2 \nonumber \\
     \leq & 36d \mathbb E \left [  \| \nabla f_{i} (\mathbf x_k^{s}) - \nabla f (\mathbf x_k^{s})  \|_2^2 \right ]  + 36d \mathbb E \left [\| \nabla f  (\mathbf x_k^{s}) \|_2^2 \right ] \nonumber \\
     & + (12d+3)L^2  \mathbb E \left [ \| \mathbf x_0^{s} - \mathbf x_k^{s} \|_2^2 \right ] + 3  L^2 d^2 \mu^2 \nonumber \\
     \leq & 36 d \sigma^2 + 36d \mathbb E \left [\| \nabla f  (\mathbf x_k^{s}) \|_2^2 \right ] + (12d+3)L^2  \mathbb E \left [ \| \mathbf x_0^{s} - \mathbf x_k^{s} \|_2^2 \right ] + 3  L^2 d^2 \mu^2,
 \end{align}
where the last inequality holds due to Assumption A2. 

\textcolor{black}{
Substituting \eqref{eq: 1st_term_norm_vks_2} into \eqref{eq: vks_norm_SVRG}, we
have
\begin{align}
    \mathbb E \left [ \| \hat {\mathbf v}_k^s \|_2^2  \right ]  
    \leq   & \frac{6\delta_n(4d+1)L^2}{b}  \mathbb E \left [ \| \mathbf x_0^{s} - \mathbf x_k^{s} \|_2^2 \right ] \nonumber \\
    &+ \left (4d + \frac{72d \delta_n}{b} \right ) \mathbb E \left [
    \| \nabla f(\mathbf x_k^s)  \|_2^2 
    \right ] 
    + \left (1+
    \frac{6 \delta_n}{b}
    \right) d^2 L^2 \mu^2 + \frac{72 d \sigma^2 \delta_n}{b}.
\end{align}
}
 The proof is now complete.
\hfill $\square$

 \subsection{Proof of Theorem~\ref{thr: ZO_SVRG_rand}}\label{supp: thr1}
Since  $f_\mu$ is $L$-smooth (Lemma\,\ref{lemma: smooth_f_random}), from    Lemma\,\ref{lemma: Lsmooth_v2} in Sec.\,\ref{app: aux} we have 
\begin{align}\label{eq: fsmooth_def_SVRG}
f_{\mu} (\mathbf x_{k+1}^s) \leq & f_{\mu} (\mathbf x_k^s) +  \inp{\nabla f_{\mu} (\mathbf x_k^s)  }{\mathbf x_{k+1}^s - \mathbf x_k^s}  + \frac{L}{2} \| x_{k+1}^s - \mathbf x_k^s \|_2^2 \nonumber \\
= &  f_{\mu} (\mathbf x_k^s) -  \eta_k \inp{\nabla f_{\mu} (\mathbf x_k^s)  }{\hat {\mathbf v}_k^s}  + \frac{L}{2} \eta_k^2 \| \hat {\mathbf v}_k^s \|_2^2,
\end{align}
where the last equality holds due to $\mathbf x_{k+1}^{s} = \mathbf  x_{k}^{s} - \eta_{k} \hat {\mathbf v}_{k}^{s}$.
Since $\mathbf x_{k}^s$ and $\mathbf x_0^s$  are independent of $\mathcal I_{k}$ and   random directions $\mathbf u$ used for ZO gradient estimates, from \eqref{eq: smooth_est_grad}   we obtain
\begin{align}\label{eq: E_uIk_v}
    \mathbb E_{\mathbf u, \mathcal I_{k}} \left [\hat {\mathbf v}_{k}^s \right ]   = & \mathbb E_{\mathbf u, \mathcal I_{k}} \left [ \hat{\nabla} f_{{\mathcal I}_{k}} ( {\mathbf x}_{k}^{s}) - \hat{\nabla} f_{{\mathcal I}_{k}} ( {\mathbf x}_{0}^{s}) + \hat{\nabla} f (\mathbf x_0^s) \right ] \nonumber \\
    = &  \nabla f_\mu(\mathbf x_{k}^s)  + \nabla  f_{\mu} (\mathbf x_0^s) - \nabla f_\mu (\mathbf x_0^s)     =  \nabla f_\mu(\mathbf x_{k}^s).
\end{align}
Combining \eqref{eq: fsmooth_def_SVRG} and \eqref{eq: E_uIk_v}, we have
\begin{align}\label{eq: fsmooth_def_SVRG2}
 \mathbb E \left [ f_{\mu} (\mathbf x_{k+1}^s) \right ] \leq & \mathbb E \left [ f_{\mu} (\mathbf x_k^s) \right ] - \eta_k \mathbb E \left [ \| \nabla f_\mu(\mathbf x_{k}^s) \|_2^2 \right ]  + \frac{L}{2} \eta_k^2 \mathbb E \left [ \| \hat {\mathbf v}_k^s \|_2^2 \right ],   
\end{align}
where the expectation is taken with respect to all random variables. 
 
 At RHS of \eqref{eq: fsmooth_def_SVRG2}, the upper bound on $\mathbb E \left [ \| \hat {\mathbf v}_k^s \|_2^2 \right ]$ is given by %\eqref{eq: vks_norm_SVRG2} in
 Proposition\,\ref{prop: vks},
\begin{align}
\label{eq: vks_norm_SVRG2_v2}
%     \mathbb E[ \| \hat{ \mathbf v}_{k}^s \|_2^2]  \leq &
%   \frac{4(b+18)d}{b}  \mathbb E\left [ \| \nabla f  (\mathbf x_k^{s}) \|_2^2  \right ]+ \frac{6(4d+1)L^2}{b} \mathbb E \left [ \| \mathbf x_k^s - \mathbf x_0^s \|_2^2 \right ] \nonumber \\
%   &+ \frac{  (6 +b)  L^2 d^2 \mu^2}{b} + \frac{ 72d \sigma^2}{b}.
  \mathbb E[ \| \hat{ \mathbf v}_{k}^s \|_2^2]   \hspace*{-0.03in} \leq \hspace*{-0.03in}
 & \frac{4(b+18\textcolor{black}{\delta_n})d}{b}  \mathbb E\left [ \| \nabla f  (\mathbf x_k^{s}) \|_2^2  \right ]+ \frac{6(4d+1)L^2 \textcolor{black}{\delta_n}}{b} \mathbb E \left [ \| \mathbf x_k^s - \mathbf x_0^s \|_2^2 \right ] \nonumber \\
  &+ \frac{  (6\textcolor{black}{\delta_n} +b)  L^2 d^2 \mu^2}{b} + \frac{ 72d \sigma^2 \textcolor{black}{\delta_n}}{b}.
\end{align} 
In \eqref{eq: vks_norm_SVRG2_v2}, we   further bound $\mathbb E \left [\| \mathbf x_{k+1}^s - \mathbf x_0^s \|_2^2 \right ]$ as
\begin{align}\label{eq: norm_xk_dif}
  & \mathbb E \left [\| \mathbf x_{k+1}^s - \mathbf x_0^s \|_2^2 \right ] 
  =   \mathbb E \left [\| \mathbf x_{k+1}^s -\mathbf x_{k}^s + \mathbf x_{k}^s - \mathbf x_0^s \|_2^2 \right ] \nonumber \\
  = & 
   \eta_k^2 \mathbb E \left [
   \| \hat {\mathbf v}_{k}^s\|_2^2
   \right ] + \mathbb E \left [
   \|
  \mathbf x_{k}^s - \mathbf x_0^s
  \|_2^2
   \right ]
  - 2\eta_k  \mathbb E \left [ \inp{\hat {\mathbf v}_{k}^s}{\mathbf x_{k}^s - \mathbf x_0^s}
  \right ] \nonumber \\
  \overset{\eqref{eq: E_uIk_v}}{=} & \eta_k^2 \mathbb E \left [
   \| \hat {\mathbf v}_{k}^s\|_2^2
   \right ] + \mathbb E \left [
   \|
  \mathbf x_{k}^s - \mathbf x_0^s
  \|_2^2
   \right ]
  - 2\eta_k  \mathbb E \left [ \inp{\nabla f_\mu(\mathbf x_k^s)}{\mathbf x_{k}^s - \mathbf x_0^s}
  \right ] \nonumber \\
  \leq & \eta_k^2 \mathbb E \left [
   \| \hat {\mathbf v}_{k}^s\|_2^2
   \right ] + \mathbb E \left [
   \|
  \mathbf x_{k}^s - \mathbf x_0^s
  \|_2^2
   \right ]
   + 2 \eta_k\mathbb E \left [
  \frac{1}{2\beta_k}   \| \nabla f_\mu(\mathbf x_{k}^s) \|_2^2 + \frac{\beta_k}{2}  \| \mathbf x_{k}^s - \mathbf x_0^s \|_2^2
  \right ],
\end{align}
where $\beta_k $ is a positive coefficient, and the last inequality holds since $\inp{\mathbf a}{\mathbf b} \leq \frac{\beta \| \mathbf a \|_2^2 + (1/\beta) \| \mathbf b \|_2^2}{2}$ for any $\mathbf a$ and $\mathbf b$, and $\beta > 0$.

Now with \eqref{eq: vks_norm_SVRG2_v2} and \eqref{eq: norm_xk_dif} at hand, we introduce 
a   Lyapunov function  \citep{reddi2016stochastic} with respect to $f_\mu$,
\begin{align}\label{eq: Def_Rks}
    R_{k}^s = \mathbb E \left [ f_\mu (\mathbf x_k^s) + c_k \| \mathbf x_k^s - \mathbf x_0^s \|_2^2 \right ],
\end{align}
for some $c_k > 0 $.
Substituting \eqref{eq: fsmooth_def_SVRG2} and \eqref{eq: norm_xk_dif} into $R_{k+1}^s$, we obtain
\begin{align}
\label{eq: Rk_new_SCRG}
   R_{k+1}^s 
    =  & \mathbb E \left [ f_\mu (\mathbf x_{k+1}^s) + c_{k+1} \| \mathbf x_{k+1}^s - \mathbf x_0^s \|_2^2 \right ] \nonumber \\
    \leq & \mathbb E \left [f_{\mu} (\mathbf x_k^s) - \eta_k \| \nabla f_{\mu} (\mathbf x_k^s)  \|_2^2 + \frac{L}{2} \eta_k^2 \| \hat {\mathbf v}_k^s \|_2^2 \right ] \nonumber \\
    & + \mathbb E \left [
    c_{k+1} \eta_k^2 \| \hat {\mathbf v}_k^s \|_2^2 + c_{k+1} \| \mathbf x_k^s - \mathbf x_0^s \|_2^s
    \right ] \nonumber \\
    & + \mathbb E \left [ \frac{c_{k+1} \eta_k}{\beta_k} \| \nabla f_\mu (\mathbf x_k^s) \|_2^2 + c_{k+1} \beta_k \eta_k \| \mathbf x_k^{s} - \mathbf x_0^s \|_2^2 \right ] \nonumber\\
    = & \mathbb E \left [ f_\mu (\mathbf x_k^s) \right ] - \left (\eta_k-  \frac{c_{k+1} \eta_k}{\beta_k} \right ) \mathbb E \left [ \| \nabla f_\mu (\mathbf x_k^s) \|_2^2 \right ] \nonumber \\
   & + \left (c_{k+1} + c_{k+1} \beta_k \eta_k \right ) \mathbb E \left [ \| \mathbf x_k^{s} - \mathbf x_0^s \|_2^2 \right ]  + \left ( \frac{L}{2} \eta_k^2 + c_{k+1} \eta_k^2  \right ) \mathbb E \left [ \| \hat {\mathbf v}_k^s \|_2^2 \right ].
\end{align}
Moreover, substituting \eqref{eq: vks_norm_SVRG2_v2} into \eqref{eq: Rk_new_SCRG}, we have
\begin{align}
\label{eq: Rk_new_SCRG_2}
  R_{k+1}^s  
  \leq &   \mathbb E \left [ f_\mu (\mathbf x_k^s) \right ] - \left (\eta_k-  \frac{c_{k+1} \eta_k}{\beta_k} \right ) \mathbb E \left [ \| \nabla f_\mu (\mathbf x_k^s) \|_2^2 \right ]   + \left (c_{k+1} + c_{k+1} \beta_k \eta_k \right ) \mathbb E \left [ \| \mathbf x_k^{s} - \mathbf x_0^s \|_2^2 \right ] \nonumber \\
   & +  \left ( \frac{L}{2} \eta_k^2 + c_{k+1} \eta_k^2 \right )\frac{6(4d+1)L^2 \textcolor{black}{\delta_n}}{b} \mathbb E \left [ \| \mathbf x_k^s - \mathbf x_0^s \|_2^2 \right ] \nonumber \\
   & + \left ( \frac{L}{2} \eta_k^2 + c_{k+1} \eta_k^2 \right ) \frac{4db + 72d \textcolor{black}{\delta_n} }{b } \mathbb E \left [ \| \nabla f  (\mathbf x_k^{s}) \|_2^2 \right ] \nonumber \\ 
& +  \left ( \frac{L}{2} \eta_k^2 + c_{k+1} \eta_k^2 \right )\frac{ (6\textcolor{black}{\delta_n}+b) L^2 d^2 \mu^2 + 72d \sigma^2\textcolor{black}{\delta_n} }{b}.
\end{align}
Based on the definition of $
    c_k = c_{k+1} + \beta_k \eta_k  c_{k+1} + \frac{6 (4d+1)  L^2 \textcolor{black}{\delta_n} \eta_k^2}{b} c_{k+1} + \frac{3 (4d+1) L^3 \textcolor{black}{\delta_n} \eta_k^2}{b}
$ and the definition of $R_k^s$ in \eqref{eq: Def_Rks}, we can simplify the inequality \eqref{eq: Rk_new_SCRG_2} as
\begin{align}
\label{eq: Rk_new_SCRG_2_2}
 R_{k+1}^s  
  \leq &  R_k^s - \left (\eta_k-  \frac{c_{k+1} \eta_k}{\beta_k} \right ) \mathbb E \left [ \| \nabla f_\mu (\mathbf x_k^s) \|_2^2 \right ]  \nonumber \\
  & + \left ( \frac{L}{2} \eta_k^2 + c_{k+1} \eta_k^2 \right ) \frac{4db + 72d\textcolor{black}{\delta_n} }{b } \mathbb E \left [ \| \nabla f  (\mathbf x_k^{s}) \|_2^2 \right ] \nonumber \\ 
& +  \left ( \frac{L}{2} \eta_k^2 + c_{k+1} \eta_k^2 \right )\frac{ (6\textcolor{black}{\delta_n}+b) L^2 d^2 \mu^2 + 72d \sigma^2 \textcolor{black}{\delta_n} }{b} \nonumber \\
\overset{\eqref{eq: up_grad_fsmooth}}{\leq } & R_k^s - \frac{1}{2} \left (\eta_k-  \frac{c_{k+1} \eta_k}{\beta_k} \right ) \mathbb E \left [ \| \nabla f(\mathbf x_k^s) \|_2^2 \right ]   
+ \left (\eta_k-  \frac{c_{k+1} \eta_k}{\beta_k} \right ) \frac{\mu^2 d^2 L^2}{4} \nonumber \\
& + \left ( \frac{L}{2} \eta_k^2 + c_{k+1} \eta_k^2 \right ) \frac{4db + 72d\textcolor{black}{\delta_n}  }{b } \mathbb E \left [ \| \nabla f  (\mathbf x_k^{s}) \|_2^2 \right ] \nonumber \\ 
& +  \left ( \frac{L}{2} \eta_k^2 + c_{k+1} \eta_k^2 \right )\frac{ (6 \textcolor{black}{\delta_n} +b) L^2 d^2 \mu^2 + 72d \sigma^2 \textcolor{black}{\delta_n} }{b} \nonumber \\
= & R_k^s  - \gamma_k \mathbb E \left [ \| \nabla f(\mathbf x_k^s) \|_2^2 \right ]   
+ \chi_k,
\end{align}
where $\gamma_k$ and $\chi_k$ are coefficients given by
\begin{align}
&    \gamma_k = \frac{1}{2} \left (1-  \frac{c_{k+1} }{\beta_k} \right ) \eta_k
    - \left ( \frac{L}{2} + c_{k+1} \right ) \frac{4db + 72d \textcolor{black}{\delta_n} }{b } \eta_k^2,\\
    & 
    \chi_k = \left ( \frac{L}{2}  + c_{k+1} \right )\frac{ (6\textcolor{black}{\delta_n} +b) L^2 d^2 \mu^2 + 72d \sigma^2 \textcolor{black}{\delta_n}  }{b}\eta_k^2 + \left (1-  \frac{c_{k+1} }{\beta_k} \right ) \frac{\mu^2 d^2 L^2}{4}\eta_k.
\end{align}
\textcolor{black}{In the second inequality of \eqref{eq: Rk_new_SCRG_2_2}, we have used the fact that $1 - \frac{c_{k+1}}{\beta_k} >0$. This   holds for some parameter $\eta_k$ under the condition that $\gamma_k > 0$.
Even if $1 - \frac{c_{k+1}}{\beta_k} <0$ (relaxing the condition $\gamma_k > 0$), a similar inequality can be obtained using the upper bound of  $\| \nabla f_\mu (\mathbf x_k^s) \|_2^2$ in \eqref{eq: up_grad_fsmooth}. Therefore, without loss of generality, we consider $1 - \frac{c_{k+1}}{\beta_k} >0$.
}

Taking a telescopic sum for \eqref{eq: Rk_new_SCRG_2_2}, we obtain 
\begin{align}\label{eq: Rk_new_SCRG_4}
    R_m^s \leq & R_0^s - \sum_{k=0}^{m-1} \gamma_k \mathbb E \left [\| \nabla f(\mathbf x_k^s) \|_2^2 \right ]   + \chi_m,
\end{align}
where  $\chi_m = \sum_{k=0}^{m-1} \chi_k$.
It is known from \eqref{eq: Def_Rks} that 
\begin{align}
    R_0^s = \mathbb E \left [f_{\mu} (\mathbf x_0^s) \right ], \quad
    R_m^s =  \mathbb E \left [f_{\mu} (\mathbf x_m^s) \right ],
\end{align}
where the last equality used the fact that $c_m = 0$. 
Since $\tilde{\mathbf x}_{s-1} = \mathbf x_0^s $ and $\tilde{\mathbf x}_s = \mathbf x_m^s$, we obtain
\begin{align}\label{eq: R0m_SVRG}
    R_0^s - R_m^s = & \mathbb E \left [f_\mu(\tilde {\mathbf x}_{s-1}) - f_\mu(\tilde {\mathbf x}_s ) \right ]. %\nonumber %\\
   % \leq & \mathbb E[  f(\tilde {\mathbf x}_{s-1}) - f(\tilde {\mathbf x}_s ) ] + L\mu^2,
\end{align}
Substituting \eqref{eq: R0m_SVRG} into \eqref{eq: Rk_new_SCRG_4} and telescoping the sum for $s = 1,2,\ldots,S$, we obtain 
\begin{align}\label{eq: grad_mid_Sm}
     \sum_{s=1}^S \sum_{k=0}^{m-1}\gamma_k  \mathbb E[\| \nabla f(\mathbf x_k^s) \|_2^2]  % \nonumber \\
    \leq    \mathbb E[  f_\mu(\tilde {\mathbf x}_{0}) - f_\mu(\tilde {\mathbf x}_S ) ]  + S \chi_m.
   % \leq  \mathbb E[  f_\mu(\tilde {\mathbf x}_{0}) - f_\mu^* ]  + S \chi_m  
%    \leq  \mathbb E[  f(\tilde {\mathbf x}_{0}) - f^* ] + L\mu^2 + S \chi_m,
\end{align}
Denoting $f_{\mu}^* = \min_{\mathbf x} f_\mu (\mathbf x)$, from \eqref{eq: smooth_est_grad} we have $f_\mu(\tilde {\mathbf x}_{0}) - f(\tilde {\mathbf x}_{0}) \leq \frac{\mu^2 L}{2}$ and $f^* - f_\mu^* \leq  \frac{\mu^2 L}{2}$, where $f^* =  \min_{\mathbf x} f (\mathbf x)$. This yields
 \begin{align}\label{eq: dist_func_random_grad}
  f_\mu(\tilde {\mathbf x}_{0}) - f_\mu(\tilde {\mathbf x}_S ) \leq 
  f_\mu(\tilde {\mathbf x}_{0}) - f_\mu^* \leq ( f(\tilde {\mathbf x}_{0}) - f^* )
   + \mu^2 L.
\end{align}
Substituting \eqref{eq: dist_func_random_grad} into \eqref{eq: grad_mid_Sm}, we have
\begin{align}\label{eq: grad_mid_Sm_v2}
   \sum_{s=1}^S \sum_{k=0}^{m-1}\gamma_k  \mathbb E[\| \nabla f(\mathbf x_k^s) \|_2^2]  % \nonumber \\
    \leq    \mathbb E[  f(\tilde {\mathbf x}_{0}) - f^* ] + L \mu^2 + S \chi_m.
\end{align}

Let $\bar{\gamma} = \min_k \gamma_k$ and we choose
 $\bar{\mathbf x}$  uniformly random from $\{ \{ {\mathbf x}_k^s \}_{k=0}^{m-1} \}_{s=1}^S$, then ZO-SVRG satisfies
\begin{align}\label{eq: conv_SVRG_0}
       \mathbb E[\| \nabla f(\bar {\mathbf x}) \|_2^2]  
    \leq     \frac{ \mathbb E[  f(\tilde {\mathbf x}_{0}) - f^* ] }{T \bar \gamma} + \frac{L\mu^2}{T \bar \gamma} + \frac{S \chi_m}{T \bar \gamma}.
\end{align}
The proof is now complete.
\hfill $\square$

\subsection{Proof of Corollary~\ref{col: SVRG_simple}}\label{supp: cor1}
We start by rewriting $c_k$ in \eqref{eq: ck_coeff} as
\begin{align}\label{eq: ck_recursion}
    c_k =  (1+ \theta) c_{k+1}  + \frac{3 (1+4d)L^3  \textcolor{black}{\delta_n} \eta^2}{b},
\end{align}
where $\theta = \beta \eta + \frac{6(1+4d)L^2 \textcolor{black}{\delta_n} \eta^2}{b}$. The recursive formula   \eqref{eq: ck_recursion} implies that $c_k \leq c_0$ for any $k$, and
\begin{align}\label{eq: c0}
    c_0  = \frac{3 (1+4d) L^3 \textcolor{black}{\delta_n} \eta^2}{b} \frac{(1+ \theta)^m - 1}{\theta}.
\end{align}
Based on the choice of $\eta = \frac{\rho}{L d}$ and $\beta = L$, we have 
\begin{align}\label{eq: theta}
    \theta = \frac{\rho}{ d} + \frac{6  \rho^2 \textcolor{black}{\delta_n}}{ b  d^2}  + \frac{24  \rho^2 \textcolor{black}{\delta_n}}{  b d} \leq \frac{31 \rho}{ d},
\end{align}
\textcolor{black}{where we have used the fact that $\delta_n \leq 1$.}
Substituting \eqref{eq: theta} into \eqref{eq: c0}, we have
\begin{align}\label{eq: ck_upper}
   c_k \leq c_0 & = \frac{3 (1+4d) L^3 \textcolor{black}{\delta_n}}{b} \frac{\eta^2}{\theta } [(1+ \theta)^m - 1]   = \frac{3(1+4d)L\rho \textcolor{black}{\delta_n}}{db + 24\rho d + 6\rho}  [(1+ \theta)^m - 1] \nonumber \\
   & \leq \frac{15 d  L \rho \textcolor{black}{\delta_n} }{db}[(1+ \theta)^m - 1]   \leq \frac{15 L \rho \textcolor{black}{\delta_n}}{b} (e - 1) \leq \frac{30L\rho \textcolor{black}{\delta_n}}{b},
 \end{align}
 where the third inequality holds since
 $
 (1+\theta)^m \leq (1+ \frac{31\rho}{d})^m
 $, $m = \ceil{ \frac{d}{31 \rho} } $,   
 $(1 + 1/a)^a \leq   \lim_{a \to \infty} (1 + \frac{1}{a})^a = e
 $ for $a > 0$ \cite[Appendix\,E]{reddi2016stochastic}, and for east of representation, the last inequality   loosely uses the notion `$\leq$' since $e < 3$.
 
  We recall from \eqref{eq: grad_norm_SVRG} and \eqref{eq: coeff_gammak} that
 \begin{align}
     \bar \gamma = \min_{0 \leq k \leq m-1}  \left \{ 
     \frac{\eta_k}{2} - \frac{c_{k+1} \eta_k}{2 \beta_k} 
     - \eta_k^2 \left (\frac{L}{2} + c_{k+1} \right ) \left ( 4d + \frac{72d \textcolor{black}{\delta_n} }{b} \right ) \right \}.
 \end{align}
Since $\eta_k =\eta $, $\beta_k = \beta$, and $c_k \leq c_0$, we have
 \begin{align}\label{eq: min_gam}
     \bar \gamma \geq \frac{\eta}{2} - \frac{c_0}{2 \beta} \eta - \eta^2 L \left (2d + \frac{36d}{b} \right ) - \eta^2 c_0 \left (4d + \frac{72d \textcolor{black}{\delta_n}}{b} \right ).
 \end{align}
 From \eqref{eq: ck_upper} and the definition of $\beta$, we have
 \begin{align}
   &  \frac{c_0}{2\beta} \leq \frac{15\rho}{b} \label{eq: c0_beta} \\
   &  \eta L \left (2d + \frac{36d}{b} \right ) =
     \rho \left (2 + \frac{36}{b} \right )   \label{eq: eta_L} \\
     &   \eta c_0 \left  (4d + \frac{72d \textcolor{black}{\delta_n}}{b} \right ) \overset{\eqref{eq: ck_upper}}{\leq}  
     \frac{\rho}{L d}\frac{30 L \rho}{b } \left  (4d + \frac{72d}{b} \right ) 
    \leq  \frac{120 \rho^2  }{b} +
    \frac{2160 \rho^2  }{b^2}. \label{eq: eta_c0}
 \end{align}
  
 %When $b \geq 18$, 
 Substituting \eqref{eq: c0_beta}-\eqref{eq: eta_c0} into \eqref{eq: min_gam}, we obtain
  \begin{align}
     \bar \gamma \geq & \eta \left (
     \frac{1}{2} - \frac{15 \rho}{b}
   - 4 \rho - \frac{240 \rho^2}{b}
     \right ) 
     \geq 
     \eta \left (
     \frac{1}{2} -
   259 \rho
     \right ), % \nonumber \\
   %  \geq &  \eta \left (
    % \frac{1}{2} -
    % 15c - \frac{244cb}{d^{\frac{2}{3}}}
    % \right ) 
 \end{align}
 where  we have used the fact that
$\rho^2 \leq \rho$. Moreover, if we set $\rho \leq \frac{1}{518}$, then $\bar{\gamma} > 0$. In other words,  the current parameter setting is valid for Theorem\,\ref{thr: ZO_SVRG_rand}.
Upon defining a universal constant $ \alpha_0 =  \left (
     \frac{1}{2} -
   259 \rho
     \right )$, 
we have
\begin{align}\label{eq: gamma_bar_bound}
     \bar \gamma \geq & 
      \eta \alpha_0.
 \end{align}
% where $\alpha_0 > 0$   is independent of $T$, $d$ and $b$.

 Next, we find the upper bound on $\chi_m$ in \eqref{eq: coeff_chik} given the current parameter setting and $c_k \leq c_0$,
 \begin{align}
     \chi_m \leq  m \eta \frac{\mu^2 d^2 L^2}{4}  + m \eta^2 \left ( \frac{L}{2}  + c_{0} \right )\frac{  72d \sigma^2 \textcolor{black}{\delta_n} + (6\textcolor{black}{\delta_n} +b) L^2 d^2 \mu^2 }{b}.
 \end{align}
 Since 
 $
 \frac{L}{2} + c_0 \leq \frac{L}{2} + 30 L \rho b^{-1} \leq \frac{L}{2} + 2 L = \frac{5 L}{2}
 $ (suppose $b \geq 18$ without loss of generality), based on \eqref{eq: gamma_bar_bound} we have
 \begin{align}
     \frac{\chi_m}{\bar \gamma} \leq&
   m  \frac{d^2 L^2 \mu^2}{4 \alpha_0}  +
    m  \frac{5 L}{2 \alpha_0} \frac{72 d \sigma^2  \textcolor{black}{\delta_n}}{b} \frac{\rho}{L d}   + m \frac{5L}{2 \alpha_0} \left (
\frac{6L^2 d^2 \mu^2 \textcolor{black}{\delta_n}}{b} + L^2 d^2 \mu^2
\right )\frac{\rho}{L d}.
 \end{align}
 Since  $T = S m$, and  $\mu = \frac{1}{\sqrt{d T}}$, the above inequality yields
 \begin{align}\label{eq: const_error_bd}
     \frac{S\chi_m}{ T \bar \gamma} \leq&
    \frac{d L^2}{4 \alpha_0 T }  +
     \frac{180 \sigma^2 \rho \textcolor{black}{\delta_n} }{b \alpha_0}  +  \frac{5 L^2   }{2 \alpha_0} \left (
\frac{6  }{b} +  1
\right )  \frac{\rho}{  T} 
=  O \left (
\frac{d}{T} + \frac{\textcolor{black}{\delta_n}}{b}
\right ),
 \end{align}
 where in the big $O$ notation, we only keep the dominant terms and ignore the constant numbers that are independent of  $d$, $b$, and $T$.
 
  Substituting \eqref{eq: gamma_bar_bound} and 
 \eqref{eq: const_error_bd} into \eqref{eq: grad_norm_SVRG}, we have
\begin{align} \label{eq: grad_norm_SVRG_special}
    \mathbb E[ \| \nabla f(\bar { \mathbf x }) \|_2^2 ] \leq &
\frac{[f(\tilde{\mathbf x}_0) - f^*]}{T \alpha_0 }
\frac{L d}{ \rho }
+   \frac{L^2 }{T^2 \alpha_0 \rho}
+ \frac{S\chi_m}{ T \bar \gamma}  
=  O \left (
\frac{d}{T} + \frac{\textcolor{black}{\delta_n}}{b}
\right ).
\end{align}

 The proof is now complete. 
 \hfill $\square$

  \subsection{Proof of Proposition~\ref{prop: grad_err}}\label{supp: grad_err}

For  \ref{eq: grad_rand}, based on \eqref{eq: dist_smooth_true} and \eqref{eq: second_moment_grad_random}, we have
\begin{align}\label{eq: diff_grad_estg}
&     \mathbb E \left [
    \| \hat \nabla f(\mathbf x) - \nabla f(\mathbf x) \|_2^2
    \right ] 
\leq   \mathbb E \left [
    \| \hat \nabla f(\mathbf x) - \nabla f_\mu (\mathbf x) 
    + \nabla f_\mu(\mathbf x)
    - \nabla f(\mathbf x) \|_2^2
    \right ] \nonumber \\
    \leq & 2 \mathbb E \left [  \| \hat \nabla f(\mathbf x) - \nabla f_\mu (\mathbf x) \|_2^2 \right ] + 2  \| \nabla f_\mu(\mathbf x)
    - \nabla f(\mathbf x) \|_2^2 \nonumber \\
    \leq & 4d \| \nabla f(\mathbf x) \|_2^2  + \frac{3\mu^2  L^2 d^2}{2}
    = O \left (d \| \nabla f(\mathbf x) \|_2^2 + \mu^2  L^2 d^2
    \right ).
     %\nonumber \\
  %  \leq & 4d \sigma + \frac{3\mu^2 d^2 L^2}{2}.
\end{align}

Similarly, for    \ref{eq: grad_rand_ave}, based on \eqref{eq: dist_smooth_true} and \eqref{eq: var_grad_est_random_ave}, we have 
\begin{align}\label{eq: diff_grad_estg_ave}
     \mathbb E \left [
    \| \hat \nabla f(\mathbf x) - \nabla f(\mathbf x) \|_2^2
    \right ] 
% \leq   \mathbb E \left [
%     \| \hat \nabla f(\mathbf x) - \nabla f_\mu (\mathbf x) 
%     + \nabla f_\mu(\mathbf x)
%     - \nabla f(\mathbf x) \|_2^2
%     \right ] \nonumber \\
%     \leq & 2 \mathbb E \left [  \| \hat \nabla f(\mathbf x) - \nabla f_\mu (\mathbf x) \|_2^2 \right ] + 2  \| \nabla f_\mu(\mathbf x)
%     - \nabla f(\mathbf x) \|_2^2
    \leq &  4 \left (1 + \frac{d}{q} \right ) \| \nabla f(\mathbf x) \|_2^2 + \left (3+\frac{2}{q} \right ) \frac{\mu^2 L^2 d^2}{2} \nonumber \\
    = & O \left (
    \frac{q+d}{q} \| \nabla f(\mathbf x) \|_2^2 + \mu^2 L^2 d^2
    \right ),
  %  \leq & 4d \sigma + \frac{3\mu^2 d^2 L^2}{2}.
\end{align}
where we have used the fact that $\frac{2}{q} \leq 3$.

Finally,    using \eqref{eq: grad_diff_allcoord}, the proof is then complete. 
\hfill $\square$

 \subsection{Proof of Theorem~\ref{thr: ZO_SVRG_rand_ave}} \label{supp: thr2}
 
 Motivated by Proposition\,\ref{prop: vks}, we first bound $\| \hat {\mathbf v}_k^s \|_2^2$. Following \eqref{eq: E_Iv_v}-\eqref{eq: vks_norm_SVRG}, we have
 \begin{align}\label{eq: vks_norm_SVRG_ave}
      \mathbb E \left [ \| \hat {\mathbf v}_k^s \|_2^2 \right ]  \leq 
      & 
      %\frac{2}{b} \mathbb E \left [ \| \hat \nabla f_{i}(\mathbf x_k^s) - \hat \nabla f_{i}(\mathbf x_0^s)  \|_2^2 \right ]   + 2 \mathbb E \left [\|\hat \nabla f(\mathbf x_k^s) \|_2^2  \right ] 
   \textcolor{black}{\frac{2 \delta_n}{bn}  \sum_{i=1}^n \mathbb E \left [
\| \hat \nabla f_{i} ( \mathbf x_k^s ) - \hat \nabla f_{i} ( \mathbf x_0^s ) \|_2^2
\right ]+   2 \mathbb E \left [\|\hat \nabla f(\mathbf x_k^s) \|_2^2  \right ]} 
      \nonumber \\
     \overset{\eqref{eq: var_grad_est_random_ave}}{\leq} &
     \frac{2 \delta_n}{bn}  \sum_{i=1}^n \mathbb E \left [
\| \hat \nabla f_{i} ( \mathbf x_k^s ) - \hat \nabla f_{i} ( \mathbf x_0^s ) \|_2^2
\right ]   + 
    4 \left (1 + \frac{d}{q} \right ) \| \nabla f(\mathbf x_k^s) \|_2^2 + \left (1+\frac{1}{q} \right )  \mu^2 L^2 d^2.
 \end{align}
 Moreover, following \eqref{eq: 1st_term_norm_vks}-\eqref{eq: 1st_term_norm_vks_2} together with \eqref{eq: var_grad_est_random_ave}, we can obtain that
 \begin{align}\label{eq: 1st_term_norm_vks_2_ave}
   & \mathbb E \left [ \| \hat \nabla f_{i} ( \mathbf x_k^s ) - \hat \nabla f_{i} ( \mathbf x_0^s ) \|_2^2 \right  ]   \nonumber \\
     \leq &  36 \left (1+\frac{d}{q} \right ) \sigma^2 + 36 \left (1+ \frac{d}{q} \right ) \mathbb E \left [ \| \nabla f (\mathbf x_k^{s}) \|_2^2 \right ]    + \left (12 \frac{d}{q}+15 \right ) L^2 \| \mathbf x_k^s - \mathbf x_0^s \|_2^2 \nonumber \\
     & + 3 \left (1+\frac{1}{q} \right ) L^2  \mu^2 d^2.
 \end{align}
 Substituting \eqref{eq: 1st_term_norm_vks_2_ave} into \eqref{eq: vks_norm_SVRG_ave}, we have
 \begin{align}
\label{eq: vks_norm_SVRG2_ave}
    \mathbb E \left [ \| \hat {\mathbf v}_{k}^s \|_2^2 \right ]  
    \leq & 
   \frac{4(b+18\textcolor{black}{\delta_n})}{b} \left (1+\frac{d}{q} \right )\mathbb E[ \| \nabla f  (\mathbf x_k^{s}) \|_2^2 ]  + \frac{6\textcolor{black}{\delta_n}}{b}\left (\frac{4d}{q}+5 \right )L^2 \| \mathbf x_k^s - \mathbf x_0^s \|_2^2  \nonumber \\
    & + 
    \frac{6\textcolor{black}{\delta_n}+b}{b} \left (1+\frac{1}{q} \right ) L^2 \mu^2 d^2  + \frac{72\textcolor{black}{\delta_n}}{b} \left (1+\frac{d}{q} \right ) \sigma^2.
\end{align}

% Next, we bound $\mathbb E[\| \mathbf x_{k+1}^s - \mathbf x_0^s \|_2^2]$.  
 Following \eqref{eq: norm_xk_dif}-\eqref{eq: Rk_new_SCRG} and substituting \eqref{eq: vks_norm_SVRG2_ave}  into \eqref{eq: Rk_new_SCRG}, we have
 \begin{align}
\label{eq: Rk_new_SCRG_2_ave}
  R_{k+1}^s  
  \leq &   \mathbb E \left [ f_\mu (\mathbf x_k^s) \right ] - \left (\eta_k-  \frac{c_{k+1} \eta_k}{\beta_k} \right ) \mathbb E \left [ \| \nabla f_\mu (\mathbf x_k^s) \|_2^2 \right ]   + \left (c_{k+1} + c_{k+1} \beta_k \eta_k \right ) \mathbb E \left [ \| \mathbf x_k^{s} - \mathbf x_0^s \|_2^2 \right ] \nonumber \\
   & +  \left ( \frac{L}{2} \eta_k^2 + c_{k+1} \eta_k^2 \right )\frac{6(4d+5q)L^2\textcolor{black}{\delta_n}}{bq} \mathbb E \left [ \| \mathbf x_k^s - \mathbf x_0^s \|_2^2 \right ] \nonumber \\
   & + \left ( \frac{L}{2} \eta_k^2 + c_{k+1} \eta_k^2 \right ) \frac{(72 \textcolor{black}{\delta_n} + 4b) (q+d) }{b q} \mathbb E \left [ \| \nabla f  (\mathbf x_k^{s}) \|_2^2 \right ] \nonumber \\ 
& +  \left ( \frac{L}{2} \eta_k^2 + c_{k+1} \eta_k^2 \right )\frac{ (6\textcolor{black}{\delta_n}+b)(q+1) L^2 d^2 \mu^2 + 72 (q+d) \sigma^2\textcolor{black}{\delta_n} }{bq}.
\end{align}
Based on 
%$c_k = c_{k+1} + c_{k+1} \beta_k \eta_k+ \frac{6(4d+5q)L^2}{bq} \eta_k^2 c_{k+1} + 
%\frac{3(4d+5q)L^3}{bq} \eta_k^2$ and 
the definitions of $
 c_k  =  \left [ 1 +  \beta_k \eta_k+ \frac{6(4d+5q)L^2\textcolor{black}{\delta_n}}{bq} \eta_k^2 \right ] c_{k+1} + 
\frac{3(4d+5q)L^3\textcolor{black}{\delta_n}}{bq} \eta_k^2
 $  and $R_k^s$ given by \eqref{eq: Def_Rks}, we can simplify \eqref{eq: Rk_new_SCRG_2_ave} to
\begin{align}
\label{eq: Rk_new_SCRG_2_2_ave}
 R_{k+1}^s  
%   \leq &  R_k^s - \left (\eta_k-  \frac{c_{k+1} \eta_k}{\beta_k} \right ) \mathbb E \left [ \| \nabla f_\mu (\mathbf x_k^s) \|_2^2 \right ]    \nonumber \\
%   & + \left ( \frac{L}{2} \eta_k^2 + c_{k+1} \eta_k^2 \right ) \frac{(72 + 4b)(q+d) }{b q } \mathbb E \left [ \| \nabla f  (\mathbf x_k^{s}) \|_2^2 \right ] \nonumber \\ 
% & + \left ( \frac{L}{2} \eta_k^2 + c_{k+1} \eta_k^2 \right )\frac{ (6+b)(q+1) L^2 d^2 \mu^2 + 72 (q+d) \sigma^2 }{bq} \nonumber \\
\overset{\eqref{eq: up_grad_fsmooth}}{\leq } & R_k^s - \frac{1}{2} \left (\eta_k-  \frac{c_{k+1} \eta_k}{\beta_k} \right ) \mathbb E \left [ \| \nabla f(\mathbf x_k^s) \|_2^2 \right ]   
+ \left (\eta_k-  \frac{c_{k+1} \eta_k}{\beta_k} \right ) \frac{\mu^2 d^2 L^2}{4} \nonumber \\
& + \left ( \frac{L}{2} \eta_k^2 + c_{k+1} \eta_k^2 \right ) \frac{(72 \textcolor{black}{\delta_n} + 4b)(q+d) }{b q } \mathbb E \left [ \| \nabla f  (\mathbf x_k^{s}) \|_2^2 \right ] \nonumber \\ 
& + \left ( \frac{L}{2} \eta_k^2 + c_{k+1} \eta_k^2 \right )\frac{ (6 \textcolor{black}{\delta_n} +b)(q+1) L^2 d^2 \mu^2 + 72 (q+d) \sigma^2 \textcolor{black}{\delta_n} }{bq} \nonumber \\
= & R_k^s  - \gamma_k \mathbb E \left [ \| \nabla f(\mathbf x_k^s) \|_2^2 \right ]   
+ \chi_k,
\end{align}
where $ \gamma_k$ and $\chi_k $ are defined coefficients in Theorem\,\ref{thr: ZO_SVRG_rand_ave}.
% \begin{align}
% &    \gamma_k = \frac{1}{2} \left (1-  \frac{c_{k+1} }{\beta_k} \right ) \eta_k
%     - \left ( \frac{L}{2} + c_{k+1} \right ) \frac{(72 + 4b)(q+d) }{b q } \eta_k^2,\\
%     & 
%     \chi_k = \left ( \frac{L}{2}  + c_{k+1} \right )\frac{ (6+b)(q+1) L^2 d^2 \mu^2 + 72 (q+d) \sigma^2 }{bq}\eta_k^2 + \left (1-  \frac{c_{k+1} }{\beta_k} \right ) \frac{\mu^2 d^2 L^2}{4}\eta_k.
% \end{align}

Based on \eqref{eq: Rk_new_SCRG_2_2_ave}
and following the same argument in \eqref{eq: Rk_new_SCRG_4}-\eqref{eq: conv_SVRG_0}, we then achieve
\begin{align} 
       \mathbb E[\| \nabla f(\bar {\mathbf x}) \|_2^2]  
    \leq     \frac{ \mathbb E[  f(\tilde {\mathbf x}_{0}) - f^* ] }{T \bar \gamma} + \frac{L\mu^2}{T \bar \gamma} + \frac{S \chi_m}{T \bar \gamma}. \label{eq: conv_SVRG_0_ave}
\end{align}

The rest of the proofs essentially follow along the lines of Corollary\,\ref{col: SVRG_simple} with the added complexity of the mini-batch parameter $q$
in $c_k$, $\gamma_k$ and $\chi_k$.

Let $\theta = \beta_k \eta_k+ \frac{6(4d+5q)L^2\textcolor{black}{\delta_n}}{bq} \eta_k^2$, then $c_k = c_{k+1} (1+ \theta)  + \frac{3 (4d+5q) L^3 \eta_k^2 \textcolor{black}{\delta_n}}{bq} $.
This leads to
\begin{align}\label{eq: c0_ave}
    c_0 & = \frac{3 (4d+5q) L^3 \eta^2 \textcolor{black}{\delta_n} }{bq} \frac{(1+ \theta)^m - 1}{\theta}.
\end{align}
Based on the choice of $\eta$ and $\beta$, we have 
\begin{align}\label{eq: theta_ave}
    \theta = \frac{\rho}{ d} + \frac{24  \rho^2 \textcolor{black}{\delta_n} }{ b  dq}  + \frac{30  \rho^2 \textcolor{black}{\delta_n} }{  b d^2} \leq \frac{55 \rho}{d}. %\leq \frac{31 b c}{ d}.
\end{align}
Substituting \eqref{eq: theta_ave} into \eqref{eq: c0_ave}, we have
\begin{align}\label{eq: ck_upper_ave}
   c_k \leq c_0 & = \textcolor{black}{\delta_n} \frac{3 (5+4d/q) L^3}{b} \frac{\eta^2}{\theta } [(1+ \theta)^m - 1]   = \textcolor{black}{\delta_n} \frac{3(5+4d/q)L\rho}{db + 24\rho d/q + 30\rho}  [(1+ \theta)^m - 1] \nonumber \\
   & \leq \textcolor{black}{\delta_n} \frac{3(5+4d/q)L\rho}{db}[(1+ \theta)^m - 1]  \leq \textcolor{black}{\delta_n}  \frac{27 L \rho}{b\min\{d,q\}} [(1+ \theta)^m - 1] \nonumber \\
   & \leq \frac{27 L \rho \textcolor{black}{\delta_n} }{b\min\{d,q\}} (e - 1) \leq \frac{54L\rho \textcolor{black}{\delta_n} }{b \min \{ d,q\}},
 \end{align}
 where the third inequality holds since  $5+4d/q \leq 9 d /q$ if $d \geq q$, and $5+4d/q \leq 9 $ otherwise, and the forth inequality holds similar to  \eqref{eq: ck_upper} under $m = \ceil{\frac{d}{55 \rho} }$.
 
 According to the definition of $\bar{\gamma} = \min_k \gamma_k$, we have
  \begin{align}\label{eq: min_gam_ave}
     \bar \gamma \geq & \frac{\eta}{2} - \frac{c_0}{2 \beta} \eta - \eta^2 L \frac{(36 \textcolor{black}{\delta_n} +2b)(q+d)}{bq}  - \eta^2 c_0 \frac{(72 \textcolor{black}{\delta_n} +4b)(q+d)}{bq}.
 \end{align}

  From \eqref{eq: ck_upper_ave} and the definition of $\beta = L$, we have
 \begin{align}\label{eq: c0_beta_ave}
     \frac{c_0}{2\beta} \leq \frac{27\rho}{b \min\{d,q\}}. %\leq 15 c.
 \end{align}
 Since $\eta = \rho/(Ld)$, we have
 \begin{align}\label{eq: eta_L_ave}
     \eta L  \frac{(36\textcolor{black}{\delta_n}+2b)(q+d)}{bq}  \leq \frac{2\rho}{\min\{ d,q\}} \left (\frac{36}{b} +2 \right ),
 \end{align}
 where we used the fact that $\frac{1}{d} + \frac{1}{q } \leq \frac{2}{\min\{ d,q\}}$. Moreover, we have
  \begin{align}\label{eq: eta_c0_ave}
     \eta c_0   \frac{(72 \textcolor{black}{\delta_n}+4b)(q+d)}{bq} \leq  &
     \frac{\rho}{L}\frac{54L  \rho}{b \min\{ q,d\}}  \left (4 + \frac{72}{b} \right ) \left (\frac{1}{d}+ \frac{1}{q} \right ) \nonumber \\
    \leq  & \frac{108 \rho^2}{b \min\{ d,q\}^2} \left (4 + \frac{72}{b} \right )
 \end{align}
Substituting \eqref{eq: c0_beta_ave}-\eqref{eq: eta_c0_ave} into \eqref{eq: min_gam_ave}, and following the arguments in \eqref{eq: gamma_bar_bound}, we   obtain 
\begin{align}\label{eq: gamma_bar_bound_ave}
     \bar \gamma \geq & 
      \alpha_0 \eta,
 \end{align}
where $\alpha_0 > 0$  is a universal constant that is   independent of $T$, $d$ and $b$.

Based on $ \chi_k =    \left (1-  \frac{c_{k+1} }{\beta_k} \right ) \frac{\mu^2 d^2 L^2}{4}\eta_k + \left ( \frac{L}{2}  + c_{k+1} \right )\frac{ (6\textcolor{black}{\delta_n}+b)(q+1) L^2 d^2 \mu^2 + 72 (q+d) \sigma^2 \textcolor{black}{\delta_n}}{bq}\eta_k^2$, the upper bound on $\chi_m = \sum_k \chi_k$ is given by
 \begin{align}
     \chi_m \leq  &  \eta m  
     %\left (1-  \frac{c_{0} }{\beta} \right ) 
     \frac{\mu^2 d^2 L^2}{4} +
    \eta m \left ( \frac{L}{2}  + c_{0} \right )\frac{ (6 \textcolor{black}{\delta_n} +b)(q+1) L^2 d^2 \mu^2 + 72 (q+d) \sigma^2 \textcolor{black}{\delta_n} }{bq} \eta .
 \end{align}
Using \eqref{eq: ck_upper_ave} and assuming $b \geq 18$ (without loss of generality), then
 $
 \frac{L}{2} + c_0 \leq \frac{L}{2} + 54 L \rho b^{-1}  \leq \frac{7 L}{2}
 $. This yields
 \begin{align}
     \frac{\chi_m}{\bar \gamma} \leq&
 \frac{m}{\alpha_0} \frac{d^2 L^2 }{4} \frac{1}{dT}  + \frac{m}{\alpha_0} 
   \frac{7L}{2} \frac{(6\textcolor{black}{\delta_n}+b)(q+1)L^2 d^2}{bq} \frac{1}{dT} \frac{\rho}{Ld } +  \frac{m}{\alpha_0} \frac{7L}{2} \frac{72 \sigma^2}{b} \left(\frac{1}{d} + \frac{1}{q} \right ) \rho \textcolor{black}{\delta_n} \nonumber \\
   \leq & O \left (
   \frac{md}{T}  + \frac{m \textcolor{black}{\delta_n}}{b \min\{d,q \}}
   \right )
 \end{align}
Since  $T = S m$, we have
 \begin{align}\label{eq: const_error_bd_ave}
     \frac{S\chi_m}{ T \bar \gamma} \leq O \left (
\frac{d}{T} + \frac{\textcolor{black}{\delta_n}}{b \min\{d,q\}}
\right ).
 \end{align}

 Substituting \eqref{eq: gamma_bar_bound_ave} and 
 \eqref{eq: const_error_bd_ave} into \eqref{eq: grad_norm_SVRG}, we have
\begin{align} \label{eq: grad_norm_SVRG_ave_special}
    \mathbb E[ \| \nabla f(\bar { \mathbf x }) \|_2^2 ] \leq &
\frac{[f(\tilde{\mathbf x}_0) - f^*]}{T \alpha_0 }
\frac{L d}{ \rho }
+   \frac{L^2 }{T^2 \alpha_0 \rho}
+ \frac{S\chi_m }{ T \bar \gamma} = O\left ( 
\frac{ d}{T} + \frac{\textcolor{black}{\delta_n}}{b \min\{d,q \}}
\right ).
\end{align}

 \hfill $\square$

 \subsection{Proof of Theorem~\ref{thr: ZO_SVRG_coord}} \label{supp: thr3}

 Since $f$ is $L$-smooth, we have
\begin{align}\label{eq: fsmooth_def_SVRG_coord}
f (\mathbf x_{k+1}^s) \leq & f (\mathbf x_k^s) - \eta_k \inp{\nabla f (\mathbf x_k^s)  }{\hat {\mathbf v}_k^s}  + \frac{L}{2} \eta_k^2 \| \hat {\mathbf v}_k^s \|_2^2.
\end{align}
Since $\mathbf x_{k}^s$ and $\mathbf x_0^s$  are independent of $\mathcal I_{k}$  used in $\hat \nabla f_{\mathcal I_{k}} (\mathbf x_{k}^s)$ and $\hat \nabla f_{\mathcal I_{k}} (\mathbf x_{0}^s)$,  we obtain
\begin{align}\label{eq: E_uIk_v_coord}
    \mathbb E_{ \mathcal I_{k}} [\mathbf v_{k}^s]  &  =   \hat \nabla f(\mathbf x_{k}^s)  + \hat \nabla  f (\mathbf x_0^s) - \hat \nabla f (\mathbf x_0^s)   =  \hat \nabla f(\mathbf x_{k}^s),
\end{align}
where we recall that a deterministic gradient estimator is used.
Combining \eqref{eq: fsmooth_def_SVRG_coord} and \eqref{eq: E_uIk_v_coord}, we have
\begin{align}\label{eq: fsmooth_def_SVRG2_coord}
 \mathbb E \left [ f (\mathbf x_{k+1}^s)  \right ] 
 \leq & \mathbb E \left [ f (\mathbf x_k^s)  \right ] - \eta_k \mathbb E \left [ \inp{\nabla f(\mathbf x_k^s)}{\hat \nabla f(\mathbf x_k^s)} \right ] + \frac{L}{2} \eta_k^2 \mathbb E \left [  \| \hat {\mathbf v}_k^s \|_2^2 \right ].   
\end{align}

 In \eqref{eq: fsmooth_def_SVRG2_coord}, we   bound $ -2 \mathbb E \left [ \inp{\nabla f(\mathbf x_k^s)}{\hat \nabla f(\mathbf x_k^s)} \right ]$ as,
\begin{align}\label{eq: prod_est_coord} 
 -2  \mathbb E \left [ \inp{\nabla f (\mathbf x_k^s)}{\hat{\nabla} f (\mathbf x_k^s)} \right ] 
\leq  & \mathbb E \left [ \|  {\nabla} f  ( {\mathbf x}_{k}^s) - \hat {\nabla} f  ( {\mathbf x}_{k}^s)  \|_2^2 \right ] - \left [ \mathbb E \| \nabla f(\mathbf x_k^s) \|_2^2 \right ] \nonumber \\
\overset{\eqref{eq: grad_diff_allcoord}}{\leq}
  & \frac{L^2 d^2 \mu^2}{4}
  - \mathbb E \left [ \| \nabla f(\mathbf x_k) \|_2^2 \right ],
%\leq & \frac{L^2 d}{2 b} \sum_{\ell = 1}^d \mu_\ell^2 - \| \nabla f(\mathbf x_k) \|_2^2,
\end{align}
where the first inequality holds since $-2 \inp{\mathbf a}{\mathbf b} \leq \| \mathbf a - \mathbf b \|_2^2 - \| \mathbf a \|_2^2$, and we have used the fact that $\mu_\ell = \mu$ in the second inequality.

Substituting \eqref{eq: prod_est_coord} into \eqref{eq: fsmooth_def_SVRG2_coord}, we have
\begin{align}\label{eq: fsmooth_def_SVRG2_coord2}
  \mathbb E \left [ f (\mathbf x_{k+1}^s) \right ]  
 \leq & \mathbb E \left [ f (\mathbf x_k^s) \right ] - \frac{\eta_k}{2} \mathbb E \left [ \| \nabla f(\mathbf x_k) \|_2^2 \right ]  + \frac{L}{2} \eta_k^2 \mathbb E\left [ \| \hat{\mathbf v}_k^s \|_2^2 \right ]   + \frac{L^2 d^2 \mu^2 \eta_k}{8}.   
\end{align}

In \eqref{eq: fsmooth_def_SVRG2_coord2}, we next bound $\mathbb E \left [ \| \hat {\mathbf v}_k^s \|_2^2 \right ] $.
Following \eqref{eq: E_Iv_v}-\eqref{eq: vks_norm_SVRG}, we have
 \begin{align}\label{eq: vks_norm_SVRG_coord0}
      \mathbb E \left [ \| \hat {\mathbf v}_k^s \|_2^2 \right ]  \leq 
      & \frac{2 \delta_n}{bn} \sum_{i=1}^n \mathbb E \left [ \| \hat \nabla f_{i}(\mathbf x_k^s) - \hat \nabla f_{i}(\mathbf x_0^s)  \|_2^2 \right ]   + 2 \mathbb E \left [\|\hat \nabla f(\mathbf x_k^s) \|_2^2  \right ].
 \end{align}
 The first term at RHS of \eqref{eq: vks_norm_SVRG_coord0} yields
  \begin{align} \label{eq: dist_est_twopoint_coord}
  &  \mathbb E \left [ \| \hat \nabla f_{i} ( \mathbf x_k^s ) - \hat \nabla f_{i} ( \mathbf x_0^s ) \|_2^2 \right   ]  \overset{\eqref{eq: grad_f_smooth_coord_all}}{=} \mathbb E \left [ \left \|  \sum_{\ell=1}^d \left  ( \frac{\partial f_{i,\mu_\ell}}{ \partial x_{k,\ell}^s }\mathbf e_\ell -\frac{\partial f_{i,\mu_\ell}}{ \partial x_{0,\ell}^s } \mathbf e_\ell  \right )  \right \|_2^2 \right ] \nonumber \\
\overset{\eqref{eq: sum_exp_bd}}{\leq}  & d  \sum_{\ell=1}^d \mathbb E \left [\left  \|   \frac{\partial f_{i,\mu_\ell}}{ \partial x_{k,\ell}^s }  - \frac{\partial f_{i,\mu_\ell}}{ \partial x_{0,\ell}^s } \right \|_2^2 \right ]
  \leq  L^2 d \sum_{\ell = 1}^d \mathbb E \left [ \| x_{\ell,k}^s - x_{\ell,0}^s \|_2^2 \right ] = L^2 d \mathbb E \left [ \| \mathbf x_{k}^s - \mathbf x_{0}^s \|_2^2 \right ],
 \end{align}
 where   $ f_{i,\mu_\ell} (\mathbf x) = \mathbb E_{u \sim U [-\mu_\ell, \mu_\ell]} f_i(\mathbf x + u \mathbf e_\ell)$ denotes the smooth function of $f_i$ with respect to its $\ell$th coordinate   (Lemma\,\ref{lemma: deter_smooth_gradest}), $x_{k,\ell}^s$ denotes the $\ell$th coordinate of $\mathbf x_k^s$,
 $\frac{\partial f_{i,\mu_\ell}}{ \partial x_{k,\ell}^s }$ is the $\ell$th partial derivative of $f_{i,\mu_\ell}$  at   $\mathbf x_k^s$, and
 the second inequality holds since $f_{i,\mu_\ell}(\mathbf x)$ is $L$-smooth (Lemma\,\ref{lemma: deter_smooth_gradest}) with respect to the $\ell$th coordinate.
From \eqref{eq: grad_diff_allcoord}, the second term at RHS of  \eqref{eq: vks_norm_SVRG_coord0} yields
\begin{align}\label{eq: grad_est_norm_deter_sig}
    \| \hat \nabla f(\mathbf x) \|_2^2 & \leq 2 \| \nabla f(\mathbf x) \|_2^2 + 2 \|  \hat{\nabla} f(\mathbf x) - \nabla f(\mathbf x)   \|_2^2    \overset{\eqref{eq: grad_diff_allcoord}}{\leq} 2 \| \nabla f(\mathbf x) \|_2^2 + \frac{L^2 d^2 \mu^2}{2},
\end{align}
where we have used the fact that $\mu_\ell = \mu$.
 
 Substituting
 \eqref{eq: dist_est_twopoint_coord}
 and
 \eqref{eq: grad_est_norm_deter_sig}
 into  \eqref{eq: vks_norm_SVRG_coord0}, we have 
 \begin{align}\label{eq: vks_norm_SVRG_coord}
     & \mathbb E \left [ \| \hat {\mathbf v}_k^s \|_2^2 \right ] \leq 
    \frac{2 L^2 d \textcolor{black}{\delta_n}}{b} \mathbb E \left [ \| \mathbf x_k^s - \mathbf x_0^s \|_2^2 \right ]
    +  4 \mathbb E \left [ \| \nabla f(\mathbf x) \|_2^2 \right ] + L^2 d^2 \mu^2.
   % \leq & \frac{2}{b} \mathbb E [ \| \hat \nabla f_{i_k} ( \mathbf x_k^s ) - \hat \nabla f_{i_k} ( \mathbf x_0^s )   \|_2^2] + 2 \mathbb E [\|\hat \nabla f(\mathbf x_k^s) \|_2^2  ],
 \end{align}
 
 Similar to \eqref{eq: norm_xk_dif}, we have
 \begin{align}\label{eq: norm_xk_dif_coord}
& \mathbb E \left [\| \mathbf x_{k+1}^s - \mathbf x_0^s \|_2^2 \right ] 
  \leq  \eta_k^2 \mathbb E \left [
   \| \hat {\mathbf v}_{k}^s\|_2^2
   \right ] + \mathbb E \left [
   \|
  \mathbf x_{k}^s - \mathbf x_0^s
  \|_2^2
   \right ]
   + \eta_k\mathbb E \left [
  \frac{1}{\beta_k}   \| \hat \nabla f(\mathbf x_{k}^s) \|_2^2 +  \beta_k  \| \mathbf x_{k}^s - \mathbf x_0^s \|_2^2
  \right ] \nonumber \\
&  \overset{\eqref{eq: grad_est_norm_deter_sig}}{\leq} 
\eta_k^2 \mathbb E \left [
   \| \hat {\mathbf v}_{k}^s\|_2^2
   \right ] + \mathbb E \left [
   \|
  \mathbf x_{k}^s - \mathbf x_0^s
  \|_2^2
   \right ]
   + \eta_k\mathbb E \left [
  \frac{2}{\beta_k}   \| \nabla f(\mathbf x_{k}^s) \|_2^2 +  \beta_k  \| \mathbf x_{k}^s - \mathbf x_0^s \|_2^2
  \right ]   + \frac{L^2 \mu^2 d^2 \eta_k}{\beta_k 2}.
\end{align}
Define the following   Lyapunov function,
\begin{align}\label{eq: Def_Rks_coord}
    R_{k}^s = \mathbb E \left [ f  (\mathbf x_k^s) + c_k \| \mathbf x_k^s - \mathbf x_0^s \|_2^2 \right ],
\end{align}
where $c_k > 0 $.

Based on \eqref{eq: fsmooth_def_SVRG2_coord2}  and \eqref{eq: norm_xk_dif_coord}, we obtain
\begin{align}
\label{eq: Rk_new_SCRG_coord}
    R_{k+1}^s  
    =  & \mathbb E \left [ f (\mathbf x_{k+1}^s) + c_{k+1} \| \mathbf x_{k+1}^s - \mathbf x_0^s \|_2^2 \right ] \nonumber \\
   \leq
    & \mathbb E [ f (\mathbf x_k^s)] - \left ( \frac{\eta_k}{2}-  \frac{c_{k+1} \eta_k}{2\beta_k} \right ) \mathbb E \left [ \| \nabla f (\mathbf x_k^s) \|_2^2 \right ]   + (c_{k+1} + c_{k+1} \beta_k \eta_k) \mathbb E \left [ \| \mathbf x_k^{s} - \mathbf x_0^s \|_2^2 \right ] \nonumber \\
   & + \left ( \frac{L}{2} \eta_k^2 + c_{k+1} \eta_k^2 \right ) \mathbb E \left [ \| \hat {\mathbf v}_k^s \|_2^2 \right ]   + \frac{d^2 L^2 \mu^2 \eta_k}{8}+ \frac{d^2 L^2 \mu^2  c_{k+1} \eta_k}{2 \beta_k}.
\end{align}
Substituting 
\eqref{eq: vks_norm_SVRG_coord} into \eqref{eq: Rk_new_SCRG_coord}, we have
\begin{align}
\label{eq: Rk_new_SCRG_2_coord}
   R_{k+1}^s  
  \leq &   \mathbb E \left [ f (\mathbf x_k^s) \right ] - \left ( \frac{\eta_k}{2}-  \frac{c_{k+1} \eta_k}{2\beta_k} \right ) \mathbb E \left [ \| \nabla f (\mathbf x_k^s) \|_2^2 \right ]   + (c_{k+1} + c_{k+1} \beta_k \eta_k) \mathbb E \left [ \| \mathbf x_k^{s} - \mathbf x_0^s \|_2^2 \right ] \nonumber \\
   & + \left ( \frac{L}{2} \eta_k^2 + c_{k+1} \eta_k^2 \right )\frac{2L^2d \textcolor{black}{ \delta_n }}{b} \mathbb E \left [ \| \mathbf x_k^s - \mathbf x_0^s \|_2^2 \right ]   + \left ( \frac{L}{2} \eta_k^2 + c_{k+1} \eta_k^2 \right ) 4 \mathbb E \left [ \| \nabla f  (\mathbf x_k^{s}) \|_2^2 \right ] \nonumber \\ 
& + \left ( \frac{L}{2} \eta_k^2 + c_{k+1} \eta_k^2 \right ) \mu^2 L^2 d^2   + \frac{d^2 L^2 \mu^2 \eta_k}{8}+ \frac{d^2 L^2 \mu^2  c_{k+1} \eta_k}{2 \beta_k}.
\end{align}

Based on the definition of $c_k$, i.e.,
\begin{align*}
    c_k =& \left ( 1 +   \beta_k \eta_k + \frac{2d L^2 \eta_k^2 \textcolor{black}{\delta_n}}{b}  \right )  c_{k+1} + \frac{ d L^3 \eta_k^2 \textcolor{black}{\delta_n}}{b},
\end{align*}
we can simplify \eqref{eq: Rk_new_SCRG_2_coord} to
\begin{align}
\label{eq: Rk_new_SCRG_2_2_coord}
   R_{k+1}^s  
\overset{ \eqref{eq: Def_Rks_coord}}{\leq} 
  R_k^s - \gamma_k \mathbb E [ \| \nabla f (\mathbf x_k^s) \|_2^2 ]  + \chi_k,
\end{align}
where we recall that
\begin{align*}
\gamma_k = & 
\frac{1}{2}\left( 1 - \frac{c_{k+1}}{\beta_k} \right ) \eta_k - 4\left(
\frac{L}{2} + c_{k+1}
\right ) \eta_k^2, \nonumber \\
\chi_k =  & 
\frac{1}{2}\left( \frac{1}{4} + \frac{c_{k+1}}{\beta_{k}} \right) L^2 d^2 \mu^2 \eta_k
+ \left (\frac{L}{2} + c_{k+1} \right ) \mu^2 L^2 d^2  \eta_k^2.
\end{align*}

Based on \eqref{eq: Rk_new_SCRG_2_2_coord}
and following the similar argument in \eqref{eq: Rk_new_SCRG_4}-\eqref{eq: grad_mid_Sm}, we have 
 \begin{align*}
     \sum_{s=1}^S \sum_{k=0}^{m-1}\gamma_k  \mathbb E[\| \nabla f(\mathbf x_k^s) \|_2^2] 
   \leq \mathbb E[  f(\tilde {\mathbf x}_{0}) - f^* ]  + S \chi_m.
\end{align*}
Consider $\bar{\gamma} = \min_k \gamma_k$ and the distribution of choosing $\bar{\mathbf x}$, we obtain
\begin{align}
       \mathbb E[\| \nabla f(\bar {\mathbf x}) \|_2^2]  
    \leq     \frac{ \mathbb E[  f(\tilde {\mathbf x}_{0}) - f^* ] }{T \bar \gamma} + \frac{S \chi_m}{T \bar \gamma}.
\end{align}

The rest of the proofs essentially follow along the lines of Corollary\,\ref{col: SVRG_simple} under a different parameter setting.

Since $c_k = c_{k+1} (1+ \theta)  + \frac{dL^3 \eta^2 \textcolor{black}{\delta_n} }{b} $, we 
have   $ c_k \leq c_0$ for any $k$, and  $\theta =  \beta \eta + \frac{2dL^2  \eta^2\textcolor{black}{\delta_n}}{b}$. This yields
\begin{align}\label{eq: c0_coord}
    c_0 & = \frac{dL^3 \eta^2 \textcolor{black}{\delta_n} }{b} \frac{(1+ \theta)^m - 1}{\theta}.
\end{align}
When $\eta = \rho/(Ld)$ and $\beta = L$ we have 
\begin{align}\label{eq: theta_coord}
    \theta = \frac{\rho}{ d} +   \frac{2  \rho^2 \textcolor{black}{\delta_n} }{  b d} \leq \frac{3 \rho }{ d}.
\end{align}
Substituting \eqref{eq: theta_coord} into \eqref{eq: c0_coord}, we have
\begin{align}\label{eq: ck_upper_coord}
   c_k \leq c_0 & = \textcolor{black}{\delta_n} \frac{d L^3}{b} \frac{\eta^2}{\theta } [(1+ \theta)^m - 1]   = \textcolor{black}{\delta_n} \frac{\rho L}{b + 2 \rho}  [(1+ \theta)^m - 1]   \leq  \textcolor{black}{\delta_n} \frac{L \rho}{b} (e - 1) \leq \textcolor{black}{\delta_n} \frac{2L \rho}{b},
 \end{align}
 where the second equality holds similar to  \eqref{eq: ck_upper} under $m = \ceil{\frac{d}{3 \rho} }$.
 
 Based on \eqref{eq: ck_upper_coord} and the definition of $\bar{\gamma}$, similar to \eqref{eq: c0_beta}-\eqref{eq: gamma_bar_bound} we can obtain
 \begin{align}\label{eq: gamma_bar_bound_coord}
     \bar \gamma \geq & 
      \eta \alpha_0,
 \end{align}
where $\alpha_0 > 0$   is independent of $T$, $d$ and $b$.

Since $\chi_m = \sum_k \chi_k $, it can be bounded as 
 \begin{align}
     \chi_m \leq  & m \eta^2   \left (\frac{L}{2} + c_0 \right ) \mu^2 L^2 d^2 + m \eta \frac{d^2 L^2 \mu^2 }{8}  + m \eta \frac{d^2 L^2 \mu^2  c_{0} }{2 \beta}.
 \end{align}
 From \eqref{eq: ck_upper_coord}, we have   
 $
 \frac{L}{2} + c_0 \leq \frac{L}{2} + 2 L \rho b^{-1} \textcolor{black}{\delta_n} \leq \frac{5 L}{2}
 $.  
Moreover, based on  $T = S m$ and  $\mu = \frac{1}{\sqrt{d}\sqrt{T}}$, we have
 \begin{align}\label{eq: const_error_bd_coord}
     \frac{S\chi_m}{ T \bar \gamma} \leq&
    \frac{5 L^2 \rho }{2 \alpha_0 T }  +
     \frac{dL^2}{8 \alpha_0 T}   +  \frac{d \rho L^2}{\alpha_0 bT} = O \left ( \frac{1}{T} + 
\frac{d }{T}   + \frac{d }{bT}
\right ),
 \end{align}
 where in the big $O$ notation, we  ignore the constant numbers that are independent of $L$, $d$, $b$, and $T$.
 
 Substituting \eqref{eq: gamma_bar_bound_coord} and 
 \eqref{eq: const_error_bd_coord} into \eqref{eq: grad_norm_SVRG_coord}, we have
\begin{align} \label{eq: grad_norm_SVRG_special_coord}
    \mathbb E[ \| \nabla f(\bar { \mathbf x }) \|_2^2 ] \leq &
\frac{[f(\tilde{\mathbf x}_0) - f^*]}{T \alpha_0 }
\frac{L d}{ \rho }
+ \frac{S \chi_m }{ T \bar \gamma}  
= O\left ( 
\frac{d}{T}  
\right ).
\end{align}

 \hfill $\square$
 
 \subsection{Auxiliary Lemmas}\label{app: aux}
 
 \begin{mylemma}\label{lemma: minibatch}
Let $\{ \mathbf z_i \}_{i=1}^n$ be a   sequence of $n$ vectors. Let $\mathcal I$ be a mini-batch of size $b$, which contains i.i.d. samples selected uniformly randomly (with replacement) from  $[n]$. Then
\begin{align}\label{eq: exp_minibatch_2}
    \mathbb E_{\mathcal I}\left [  \frac{1}{b} \sum_{i \in \mathcal I} \mathbf z_i    \right ]  
 %= \mathbb E_i[\mathbf z_i] 
 =  \frac{1}{n}\sum_{j=1}^n \mathbf z_j.
\end{align}
When $\sum_{i=1}^n \mathbf z_i = \mathbf 0$, then
\begin{align}\label{eq: exp_minibatch_special}
     \mathbb E_{\mathcal I}\left [ \left \|  \frac{1}{b} \sum_{i \in \mathcal I} \mathbf z_i \right \|_2^2 \right ] 
     %= \frac{1}{b} \mathbb E_i [\| \mathbf z_i \|_2^2]
 = \frac{1}{bn}\sum_{i=1}^n \| \mathbf z_i \|_2^2.
\end{align}
\end{mylemma}

\textbf{Proof:}
Based on the definition of $\mathcal I$, we immediately obtain
$
    \mathbb E_{\mathcal I}\left [  \frac{1}{b} \sum_{i \in \mathcal I} \mathbf z_i    \right ]  
 = \mathbb E_i[ \mathbf z_i] = \frac{1}{n}\sum_{j=1}^n \mathbf z_j$.

Since $\mathbb E_{i,j}[\mathbf z_i \mathbf z_j] = \mathbb E_i[\mathbf z_i] \mathbb E_j[\mathbf z_j] = \mathbf 0$ for $i \neq j$,  we have
\begin{align}\label{eq: exp_z}
    \mathbb E \left [ \left \|  \frac{1}{b} \sum_{i \in \mathcal I} \mathbf z_i \right \|_2^2 \right ]
 =  &  \frac{1}{b^2} 
 \sum_{i \in \mathcal I} \mathbb E[ \| \mathbf z_i \|_2^2] = \frac{1}{b} \mathbb E_i[ \| \mathbf z_i \|_2^2] = \frac{1}{bn}\sum_{i=1}^n \| \mathbf z_i \|_2^2.  
\end{align}
The proof is now complete.
\hfill $\square$

\textcolor{black}{
 \begin{mylemma}\label{lemma: minibatch_noReplace}
Let $\{ \mathbf z_i \}_{i=1}^n$ be a   sequence of $n$ vectors. Let $\mathcal I$ be a uniform random mini-batch of $[n]$ with size $b$ (no replacement in samples). Then
\begin{align}\label{eq: exp_minibatch_2_noReplace}
    \mathbb E_{\mathcal I}\left [  \frac{1}{b} \sum_{i \in \mathcal I} \mathbf z_i    \right ]  
 %= \mathbb E[\mathbf z_i]  
 =  \frac{1}{n}\sum_{j=1}^n \mathbf z_j.
\end{align}
When $\sum_{j=1}^n \mathbf z_j = \mathbf 0$, then
\begin{align}\label{eq: exp_minibatch_special_noReplace}
     \mathbb E_{\mathcal I}\left [ \left \|  \frac{1}{b} \sum_{i \in \mathcal I} \mathbf z_i \right \|_2^2 \right ]  
     %= \frac{1}{b} \mathbb E [\| \mathbf z_i \|_2^2]
 \leq \frac{\mathcal I(b < n)}{bn}\sum_{i=1}^n \| \mathbf z_i \|_2^2,
\end{align}
where $I$ is an indicator function, which is equal to $1$ if $b < n$ and $0$ if $b = n$.
\end{mylemma}
\textbf{Proof:}
See \cite[Lemma\,A.1]{lei2017non}. \hfill $\square$
}

\begin{mylemma}\label{lemma: sum_exp_bd}
For  variables $\{ \mathbf z_i \}_{i=1}^n$, we have
%{\color{green} Do we really need this lemma as it trivially follows from Jensen's inequality}
\begin{align}\label{eq: sum_exp_bd}
 \left \| \sum_{i=1}^n \mathbf z_i \right  \|_2^2  \leq n \sum_{i=1}^n  \| \mathbf z_i \|_2^2.
\end{align}
\end{mylemma}
\textbf{Proof}: Since $\phi(\mathbf x) = \| \mathbf x \|_2^2$ is convex, the 
 Jensen's inequality yields $\| \frac{1}{n} \sum_i \mathbf z_i \|_2^2 \leq \frac{1}{n} \sum_i \| \mathbf z_i \|_2^2$.
 \hfill $\square$

\begin{mylemma}\label{lemma: Lsmooth_v2}
if $f$ is $L$-smooth, then for any $\mathbf x, \mathbf y \in \mathbb R^d$
\begin{align}\label{eq: Lsmooth_v2}
    | f(\mathbf x) - f(\mathbf y) - \inp{\nabla f_i(\mathbf y)}{\mathbf x - \mathbf y} | \leq \frac{L}{2} \| \mathbf x - \mathbf y \|_2^2.
\end{align}
\end{mylemma}
 \textbf{Proof}: This is a direct consequence of A2 \citep{lei2017non}. \hfill $\square$
 
\subsection{Application: black-box classification}
\label{supp: experiment_classification}
\paragraph{Real dataset}
Our dataset consists of 
 $N = 1000$ crystalline materials/compounds, each of which corresponds to a numerical valued feature vector $\mathbf a_i$. The feature vector encodes chemical information regarding constituent elements. There exist $d=145$ attributes, such as, stoichiometric properties, elemental statistics, electronic structure properties attributes, and ionic compound attributes \citep{magpie}. The label information  $y_i\in \{0,1\}$ (conductor against insulator)  is determined using DFT calculations \citep{vasp}. We equally divided the data into a training and test set.
 
\paragraph{Parameter setting}
In our ZO algorithms, unless specified otherwise, 
the length of each epoch is set by $m = 50$, the mini-batch size is $b = 10$, the number of random direction samples is $q = 10$, the initial value is given by 
 $\tilde{\mathbf x}_0 = \mathbf 0$, and the smoothing parameter  follows
$
\mu = O( 1/\sqrt{dT} )
$. 
%For SGD, we   choose $\eta = 0.1$.  
% For SVRG, we choose
% $\eta = O( {b}/{n^{\frac{2}{3}}} )$ suggested by   \cite[Theorem\,7]{reddi2016stochastic}. 
For ZO-SGD, ZO-SVRC and ZO-SVRG, we choose
$
\eta = O( 1/d )
$ suggested by  Corollary \ref{col: SVRG_simple} and \citep[Corollary\,3.3]{ghadimi2013stochastic}. 
Also ZO-SVRC updates $J=1$ coordinates per iteration within an epoch.
%where $\alpha = 0.8$. 

\subsection{Application: generating universal adversarial perturbations from black-box DNNs} 
\label{appendix: adv_app}
\paragraph{Problem formulation}
In image classification, adversarial examples refer to carefully crafted perturbations such that, when added to the natural images, are visually imperceptible but will lead the target model to misclassify.
When testing the robustness of a deployed black-box DNN (e.g., an online image classification service), the model parameters are hidden and acquiring its gradient is inadmissible. But one has access to the input-output correspondence of the target model  $F(\cdot)$, rendering generating adversarial examples a ZO optimization problem. 

We consider the task of generating a universal perturbation to a batch of $n = 10$ images via iteratively querying the target DNN. These images are selected from the class of digit ``1'' and are all originally correctly classified by the DNN. In problem \eqref{eq: prob_ori}, let $f_i(\mathbf x)= c \cdot \max \{ F_{y_i} (0.5 \cdot \tanh ( \tanh^{-1} 2 \mathbf  a_i + \mathbf x)) - \max_{j \neq y_i} F_j(0.5 \cdot \tanh ( \tanh^{-1} 2 \mathbf a_i + \mathbf x)) , 0 \} + \|0.5 \cdot \tanh ( \tanh^{-1} 2 \mathbf a_i + \mathbf x) - \mathbf a_i \|_2^2  $ be the designed attack loss function of the $i$th image \cite{chen2017zoo,carlini2017towards}. 
Here  $(\mathbf a_i,y_i)$ denotes the pair of the $i$th natural image $\mathbf a_i \in [-0.5,0.5]^d$ and its original class label $y_i$. The function $F(\mathbf z)=[F_1(\mathbf z),\ldots,F_K(\mathbf z)]$ outputs the model prediction scores (e.g., log-probabilities) of the input $\mathbf z$ in  all $K$ image classes.
The $\tanh$ operation ensures the generated adversarial example $0.5 \cdot \tanh ( \tanh^{-1} \mathbf a_i + \mathbf x)$ still lies in the valid image space $[-0.5,0.5]^d$.  The regularization parameter $c$  trades off adversarial success and the $\ell_2$ distortion of adversarial examples. In our experiment, we set $c = 1$ and use the log-probability as the model output. 
The reported $\ell_2$ distortion is the least averaged distortion over the $n$ successful adversarial images relative to the original images among the $S$ iterations.

\paragraph{Generated adversarial images}
Table \ref{table:digit1} displays the original images and their adversarial examples generated by ZO-SGD and ZO-SVRG. Their statistics  are given in Fig. \ref{Fig3}. Table \ref{table:digit4} shows another visual comparison chart of digit class ``4''.

% The regularization parameter $c$  trades off adversarial success and $\ell_2$ distortion, $F(\mathbf z)=[F_1(\mathbf z),\ldots,F_K(\mathbf z)]$ outputs the model prediction scores (e.g., log-probabilities) of the input $\mathbf z$ in  all $K$ image classes, and $(\mathbf a_i,y_i)$ denotes the pair of the $i$th natural image $\mathbf a_i \in [-0.5,0.5]^d$ and its original class label $y_i$. The $\tanh$ operation ensures the generated adversarial example $0.5 \cdot \tanh ( \tanh^{-1} \mathbf a_i + \mathbf x)$ still lies in the valid image space $[-0.5,0.5]^d$.

\begin{table*}
  [ht] \caption{Comparison of generated adversarial examples from a black-box DNN on MNIST: digit class ``1''.} \label{table:digit1}
  \begin{adjustbox}{max width=\textwidth }
  \begin{tabular}
      {ccccccccccc}
      \hline
      	Image ID & 2 & 5 & 14 & 29 & 31 & 37 & 39 & 40 & 46 & 57 \\
      \hline &&&&&&&&&& \vspace{-0.2cm} \\
      	Original &
        \parbox[c]{2.2em}{\includegraphics[width=0.4in]{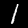}} &
        \parbox[c]{2.2em}{\includegraphics[width=0.4in]{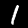}} &
        \parbox[c]{2.2em}{\includegraphics[width=0.4in]{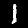}} &
        \parbox[c]{2.2em}{\includegraphics[width=0.4in]{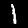}} &
        \parbox[c]{2.2em}{\includegraphics[width=0.4in]{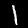}} &
        \parbox[c]{2.2em}{\includegraphics[width=0.4in]{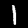}} &
        \parbox[c]{2.2em}{\includegraphics[width=0.4in]{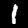}} &
        \parbox[c]{2.2em}{\includegraphics[width=0.4in]{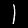}} &
        \parbox[c]{2.2em}{\includegraphics[width=0.4in]{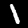}} &
        \parbox[c]{2.2em}{\includegraphics[width=0.4in]{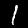}}
      \vspace{0.2cm} \\
      
      \hline &&&&&&&&&& \vspace{-0.2cm} \\
      	\thead{ZO-SGD} &
        \parbox[c]{2.2em}{\includegraphics[width=0.4in]{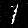}} &
        \parbox[c]{2.2em}{\includegraphics[width=0.4in]{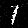}} &
        \parbox[c]{2.2em}{\includegraphics[width=0.4in]{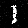}} &
        \parbox[c]{2.2em}{\includegraphics[width=0.4in]{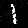}} &
        \parbox[c]{2.2em}{\includegraphics[width=0.4in]{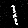}} &
        \parbox[c]{2.2em}{\includegraphics[width=0.4in]{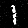}} &
        \parbox[c]{2.2em}{\includegraphics[width=0.4in]{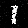}} &
        \parbox[c]{2.2em}{\includegraphics[width=0.4in]{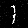}} &
        \parbox[c]{2.2em}{\includegraphics[width=0.4in]{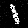}} &
        \parbox[c]{2.2em}{\includegraphics[width=0.4in]{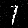}} \\
        Classified as & 7 & 7 & 3 & 3 & 3 & 3 & 3 & 3 & 3 & 7 \\
      \hline
      
      \hline &&&&&&&&&& \vspace{-0.2cm} \\
      	\thead{ZO-SVRG\\q = 10} &
        \parbox[c]{2.2em}{\includegraphics[width=0.4in]{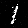}} &
        \parbox[c]{2.2em}{\includegraphics[width=0.4in]{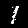}} &
        \parbox[c]{2.2em}{\includegraphics[width=0.4in]{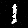}} &
        \parbox[c]{2.2em}{\includegraphics[width=0.4in]{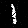}} &
        \parbox[c]{2.2em}{\includegraphics[width=0.4in]{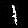}} &
        \parbox[c]{2.2em}{\includegraphics[width=0.4in]{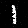}} &
        \parbox[c]{2.2em}{\includegraphics[width=0.4in]{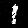}} &
        \parbox[c]{2.2em}{\includegraphics[width=0.4in]{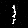}} &
        \parbox[c]{2.2em}{\includegraphics[width=0.4in]{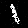}} &
        \parbox[c]{2.2em}{\includegraphics[width=0.4in]{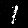}} \\
        Classified as & 3 & 3 & 3 & 3 & 3 & 3 & 3 & 3 & 3 & 3 \\
      \hline
      
      \hline &&&&&&&&&& \vspace{-0.2cm} \\
      	\thead{ZO-SVRG\\q = 20} &
        \parbox[c]{2.2em}{\includegraphics[width=0.4in]{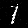}} &
        \parbox[c]{2.2em}{\includegraphics[width=0.4in]{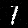}} &
        \parbox[c]{2.2em}{\includegraphics[width=0.4in]{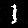}} &
        \parbox[c]{2.2em}{\includegraphics[width=0.4in]{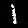}} &
        \parbox[c]{2.2em}{\includegraphics[width=0.4in]{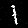}} &
        \parbox[c]{2.2em}{\includegraphics[width=0.4in]{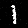}} &
        \parbox[c]{2.2em}{\includegraphics[width=0.4in]{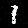}} &
        \parbox[c]{2.2em}{\includegraphics[width=0.4in]{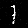}} &
        \parbox[c]{2.2em}{\includegraphics[width=0.4in]{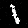}} &
        \parbox[c]{2.2em}{\includegraphics[width=0.4in]{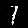}} \\
        Classified as & 7 & 7 & 3 & 3 & 3 & 3 & 3 & 3 & 3 & 7 \\
      \hline
      
      \hline &&&&&&&&&& \vspace{-0.2cm} \\
      	\thead{ZO-SVRG\\q = 30} &
        \parbox[c]{2.2em}{\includegraphics[width=0.4in]{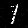}} &
        \parbox[c]{2.2em}{\includegraphics[width=0.4in]{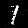}} &
        \parbox[c]{2.2em}{\includegraphics[width=0.4in]{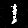}} &
        \parbox[c]{2.2em}{\includegraphics[width=0.4in]{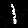}} &
        \parbox[c]{2.2em}{\includegraphics[width=0.4in]{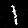}} &
        \parbox[c]{2.2em}{\includegraphics[width=0.4in]{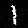}} &
        \parbox[c]{2.2em}{\includegraphics[width=0.4in]{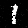}} &
        \parbox[c]{2.2em}{\includegraphics[width=0.4in]{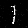}} &
        \parbox[c]{2.2em}{\includegraphics[width=0.4in]{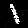}} &
        \parbox[c]{2.2em}{\includegraphics[width=0.4in]{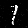}} \\
        Classified as & 7 & 7 & 3 & 3 & 3 & 3 & 3 & 3 & 3 & 7 \\
      \hline
  \end{tabular}
  \end{adjustbox}
\end{table*}

\begin{table*}
  [ht] \caption{Comparison of generated adversarial examples from a black-box DNN on MNIST: digit class ``4''.} \label{table:digit4}
  \begin{adjustbox}{max width=\textwidth}
  \begin{tabular}
      {ccccccccccc}
      \hline
      	Image ID & 4 & 6 & 19 & 24 & 27 & 33 & 42 & 48 & 49 & 56 \\
      \hline &&&&&&&&&& \vspace{-0.2cm} \\
      	Original &
        \parbox[c]{2.2em}{\includegraphics[width=0.4in]{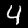}} &
        \parbox[c]{2.2em}{\includegraphics[width=0.4in]{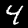}} &
        \parbox[c]{2.2em}{\includegraphics[width=0.4in]{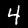}} &
        \parbox[c]{2.2em}{\includegraphics[width=0.4in]{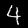}} &
        \parbox[c]{2.2em}{\includegraphics[width=0.4in]{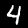}} &
        \parbox[c]{2.2em}{\includegraphics[width=0.4in]{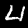}} &
        \parbox[c]{2.2em}{\includegraphics[width=0.4in]{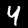}} &
        \parbox[c]{2.2em}{\includegraphics[width=0.4in]{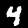}} &
        \parbox[c]{2.2em}{\includegraphics[width=0.4in]{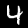}} &
        \parbox[c]{2.2em}{\includegraphics[width=0.4in]{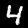}}
      \vspace{0.2cm} \\
      
      \hline &&&&&&&&&& \vspace{-0.2cm} \\
      	\thead{ZO-SGD} &
        \parbox[c]{2.2em}{\includegraphics[width=0.4in]{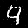}} &
        \parbox[c]{2.2em}{\includegraphics[width=0.4in]{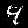}} &
        \parbox[c]{2.2em}{\includegraphics[width=0.4in]{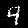}} &
        \parbox[c]{2.2em}{\includegraphics[width=0.4in]{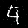}} &
        \parbox[c]{2.2em}{\includegraphics[width=0.4in]{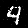}} &
        \parbox[c]{2.2em}{\includegraphics[width=0.4in]{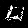}} &
        \parbox[c]{2.2em}{\includegraphics[width=0.4in]{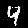}} &
        \parbox[c]{2.2em}{\includegraphics[width=0.4in]{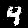}} &
        \parbox[c]{2.2em}{\includegraphics[width=0.4in]{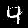}} &
        \parbox[c]{2.2em}{\includegraphics[width=0.4in]{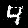}} \\
        Classified as & 9 & 9 & 9 & 9 & 9 & 2 & 9 & 9 & 9 & 9 \\
      \hline
      
      \hline &&&&&&&&&& \vspace{-0.2cm} \\
      	\thead{ZO-SVRG\\q = 10} &
        \parbox[c]{2.2em}{\includegraphics[width=0.4in]{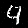}} &
        \parbox[c]{2.2em}{\includegraphics[width=0.4in]{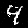}} &
        \parbox[c]{2.2em}{\includegraphics[width=0.4in]{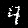}} &
        \parbox[c]{2.2em}{\includegraphics[width=0.4in]{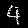}} &
        \parbox[c]{2.2em}{\includegraphics[width=0.4in]{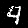}} &
        \parbox[c]{2.2em}{\includegraphics[width=0.4in]{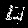}} &
        \parbox[c]{2.2em}{\includegraphics[width=0.4in]{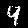}} &
        \parbox[c]{2.2em}{\includegraphics[width=0.4in]{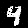}} &
        \parbox[c]{2.2em}{\includegraphics[width=0.4in]{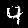}} &
        \parbox[c]{2.2em}{\includegraphics[width=0.4in]{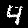}} \\
        Classified as & 9 & 9 & 7 & 9 & 9 & 5 & 9 & 9 & 9 & 9 \\
      \hline
      
      \hline &&&&&&&&&& \vspace{-0.2cm} \\
      	\thead{ZO-SVRG\\q = 20} &
        \parbox[c]{2.2em}{\includegraphics[width=0.4in]{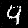}} &
        \parbox[c]{2.2em}{\includegraphics[width=0.4in]{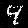}} &
        \parbox[c]{2.2em}{\includegraphics[width=0.4in]{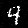}} &
        \parbox[c]{2.2em}{\includegraphics[width=0.4in]{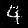}} &
        \parbox[c]{2.2em}{\includegraphics[width=0.4in]{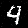}} &
        \parbox[c]{2.2em}{\includegraphics[width=0.4in]{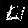}} &
        \parbox[c]{2.2em}{\includegraphics[width=0.4in]{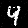}} &
        \parbox[c]{2.2em}{\includegraphics[width=0.4in]{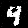}} &
        \parbox[c]{2.2em}{\includegraphics[width=0.4in]{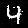}} &
        \parbox[c]{2.2em}{\includegraphics[width=0.4in]{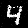}} \\
        Classified as & 9 & 9 & 9 & 9 & 9 & 5 & 9 & 9 & 9 & 9 \\
      \hline
      
      \hline &&&&&&&&&& \vspace{-0.2cm} \\
      	\thead{ZO-SVRG\\q = 30} &
        \parbox[c]{2.2em}{\includegraphics[width=0.4in]{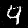}} &
        \parbox[c]{2.2em}{\includegraphics[width=0.4in]{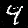}} &
        \parbox[c]{2.2em}{\includegraphics[width=0.4in]{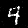}} &
        \parbox[c]{2.2em}{\includegraphics[width=0.4in]{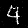}} &
        \parbox[c]{2.2em}{\includegraphics[width=0.4in]{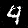}} &
        \parbox[c]{2.2em}{\includegraphics[width=0.4in]{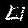}} &
        \parbox[c]{2.2em}{\includegraphics[width=0.4in]{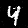}} &
        \parbox[c]{2.2em}{\includegraphics[width=0.4in]{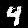}} &
        \parbox[c]{2.2em}{\includegraphics[width=0.4in]{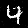}} &
        \parbox[c]{2.2em}{\includegraphics[width=0.4in]{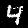}} \\
        Classified as & 9 & 9 & 7 & 9 & 9 & 0 & 9 & 9 & 9 & 9 \\
      \hline
  \end{tabular}
  \end{adjustbox}
\end{table*}

\end{document}